\documentclass[]{opendatalab}
\usepackage{xcolor}
\usepackage{makecell}
\usepackage{longtable}
\usepackage{listings}
\usepackage[numbers]{natbib}

\usepackage{soul}
\usepackage{graphicx}
\usepackage{pifont}
\usepackage{amssymb}
\usepackage{booktabs}
\usepackage{xspace}
\usepackage{todonotes}
\usepackage{array}
\usepackage{pdflscape}
\usepackage{rotating}
\usepackage{siunitx}
\usepackage{tcolorbox}
\newcommand{\cmark}{\ding{51}}
\newcommand{\xmark}{\ding{55}}
\newcommand{\method}{BioMatrix\xspace}
\newcommand{\bound}[1]{\textcolor{gray}{#1}}

\newcommand{\hficon}{\raisebox{-0.2em}{\includegraphics[height=1em]{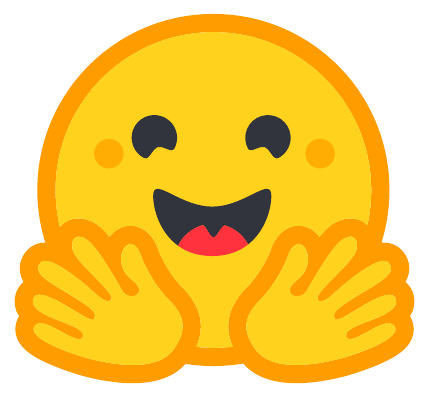}}}
\newcommand{\ghicon}{\raisebox{-0.2em}{\includegraphics[height=1em]{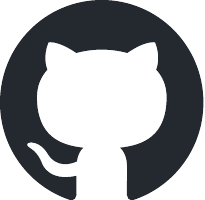}}}
\newtcolorbox{DefinitionBox}{
  colback=blue!5,
  colframe=blue!80,
  boxrule=0.5pt,
  arc=2pt,
  left=2pt,
  right=2pt,
  top=2pt,
  bottom=2pt,
}

\newtcolorbox{CorollaryBox}{
  colback=yellow!10,
  colframe=orange!60,
  boxrule=0.5pt,
  arc=2pt,
  left=2pt,
  right=2pt,
  top=2pt,
  bottom=2pt,
}

\definecolor{codegreen}{rgb}{0,0.6,0}
\definecolor{codegray}{rgb}{0.5,0.5,0.5}
\definecolor{codepurple}{rgb}{0.58,0,0.82}
\definecolor{backcolour}{rgb}{0.95,0.95,0.92}
\definecolor{promptcolor}{HTML}{D1D0F2}
\definecolor{promptcolorheader}{HTML}{bdbcec}
\newcommand{\promptbox}[2]{
\begin{tcolorbox}[
top=0.3em,bottom=0.3em,left=0.5em,right=0.5em,
toptitle=0.3em,bottomtitle=0.2em,boxsep=0pt,
colframe=promptcolorheader,colback=promptcolor!50,boxrule=0.5pt,
]
\footnotesize
\end{tcolorbox}
}
\lstdefinestyle{mystyle}{
    backgroundcolor=\color{backcolour},   
    commentstyle=\color{codegreen},
    keywordstyle=\color{magenta},
    numberstyle=\tiny\color{codegray},
    stringstyle=\color{codepurple},
    basicstyle=\ttfamily\footnotesize,
    breakatwhitespace=false,         
    breaklines=true,                 
    captionpos=b,                    
    keepspaces=true,                 
    numbers=left,                    
    numbersep=5pt,                  
    showspaces=false,                
    showstringspaces=false,
    showtabs=false,                  
    tabsize=2
}

\title{Towards a Comprehensive Biological Foundation Model Spanning the Modality Matrix of Sequences, Structures, \mbox{and Language}}
\shorttitle{BioMatrix}

\author[1,2*]{Qizhi Pei}
\author[3,4*]{Zhimeng Zhou}
\author[1,2*]{Yi Duan}
\author[2,3,4]{Yiyang Zhao}
\author[5,2]{Wei Li}
\author[2]{Han Guo}
\author[6]{Liang He}
\author[6]{Chengping Li}
\author[3]{Chang-Yu Hsieh}
\author[2]{Conghui He}
\author[7\dagger]{Rui Yan}
\author[2\dagger]{Lijun Wu}

\affiliation[1]{Gaoling School of Artificial Intelligence, Renmin University of China} 
\affiliation[2]{OpenDataLab, Shanghai Artificial Intelligence Laboratory}
\affiliation[3]{Zhejiang University}
\affiliation[4]{Shanghai Innovation Institute}
\affiliation[5]{East China Normal University}
\affiliation[6]{Zhongguancun Academy}
\affiliation[7]{School of Artificial Intelligence, Wuhan University}

\abstract{
We present BioMatrix, the first multimodal foundation model that natively integrates sequences, structures, and natural language for both molecules and proteins within a single decoder-only architecture. Existing biological foundation models pursue native multimodality and broad entity coverage separately: those that fuse multiple modalities under a shared objective remain confined to a single entity type, while those spanning multiple entity types either omit explicit structural modeling or rely on adapter-based designs in which the model cannot natively generate the very modalities it can read. BioMatrix closes this gap by mapping molecular sequences (supporting both SMILES and SELFIES notations), molecular structures, protein sequences, protein structures, and natural language into a shared discrete token space through a unified tokenization scheme, so that all modalities are consumed and produced uniformly under a single next-token prediction objective---without external encoders, projection adapters, or modality-specific output heads. Built upon the Qwen3 language model (1.7B and 4B), BioMatrix is continually pretrained on 304.4 billion tokens spanning general and domain-specific text, sequence and structure views of molecules and proteins, and cross-modal corpora that interleave biomolecular entities with scientific text and link distinct entities through molecule--protein and protein--protein interaction data. After tuning on a comprehensive suite of downstream applications covering 80 tasks across 6 categories---encompassing single-entity and multi-entity understanding and generation tasks across and within modalities---BioMatrix achieves state-of-the-art or competitive performance on 77 out of 80 tasks, demonstrating that a single, natively multimodal generalist model can effectively match or surpass specialized approaches across a wide range of biological tasks.
}

\correspondence{Rui Yan (\email{rui.yan@whu.edu.cn}) and Lijun Wu (\email{wulijun@pjlab.org.cn})}

\metadata[Links]{\ghicon~\href{https://github.com/QizhiPei/biomatrix}{GitHub: QizhiPei/biomatrix}\quad \hficon~\href{https://huggingface.co/collections/QizhiPei/biomatrix}{HuggingFace: QizhiPei/biomatrix}}

\begin{document}

\maketitle
\footnotetext[1]{Equal Contribution.}
\footnotetext[2]{Corresponding Authors.}
\enlargethispage{1.2cm}

\vspace{-0.5cm}
\noindent\begin{minipage}{\textwidth}
    \centering
    \makebox[\textwidth][c]{\includegraphics[width=1.10\textwidth,height=0.36\textheight,keepaspectratio]{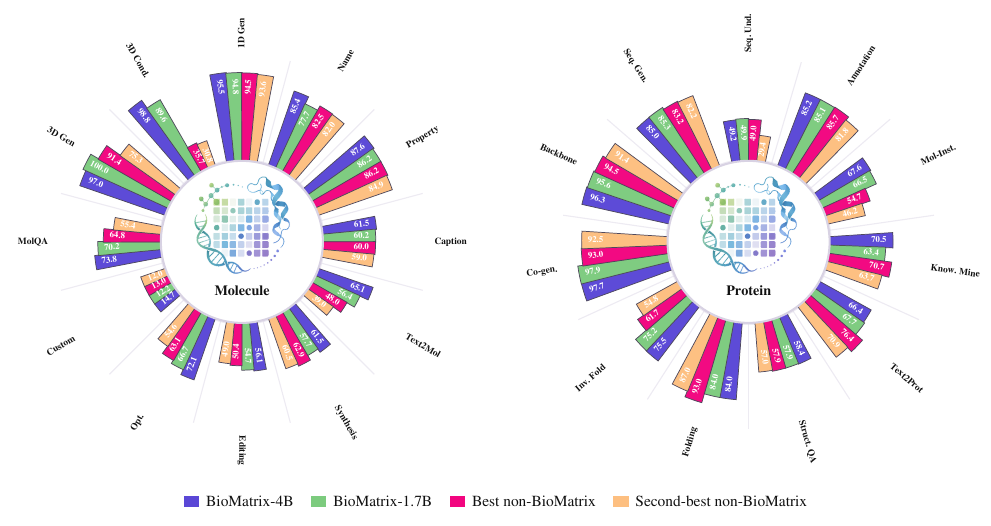}}
\end{minipage}
\newpage

\begingroup
\small
\setcounter{tocdepth}{2}
\setlength{\parskip}{0pt}
\tableofcontents
\endgroup
\newpage

\section{Introduction}
\label{section:intro}

Foundation models trained on large-scale data have achieved remarkable success in natural language processing and beyond, demonstrating strong capabilities in reasoning, generation, and cross-task generalization~\cite{anthropic2026claude46,qwen3,deepseek_r1,interns1_pro}.
Extending this paradigm to the biological and chemical sciences holds transformative potential for drug discovery, protein engineering, and molecular design~\cite{llm_sci_survey_zju,llm_sci_survey,bl_survey}.
Biological entities, however, are not defined by a single representation. A molecule or protein may be specified as a sequence, realized as a spatial structure, and interpreted through natural-language descriptions of function, activity, disease relevance, or experimental context. These views play complementary roles: sequences provide compact symbolic identity, structures determine physical interaction and function, and language carries both human design intent and the accumulated knowledge of the literature. Biological tasks also rarely involve one entity in isolation; drug discovery, protein engineering, and mechanism understanding routinely depend on molecule--protein and protein--protein relationships. A general biological foundation model should therefore support both \emph{native multimodality}---treating sequence, structure, and language as first-class inputs and outputs of the same generative model---and \emph{broad entity coverage} across molecules, proteins, and their interactions. Despite rapid progress on individual fronts, no existing model meets both requirements at once.

Existing efforts toward such a unified model can be organized along two largely independent axes: \emph{how biomolecular content is coupled with natural language}, and \emph{how broadly different biomolecular entity types are covered}. Progress along each axis has so far been pursued in isolation, and the two have yet to be reconciled within a single architecture.

\noindent \textbf{Axis 1: coupling biomolecular content with natural language.}
Work along this axis broadly falls into two paradigms. The \emph{adapter-based} paradigm keeps domain encoders and a language model as separate components, as in BioMedGPT~\cite{biomedgpt} and 3D-MoLM~\cite{3dmolm}. While this paradigm benefits from specialized encoders, it introduces an input--output asymmetry: biomolecular content enters the LLM as continuous encoder features but must be emitted through the LLM's text vocabulary, making native generation of non-textual modalities such as molecular or protein structures difficult without external decoders. The \emph{native-tokenization} paradigm instead discretizes biomolecular content into tokens consumed under the same next-token objective as language, mirroring the shift from late-fusion adapters to early-fusion token streams in vision--language modeling~\cite{chameleon,janus_pro,gpt_4o}. In biology, this direction is enabled by line notations such as SMILES, SELFIES, and amino acid strings, as well as recent SE(3)-invariant geometry tokenizers for molecules~\cite{molstructok} and proteins~\cite{esm3,foldseek,saprot,gcp_vqvae}. Existing models in this paradigm remain partial: BioT5~\cite{biot5,biot5+}, NatureLM~\cite{naturelm}, and SciReasoner~\cite{scireasoner} unify sequences with text but omit explicit structure, while ESM3~\cite{esm3} includes sequence and structure but covers only proteins.

\noindent \textbf{Axis 2: covering multiple biomolecular entity types.}
A parallel line of work pursues unified modeling across biomolecular entities themselves, often without natural language. AlphaFold3~\cite{alphafold3} jointly represents proteins, nucleic acids, and molecules for structure prediction, and ODesign~\cite{odesign} extends this direction to all-atom generative design. More recently, LOGOS~\cite{logos} from the Tongyi team broadens the scope further by encoding diverse scientific objects under a shared grammar, including proteins, antibodies, small molecules, reactions, materials, pockets, and protein--ligand complexes. This grammar-based formulation is complementary to our goal: it emphasizes symbolic scientific object descriptions and contact/constraint patterns, whereas \method unifies natural language, biomolecular sequences, and explicit structure tokens in a single decoder-only next-token interface. Overall, these models show that broad cross-entity coverage is feasible, but their token spaces and objectives remain largely separate from language-grounded prediction over both sequence and structure. Consequently, design intents must be specified through structured configurations rather than functional descriptions, and the models cannot directly leverage the textual knowledge available in biological literature.

\noindent \textbf{The remaining gap.}
As Table~\ref{tab:model_comp} makes concrete, native multimodality and broad entity coverage have so far been pursued separately: models that fuse sequence, structure, and text are confined to a single entity type, while models that span multiple entity types either lack natural language or omit explicit structural modeling. With mature sequence, text, and SE(3)-invariant geometry tokenizers now available, the remaining challenge is integration: reconciling heterogeneous tokenizers into one shared vocabulary and curating supervision dense enough to bind sequences, structures, language, and cross-entity relationships under a unified next-token objective.

\begin{table}[t]
\centering
\caption{Comparison of modality coverage across multimodal biological foundation models. $\triangle$: structural/spatial interactions are encoded as discrete token sequences rather than explicit 3D coordinates or geometric inputs.}
\label{tab:model_comp}
\begin{tabular}{l|cc|cc|c}
\toprule
\multirow{2}{*}{Model} & \multicolumn{2}{c|}{Molecule} & \multicolumn{2}{c|}{Protein} & Text \\
 & 1D Seq & 3D Struct & 1D Seq & 3D Struct & NL \\
\midrule
ESM3~\cite{esm3}       & \xmark & \xmark & \cmark & \cmark & \cmark \\
3D-MoLM~\cite{3dmolm}    & \cmark & \cmark & \xmark & \xmark & \cmark \\
AlphaFold3~\cite{alphafold3} & \cmark & \cmark & \cmark & \cmark & \xmark \\
ODesign~\cite{odesign} & \xmark & \cmark & \cmark & \cmark & \xmark \\
BioT5/BioT5+~\cite{biot5,biot5+}  & \cmark & \xmark & \cmark & \xmark & \cmark \\
BioMedGPT~\cite{biomedgpt} & \cmark & \xmark & \cmark & \xmark & \cmark \\
NatureLM~\cite{naturelm} & \cmark & \xmark & \cmark & \xmark & \cmark \\
SciReasoner~\cite{scireasoner} & \cmark & \xmark & \cmark & \xmark & \cmark \\
LOGOS~\cite{logos} & \cmark & $\triangle$ & \cmark & $\triangle$ & \xmark \\
\midrule
\textbf{\method} & \cmark & \cmark & \cmark & \cmark & \cmark \\
\bottomrule
\end{tabular}
\end{table}

We close this gap with \textbf{\method}, a multimodal foundation model that, to our knowledge, is the first to natively integrate sequence, structure, and natural language for both molecules and proteins within a single decoder-only architecture. Built upon Qwen3~\cite{qwen3} (1.7B and 4B), \method serializes all supported modalities into one discrete token stream: SMILES and SELFIES for molecular sequences, SELFIES-aligned tokens for molecular conformations, residue-level sequence tokens and GCP-VQVAE structure tokens for proteins, and the original language-model vocabulary for natural language. Modality-specific boundary tokens allow these segments to be freely interleaved, so tasks that are usually architecturally distinct---captioning, text-conditioned design, folding, inverse folding, structure generation, and interaction prediction---become different conditional generation patterns under one next-token objective.

Several design choices make this interface practical rather than merely conceptual. For molecules, we adapt MolStrucTok~\cite{molstructok} with a branch-decoupled decoder and compose frequent SELFIES--structure pairs into joint tokens, binding chemical identity to local geometry without exploding the vocabulary. For proteins, we retain a factored representation between residue tokens and GCP-VQVAE~\cite{gcp_vqvae} structure tokens, preserving residue-level sequence--structure alignment while avoiding a prohibitively large joint residue--structure vocabulary. To train the resulting model, we construct a 304.4-billion-token corpus spanning scientific text, molecule- and protein-centric multi-view data, interleaved biomedical text, and molecule--protein/protein--protein interaction data, using stochastic composition and protein three-view instances to expose both isolated and jointly grounded modality views.

We evaluate the resulting backbone after instruction tuning on 80 tasks across 6 categories, covering molecular generation and prediction, molecule--text translation, molecular conformer generation, protein understanding and design, protein structure tasks, and biomolecular interaction prediction. \method achieves state-of-the-art (SOTA) or competitive performance on \textbf{77 out of 80} tasks, with the largest gains on settings that exercise the unified design, including text-based molecule generation, property-conditioned molecular conformer generation, protein sequence--structure co-generation, and structure-grounded molecule--protein affinity prediction. Our main contributions are as follows:

\noindent $\bullet$ \textbf{A unified architecture along both axes.} We propose \method, the first foundation model that simultaneously achieves native multimodality (sequences, structures, and natural language) and broad entity coverage (molecules, proteins, and their interactions) within a single decoder-only architecture, supporting both single- and multi-entity understanding and generation under one next-token prediction objective.

\noindent $\bullet$ \textbf{A unified tokenization scheme.} We design a tokenization scheme that consolidates molecular sequences (in both SMILES and SELFIES), protein sequences, molecular and protein structures, and natural language into a single shared discrete vocabulary, enabling heterogeneous biomolecular inputs and outputs to be handled uniformly without external encoders or projection adapters.

\noindent $\bullet$ \textbf{Empirical evidence at scale.} We curate a 304.4-billion-token pretraining corpus spanning within-modality and cross-modality data, and instruction-tune across 80 tasks in 6 categories. Built on a single unified pretrained backbone, \method matches or surpasses specialized baselines on 77 out of 80 tasks, providing the first empirical evidence that the two axes above can be unified without compromising per-task performance.

\begin{figure}[t]
    \centering
    \includegraphics[width=1\linewidth]{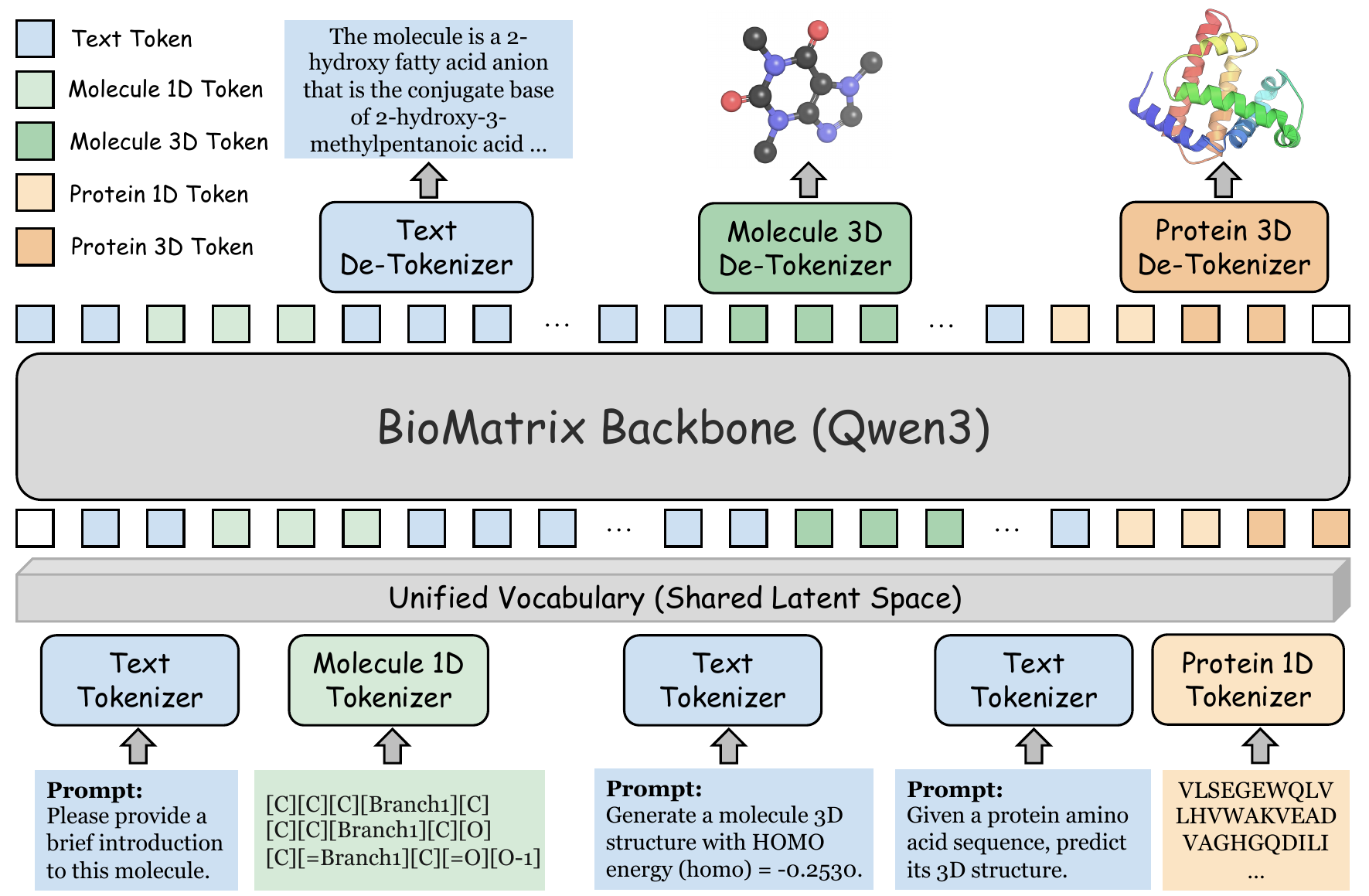}
    \caption{Overview of \method. Natural language, molecular line notations (SMILES, SELFIES), molecular structures, protein sequences, and protein structures are mapped into a shared discrete token space via a unified tokenization scheme, processed by a single Qwen3-based decoder backbone under next-token prediction, and decoded back to their respective modalities through modality-specific de-tokenizers.}
    \label{fig:biomatrix}
\end{figure}

\section{Unified Multimodal Tokenization}
\label{sec:tokenization}

A central design goal of \method is to process sequences, structures, and natural language for both molecules and proteins within a single decoder-only model. To this end, we adopt a unified tokenization scheme that maps all modalities into a shared discrete token space, so that heterogeneous biomolecular inputs can be handled by the same next-token prediction objective.

\begin{figure}[t]
    \centering
    \includegraphics[width=\linewidth]{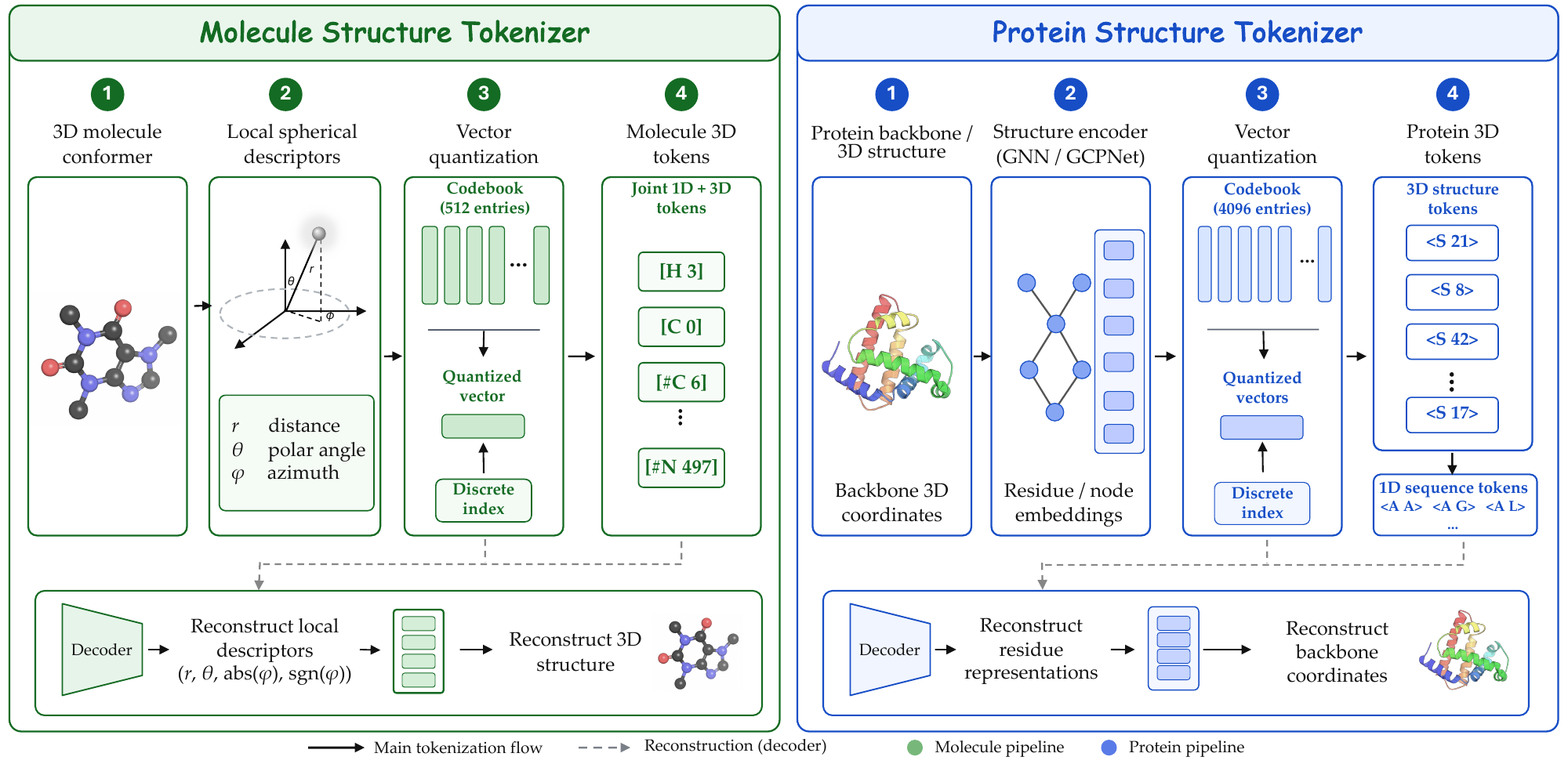}
    \caption{\textbf{3D structure tokenization pipelines in \method.} The molecule structure tokenizer converts each 3D conformer into SELFIES-aligned joint 1D--3D tokens through local spherical descriptors and a 512-entry vector-quantized codebook, while the protein structure tokenizer encodes backbone geometry into per-residue structure tokens using a GCPNet-based encoder and a 4096-entry codebook. Dashed arrows indicate the reconstruction paths used by the corresponding decoders.}
    \label{fig:3d_tokenizers}
\end{figure}

Concretely, natural language and sequence-like representations of molecules (SMILES, SELFIES) and proteins (amino acid sequences) are tokenized as ordinary token sequences, while molecular conformations and protein backbones are discretized into structural codes by two dedicated structure tokenizers: we build on MolStrucTok~\cite{molstructok} for molecules (Section~\ref{sec:mol_struc_tokenizer}, where we additionally propose a branch-decoupled decoder) and directly adopt GCP-VQVAE~\cite{gcp_vqvae} for protein backbones (Section~\ref{sec:pro_struc_tokenizer}). Figure~\ref{fig:3d_tokenizers} summarizes these two structure-tokenization pipelines. These structural codes, together with the sequence and language tokens, are then integrated into a single extended Qwen3 vocabulary alongside modality-specific control tokens that delimit each modality segment (Section~\ref{sec:vocab_ext}). Finally, the embeddings of all newly added tokens are initialized through a description-based scheme (Section~\ref{sec:embed_init}).

\subsection{Molecule Structure Tokenizer}
\label{sec:mol_struc_tokenizer}

For molecular structures, we adopt MolStrucTok~\cite{molstructok} as our structure tokenizer, which discretizes continuous atomic coordinates into a compact set of SE(3)-invariant structural codes that can be seamlessly consumed by a language model. We briefly review the components relevant to \method here, and refer readers to the original paper for full architectural and training details. 
Compared to the original design, we make two modifications tailored to our setting: we drop the \emph{understanding descriptors} (bond lengths and bond angles to nearest neighbors) and retain only the \emph{generation descriptors}, which we find sufficient for faithful structure reconstruction; and we replace the original monolithic decoder with a \emph{branch-decoupled} variant that disentangles the heterogeneous geometric targets and noticeably improves reconstruction fidelity. We describe both changes alongside the corresponding components below.

\paragraph{Geometric input.} MolStrucTok represents a molecular conformer as an augmented line notation in which every atom token is paired with a local geometric descriptor on top of its chemical identity. Following the order induced by the underlying SELFIES sequence, each atom $v_i$ is assigned a local spherical coordinate frame anchored at a focal atom and two reference atoms, all selected via 2D topological proximity to $v_i$. Within this frame, the position of $v_i$ is encoded by three scalars: the radial distance $d_i$ to the focal atom, the polar angle $\theta_i$, and the azimuthal angle $\varphi_i$. Because the frame is constructed purely from intrinsic graph topology and relative geometry, the resulting descriptors are inherently invariant to global translation and rotation. Since $\theta_i \in [0, \pi]$ and $\varphi_i \in (-\pi, \pi]$, to normalize their ranges, $\varphi_i$ is further decomposed into its absolute value and sign, yielding a four-dimensional input
\begin{equation}
\bm{g}_i = (d_i,\, \theta_i,\, \mathrm{abs}(\varphi_i),\, \mathrm{sign}(\varphi_i)).
\end{equation}
As noted above, we omit the \emph{understanding} features (bond lengths and bond angles to the four nearest neighbors) used in the original MolStrucTok and keep only these \emph{generation descriptors} $\bm{g}_i$.

\paragraph{Encoding pipeline.} The continuous descriptors $\bm{g}_i$ are discretized by a VQ-VAE~\cite{vqvae}. After log-normalizing the length component and rescaling the angular components to a unit range, an MLP encoder maps each $\bm{g}_i$ into a latent embedding, which is then assigned to its nearest entry in a learnable codebook of size \textbf{512}. The selected code index serves as the structural token for atom $v_i$, yielding a per-atom sequence of discrete geometric codes whose order follows that of the underlying SELFIES.

\paragraph{Decoding pipeline and branch-decoupled decoder.} At decoding time, each codebook entry is passed through a decoder that reconstructs the four-dimensional descriptor $\bm{g}_i$, from which 3D coordinates are recovered by sequentially traversing the molecule and reconstructing each local spherical frame. The original MolStrucTok decoder regresses all four components with a shared trunk followed by a single output head. We observe, however, that the four entries of $\bm{g}_i$ are heterogeneous in both physical meaning and numerical scale---$d_i$ is a positive length, $\theta_i$ and $\mathrm{abs}(\varphi_i)$ are bounded angles, and $\mathrm{sign}(\varphi_i)$ is a binary indicator---and that forcing a shared trunk to jointly regress all of them introduces interference among these targets and limits reconstruction accuracy. We therefore replace the monolithic decoder with a branch-decoupled variant: the shared trunk is removed, and each of the four targets---length, polar angle, azimuthal magnitude, and azimuthal sign---is predicted by its own dedicated MLP head operating directly on the quantized code. A symmetric modification on the encoder side yielded no measurable benefit, suggesting that the bottleneck lies in disentangling heterogeneous outputs rather than in encoding them. With this change alone, the average per-component RMSD improves by roughly 0.1Å on QM9-2014~\cite{qm9_2014} over the original tokenizer, while leaving the codebook size, encoder, and overall token interface unchanged.

\paragraph{Integration into \method.} For each molecule, we run the MolStrucTok encoder over its conformation to obtain a per-atom sequence of discrete structural codes drawn from the $512$-entry codebook, aligned one-to-one with the corresponding SELFIES atom tokens. Their integration into \method's joint vocabulary is described in Section~\ref{sec:vocab_ext}.

\subsection{Protein Structure Tokenizer}
\label{sec:pro_struc_tokenizer}

For protein structures, we adopt GCP-VQVAE~\cite{gcp_vqvae} as-is, without architectural modification, since it already offers strong reconstruction fidelity and generalization to unseen proteins among publicly available open-source structure tokenizers at comparable codebook sizes. Specifically, we use its \emph{Large} variant throughout \method. Conceptually, GCP-VQVAE plays the same role for proteins that MolStrucTok plays for molecules: it converts a continuous backbone into a sequence of discrete, pose-invariant structural symbols, and provides a decoder that maps those symbols back to backbone coordinates. The key difference is the unit of tokenization. Rather than assigning one code per atom, GCP-VQVAE assigns one code per residue, so a protein chain of length $L$ becomes a length-$L$ sequence of structural tokens aligned with the amino acid sequence.

\paragraph{Geometric input.} GCP-VQVAE operates on protein backbone coordinates, taking the (N, C$_\alpha$, C) atoms of each residue as input. Each residue is treated as a node in a protein graph, with geometric features derived from the relative positions and orientations of neighboring backbone frames rather than from absolute Cartesian coordinates. This is important because a structure token should describe the local fold of a residue, not the arbitrary global rotation or translation of the protein in a coordinate file. By restricting itself to backbone geometry, the tokenizer captures the global fold and local secondary-structure arrangement of a protein while remaining agnostic to side-chain conformations, which keeps the discrete vocabulary compact and broadly applicable to both experimentally resolved and predicted structures.

\paragraph{Encoding pipeline.} The backbone graph is first processed by a strictly SE(3)-equivariant GCPNet encoder~\cite{gcpnet}. GCPNet maintains scalar features together with vector features, and performs message passing in local geometric frames so that the representation can retain directional and chiral information while remaining well behaved under global rigid motions. The resulting per-residue embeddings summarize each residue's local backbone environment as well as information propagated from nearby residues. A transformer encoder then refines these residue embeddings with global sequence-level context. Finally, a vector-quantization layer replaces each continuous residue embedding with the nearest entry in a learnable codebook of size \num{4096}. The selected code index is the protein structure token for that residue. Thus the encoder maps
\begin{equation}
    \{(\mathbf{x}^{N}_i,\mathbf{x}^{C_\alpha}_i,\mathbf{x}^{C}_i)\}_{i=1}^{L}
    \;\longrightarrow\;
    (z_1,\ldots,z_L), \qquad z_i \in \{1,\ldots,\num{4096}\},
\end{equation}
where each $z_i$ is a discrete summary of the backbone geometry around residue $i$.

\paragraph{Decoding pipeline.} The decoder performs the inverse operation. Each discrete code $z_i$ is first mapped back to its corresponding codebook vector, producing a sequence of continuous embeddings with the same length as the protein chain. A transformer decoder then predicts a rigid frame for each residue, parameterized by a translation and a rotation represented with a continuous 6D rotation head. These predicted residue frames are applied to an idealized local backbone template containing the N, C$_\alpha$, and C atoms, yielding reconstructed backbone coordinates for the full chain. In this sense, the code sequence does not directly store Cartesian coordinates; it stores enough local geometric information for the decoder to recover a globally coherent backbone after the residue-level frames are assembled. Training uses reconstruction losses on the recovered coordinates and geometry together with the standard VQ codebook and commitment objectives, so the learned codes form a compact geometric vocabulary that can be generated autoregressively by \method.

\paragraph{Integration into \method.} For each protein, we run the GCP-VQVAE Large encoder over its backbone structure to obtain a length-$L$ sequence of discrete per-residue structural tokens drawn from the \num{4096}-entry codebook, one token per residue. Because this sequence is residue-aligned, it can be paired directly with the dedicated amino-acid tokens introduced in Section~\ref{sec:vocab_ext}: folding becomes generation of $z_1,\ldots,z_L$ from amino-acid tokens, while inverse folding becomes generation of amino-acid tokens from $z_1,\ldots,z_L$. At output time, generated structure-token sequences are passed through the GCP-VQVAE decoder to recover backbone coordinates. Their integration into \method's joint vocabulary is described in Section~\ref{sec:vocab_ext}.

\subsection{Vocabulary Extension}
\label{sec:vocab_ext}

\definecolor{tokblue}{RGB}{210,230,250}

\setlength{\fboxsep}{2pt}
\newcommand{\tok}[1]{%
  \tikz[baseline=(X.base)]\node[fill=tokblue,inner sep=2pt,rounded corners=1pt]
    (X) {\texttt{#1}};\hspace{2pt}}

\begin{table}[h]
\centering
\small
\renewcommand{\arraystretch}{1.5}
\setlength{\tabcolsep}{4pt}
\setlength{\fboxsep}{2pt}
\resizebox{\textwidth}{!}{
\begin{tabular}{lll}
\toprule
Category & Modality & Example \\
\midrule
\multirow{3}{*}{Molecule}
  & 1D SMILES    & \texttt{<|mol\_smi\_start|>}\,\tok{C}\tok{\#}\tok{CC}\tok{\#}\tok{N}\texttt{<|mol\_smi\_end|>} \\
  & 1D SELFIES   & \texttt{<|mol\_sfi\_start|>}\,\tok{[C]}\tok{[\#C]}\tok{[C]}\tok{[\#N]}\texttt{<|mol\_sfi\_end|>} \\
  & 3D Structure & \texttt{<|mol\_3d\_start|>}\,\tok{[H 3]}\tok{[C 0]}\tok{[\#C 6]}\tok{[C 132]}\tok{[\#N 497]}\texttt{<|mol\_3d\_end|>} \\
\midrule
\multirow{2}{*}{Protein}
  & 1D AA Sequence & \texttt{<|prot\_aa\_start|>}\,\tok{<A M>}\tok{<A R>}\tok{<A A>}\tok{<A K>}\tok{<A W>}\texttt{<|prot\_aa\_end|>} \\
  & 3D Structure  & \texttt{<|prot\_3d\_start|>}\,\tok{<S 4012>}\tok{<S 153>}\tok{<S 2091>}\tok{<S 2440>}\tok{<S 3117>}\texttt{<|prot\_3d\_end|>} \\
\bottomrule
\end{tabular}
}
\caption{Examples of wrapped sequences for each modality in the extended vocabulary. Each molecule or protein segment is delimited by a dedicated pair of start/end special tokens, while natural language text is left unwrapped and serves as the default carrier modality. Each \colorbox{tokblue}{\texttt{light-blue box}} corresponds to a single token produced by the tokenizer on the wrapped segment. Molecular 3D tokens are formed by composing each SELFIES token with its 3D code from MolStrucTok~\cite{molstructok}; protein 3D tokens form a standalone set produced by GCP-VQVAE~\cite{gcp_vqvae}.}
\label{tab:tokenize_example}
\end{table}

To accommodate the full spectrum of biomolecular modalities within a single decoder-only model, we extend the original Qwen3 vocabulary along two orthogonal axes: \emph{modality-specific tokens} that encode molecular and protein sequence and structure representations, and \emph{structural control tokens} that delimit each modality segment within a mixed-modality sequence. 
Table~\ref{tab:tokenize_example} summarizes the full set of additions, which we describe below.

\paragraph{1D tokens.} 
For natural language, we directly reuse the Qwen3 tokenizer without modification. 
For molecules, SMILES strings are tokenized directly by the Qwen3 tokenizer since SMILES is a widely used line notation that Qwen3 has already been exposed to during pretraining. SELFIES, in contrast, is a more recent and less prevalent notation whose bracketed atomic symbols are not natively recognized, and we add them as new vocabulary entries.
We retain both SMILES and SELFIES as parallel molecular line notations rather than committing to a single one since our downstream experiments (Section~\ref{sec:eval_mol_1d}) reveal that the two notations exhibit complementary strengths.
For proteins, we introduce a dedicated vocabulary of 26 residue-level tokens covering the 20 standard amino acids together with non-standard and unknown residue symbols. Although amino acid letters already lie within Qwen3's character set, its tokenizer merges them into variable-length subword pieces, which would break the one-to-one correspondence between residues and the per-residue 3D tokens produced by GCP-VQVAE (Section~\ref{sec:pro_struc_tokenizer}).
Enforcing single-token-per-residue encoding preserves this alignment and directly supports downstream tasks that operate on residue-level sequence--structure correspondences, such as folding (mapping each AA token to its structural token) and inverse folding (the reverse direction).

\paragraph{3D tokens.} The molecular and protein structure vocabularies are inherited from the two structure tokenizers introduced in Sections~\ref{sec:mol_struc_tokenizer} and~\ref{sec:pro_struc_tokenizer}, but we integrate them into \method's vocabulary in two different ways, guided by how many joint sequence--structure entries each entity type would actually need.

For molecules, we compose each MolStrucTok code with its aligned SELFIES atom token into a joint entry, following the original MolStrucTok design, so that every structure-tokenized molecule token simultaneously carries chemical identity and local geometry. In principle, pairing the 925 unique SELFIES atom tokens observed in our pretraining corpus with the 512 MolStrucTok codes could yield up to $\sim$470K combinations, but in practice the vast majority never occur. We therefore enumerate joint tokens from the pretraining data in descending order of frequency and greedily retain the smallest prefix that covers more than $99\%$ of occurrences, which yields \num{11294} joint entries. This composed vocabulary is compact enough to be learned reliably while keeping tail combinations, whose embeddings would be under-trained, out of the vocabulary.

For proteins, we keep the \num{4096} GCP-VQVAE codes as a standalone set of per-residue structure tokens, parallel to the 26 residue-level amino-acid tokens (standard amino acids plus non-standard and unknown symbols). Applying the same $99\%$-coverage procedure to AA--structure pairs would require \num{71737} joint entries (nearly half the size of the original Qwen3 vocabulary, \num{151669}), and a long tail of these combinations occurs only a handful of times in the pretraining corpus, leaving their embeddings severely under-trained. We therefore opt for a factored representation in which each residue is described by an AA token and a structure token independently.

\paragraph{Structural control tokens.} To let the model distinguish heterogeneous content within a single token stream, we further introduce five pairs of start/end special tokens that mark the boundaries of each non-text biomolecular segment (rightmost column of Table~\ref{tab:tokenize_example}). Natural language is left unwrapped and serves as the default carrier modality, while molecular and protein segments, whether sequence- or structure-based, are explicitly enclosed by their corresponding control-token pair. This design provides a clean interface between continuous prose and structured biomolecular content, and allows text, molecular representations, and protein representations to be interleaved within one training instance under \method's unified next-token prediction framework.

\subsection{Embedding Initialization}
\label{sec:embed_init}

Extending Qwen3's vocabulary with the modality-specific tokens and structural control tokens introduced in Section~\ref{sec:vocab_ext} adds several thousand new entries to both the input embedding table and, when not tied, the output projection. Initializing these new rows from scratch (e.g., Gaussian noise) discards the well-trained semantic geometry of the underlying language model and slows down the early stages of continual pretraining. We therefore adopt a simple description-based initialization scheme that grounds every newly added token in the pretrained embedding space of Qwen3 itself.

\paragraph{Description-based initialization.} For each new token $t$, we first construct a short natural-language description $\mathrm{desc}(t)$ that summarizes its semantic role. The description is then tokenized by the original Qwen3 tokenizer into a sequence of subword ids $(s_1, \dots, s_k)$, and the new embedding is initialized as the mean of the corresponding pretrained input embeddings, plus a small isotropic Gaussian perturbation:
\begin{equation}
\bm{e}_t \;=\; \frac{1}{k}\sum_{j=1}^{k} \bm{E}_{\text{Qwen3}}[s_j] \;+\; \bm{\epsilon}, \qquad \bm{\epsilon} \sim \mathcal{N}\!\left(\bm{0},\, \tfrac{1}{d}\bm{I}\right),
\end{equation}
where $\bm{E}_{\text{Qwen3}}$ denotes the pretrained Qwen3 input embedding matrix and $d$ is the embedding dimension. The noise term breaks symmetry between tokens whose descriptions happen to share subword ids, while keeping each new embedding within the typical scale of the pretrained vocabulary. 

\paragraph{Per-category descriptions.} The description $\mathrm{desc}(t)$ is constructed in a category-specific manner that reflects how much prior semantic content each token actually carries:
\begin{itemize}
    \item \textbf{Structural control tokens} are mapped to short phrases describing the modality boundary they mark, e.g., \texttt{<|mol\_smi\_start|>} $\rightarrow$ \emph{Begins a molecule SMILES string}, \texttt{<|prot\_3d\_end|>} $\rightarrow$ \emph{Ends a protein 3D structure}. This grounds each delimiter in the same semantic space as the surrounding natural-language context.
    \item \textbf{Protein 1D tokens} are mapped to the full English name of their corresponding amino acid (e.g., \texttt{<A A>} $\rightarrow$ \emph{Alanine}, \texttt{<A W>} $\rightarrow$ \emph{Tryptophan}), with non-standard or ambiguous symbols mapped to descriptive phrases such as \emph{Selenocysteine} or \emph{any amino acid}.
    \item \textbf{Molecular 1D, molecular 3D, and protein 3D tokens} have no informative natural-language gloss—a SELFIES atom token or a VQ codebook index is essentially an opaque symbol. For these categories, we use the raw token string as its own description, which yields a near-uniform initialization centered on the subword decomposition of the token literal while still benefiting from the noise term to break ties.
\end{itemize}

\paragraph{Discussion.} This scheme requires no additional training or auxiliary models: it reuses the frozen Qwen3 embedding table as a fixed semantic encoder for the description strings, and writes the resulting vectors directly into the expanded embedding rows before continual pretraining begins. In our preliminary experiments, we find that this description-based initialization provides a more stable starting point than random initialization in the early phase of continual pretraining.

\section{Multimodal Continual Pretraining}
\label{sec:pretrain}

To equip \method with joint understanding over sequences, structures, and language for both molecules and proteins, we perform continual pretraining on a large-scale, heterogeneous corpus that spans general and scientific text, molecular and protein data with both sequence and structure views, and cross-entity resources linking distinct biomolecular entities or wrapping them into free-form scientific text. All modalities are serialized into a shared token space via the unified tokenization scheme described in Section~\ref{sec:tokenization}, with each modality segment explicitly delimited by the modality-specific start/end tokens. In this section, we first describe the composition and curation of the pretraining corpus (Section~\ref{sec:cpt_data}), and then report the training dynamics of \method under this data mixture (Section~\ref{sec:cpt_dynamics}).

\subsection{Pretraining Corpus}
\label{sec:cpt_data}

We organize the corpus into four categories: (i) general- and scientific-domain text (Section~\ref{sec:text_data}); (ii) molecule-centric data, where within-molecule sequence, structure, and textual views are jointly exposed (Section~\ref{sec:mol_data}); (iii) protein-centric data, organized analogously to the molecular case (Section~\ref{sec:prot_data}); and (iv) cross-entity and interleaved data that link distinct biomolecular entities or wrap them into scientific prose (Section~\ref{sec:cross_data}).

For molecular and protein data, we adopt two complementary strategies that operate at different levels of instance construction: \emph{stochastic composition} governs how fields are sampled and ordered within a single instance, while the \emph{three-view instance pattern} controls how multiple instances are emitted from the same biomolecular entity.
\emph{Stochastic composition} applies to every molecule- and protein-centric instance. For each instance, the inclusion of optional fields is controlled by predefined probabilities, and the ordering of interchangeable components (e.g., property fields, annotation entries, or co-present modalities) is randomly permuted across samples. For example, a single PubChem molecule may yield an instance that randomly samples a subset of {SMILES, SELFIES, tokenized conformation, IUPAC name, textual description, physicochemical properties} and permutes their order. This design prevents the model from overfitting to any fixed input schema and encourages it to jointly learn associations among heterogeneous representations~\cite{moddrop}.
\emph{The three-view instance pattern} is applied specifically to protein-centric data sources that provide both sequences and structures. For each such protein, we emit (i) a sequence-only instance containing the amino acid sequence, (ii) a structure-only instance containing the tokenized backbone, and (iii) a joint instance containing both modalities with their relative order randomly permuted, exposing the model to both isolated and jointly grounded views of each protein. Protein data sources providing only sequences emit the sequence view alone; sources that additionally carry functional annotations further emit a separate composite instance that combines those annotations with the available modalities under the stochastic composition strategy. We apply this pattern to proteins because sequence and structure provide largely complementary, residue-level-aligned views of the same entity, and we want the model to handle each modality both independently and jointly.

\begin{figure}[t]
\centering
\vspace{-0.6cm}
\includegraphics[width=\linewidth]{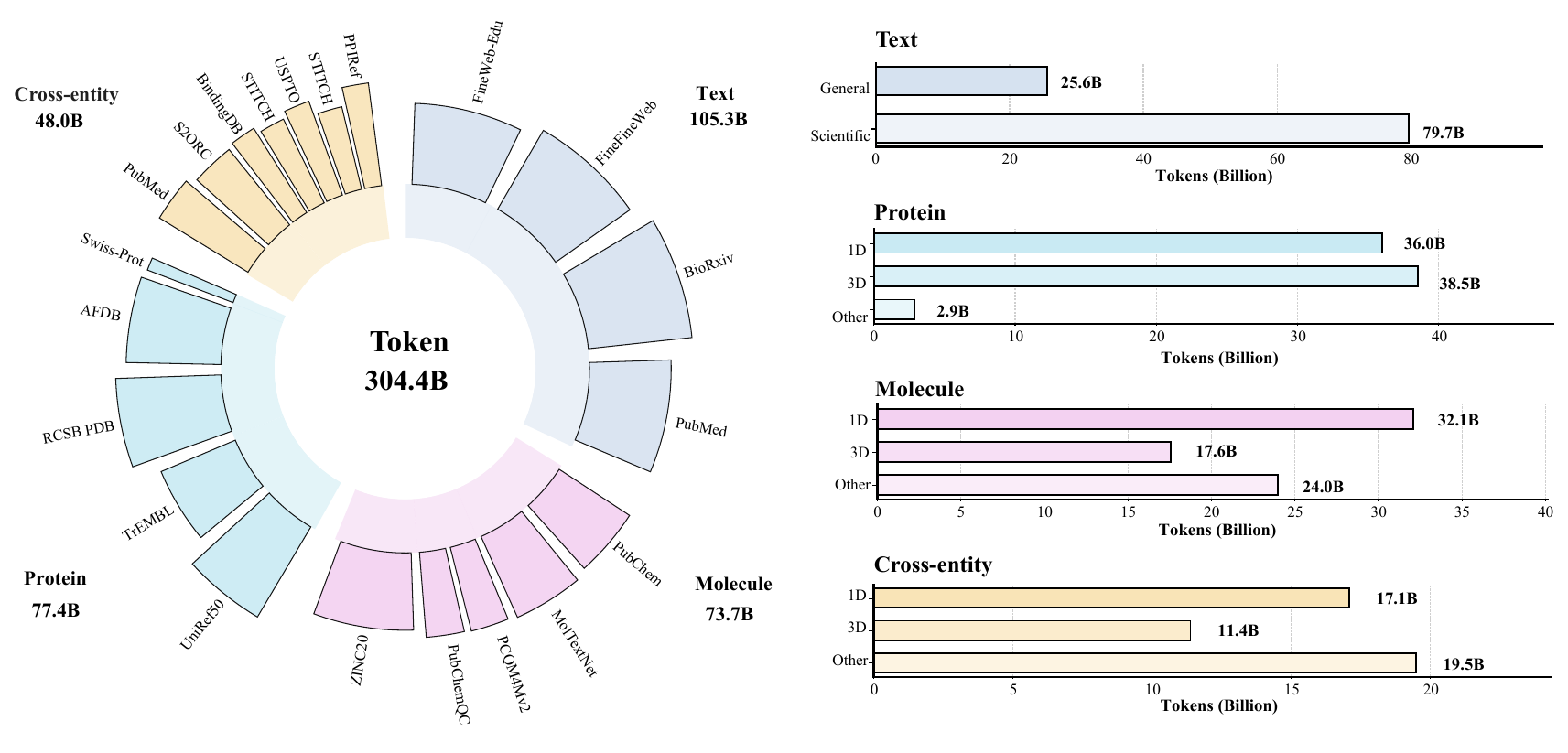}
\vspace{-0.6cm}
\caption{Overview of the continual pretraining corpus composition. Token budget distribution across four data categories — general and scientific text, molecule-centric data, protein-centric data, and cross-entity/interleaved data — spanning sequence, structure, and textual modalities, totaling 304.4 billion tokens.}
\label{fig:pretrain_data}
\end{figure}

In total, our pretraining corpus comprises approximately \textbf{304.4 billion} tokens, with text and protein data contributing the largest shares, followed by molecular and cross-entity data. Figure~\ref{fig:pretrain_data} summarizes the full token budget across categories and modalities.

\subsubsection{Text Corpora}
\label{sec:text_data}

\paragraph{General-domain text.} We use FineWeb-Edu~\cite{fineweb-edu}, a large-scale, education-oriented filtered subset of CommonCrawl, as our source of general-domain natural language, which helps prevent catastrophic forgetting of the model's general language abilities during continual pretraining.

\paragraph{Scientific-domain text.} To strengthen the model's scientific language proficiency, we curate text from two complementary sources. From FineFineWeb~\cite{finefineweb}, a domain-categorized web corpus, we select the four subsets most relevant to our target disciplines: biology, chemistry, medicine, and health. To complement web-sourced text with more rigorous scientific writing, we additionally extract a large collection of full-article texts from PubMed~\cite{pubmed}, covering peer-reviewed literature in molecular biology, pharmacology, and clinical research.

\subsubsection{Molecular Data}
\label{sec:mol_data}

\paragraph{PubChem.} We collect all compounds available in PubChem~\cite{pubchem2025} as of December 2025 and construct multi-view training instances that jointly expose the model to structure, properties, and annotations. For each molecule, we extract up to three representations---canonical SMILES, SELFIES, and a tokenized conformation produced by our molecular structure tokenizer (Section~\ref{sec:mol_struc_tokenizer})---together with available physicochemical properties (e.g., molecular weight, hydrogen bond donor/acceptor counts, topological polar surface area, XLogP3) and textual annotations such as IUPAC names and natural language descriptions. These heterogeneous components are concatenated into a single training instance following the stochastic composition strategy described above. Since PubChem's conformer coverage is limited and its conformations are generated by lightweight force fields, we further incorporate DFT-level conformations from PCQM4Mv2~\cite{ogb_lsc} and PubChemQC~\cite{pubchemqc}, processed through the same structure tokenizer and merged into the corpus under the same composition strategy.

\paragraph{MolTextNet.} To strengthen molecule--text alignment, we incorporate MolTextNet~\cite{moltextnet}, a corpus of approximately 2.5M molecule--description pairs derived from ChEMBL35~\cite{chembl_2023}, whose descriptions---synthesized by GPT-4o-mini from structural, physicochemical, bioactivity, and synthesizability annotations---are substantially longer and more informative than prior molecule--text resources. For each entry, we canonicalize the SMILES, derive the corresponding SELFIES, and assemble the three modalities into a single training instance under the same stochastic composition strategy.

\subsubsection{Protein Data}
\label{sec:prot_data}

\paragraph{UniRef50.} We include protein sequences from UniRef50~\cite{uniref} to provide broad coverage of the protein sequence space.

\paragraph{RCSB PDB.} We incorporate experimentally determined protein structures from the RCSB Protein Data Bank~\cite{rcsb_pdb}. For each chain, we extract the amino acid sequence and discretize its backbone geometry into structural tokens via our protein structure tokenizer (Section~\ref{sec:pro_struc_tokenizer}), then apply the three-view instance pattern.

\paragraph{UniProt/Swiss-Prot.} We use the manually curated Swiss-Prot subset of UniProt~\cite{uniprot_2023} as our primary source of annotated proteins, pairing each entry with its predicted structure from the latest release of the AlphaFold Protein Structure Database (AFDB)~\cite{afdb_2026} when available. Beyond sequence and structure, Swiss-Prot provides rich functional metadata: protein name, description, organism, taxonomic lineage, subcellular localization, keywords, and free-text comments on function, catalytic activity, and biological context. For each protein, we serialize these annotations together with the available sequence and structure views into a composite training instance, and additionally emit the three-view instance pattern so that the model is exposed to both isolated and jointly grounded views.

\paragraph{UniProt/TrEMBL.} To scale up coverage of the protein sequence--function space, we further include the automatically annotated TrEMBL subset of UniProt~\cite{uniprot_2023}. As most TrEMBL entries lack resolved or predicted structures at our processing time, we retain only the amino acid sequence together with its functional annotations, with evidence codes and literature citations removed. TrEMBL provides broad taxonomic and functional coverage that complements the smaller but more richly grounded Swiss-Prot samples.

\paragraph{AlphaFold Protein Structure Database.} Beyond the Swiss-Prot predictions already paired with UniProt annotations, we incorporate two additional curated subsets from the AFDB~\cite{afdb_2026}: the model organism proteomes (covering 16 reference species including human, mouse, \textit{E. coli}, and \textit{S. cerevisiae}) and the global health proteomes (covering pathogens relevant to neglected tropical and infectious diseases). 
To further scale structural diversity, we additionally sample approximately 130M predicted structures from the full AFDB, representing over 60\% of all entries in the database and covering a substantial fraction of the known protein universe.
All entries are processed through the same structure tokenizer and emit the three-view instance pattern.

\subsubsection{Cross-Entity and Interleaved Data}
\label{sec:cross_data}

While Sections~\ref{sec:mol_data} and~\ref{sec:prot_data} already expose the model to within-entity cross-modal alignment (e.g., a molecule paired with its own conformation and description), this section focuses on data that link distinct biomolecular entities to each other or wrap them into free-form scientific text.

\paragraph{Interleaved biomedical text.} To encourage tight cross-modal grounding between natural language and biomolecular representations, we construct an interleaved corpus in which biomedical entities mentioned in scientific text are directly paired with their structural representations. Source documents are drawn from four complementary corpora---PubMed abstracts~\cite{pubmed}, bioRxiv abstracts~\cite{biorxiv}, S2ORC~\cite{s2orc}, and the USPTO-Applications patent corpus~\cite{presto,uspto}---jointly covering biomedical literature, preprints, broad scientific publications, and chemical patents. We apply BERN2~\cite{bern2}, a neural biomedical named entity recognition and normalization tool, to detect mentions of molecules and proteins and link them to canonical database identifiers. For each recognized entity, we retrieve the corresponding structural representations---SMILES and SELFIES for molecules, amino acid sequences for proteins---and append them inline after the entity mention, wrapped with the appropriate special tokens. This exposes the model to natural language contexts in which textual descriptions co-occur with their underlying structural representations, facilitating fine-grained alignment between biomedical terminology and biomolecular entities.

\paragraph{Biomolecular interaction data.} To equip the model with knowledge of biomolecular interactions, we incorporate curated interaction datasets spanning both molecule--protein and protein--protein pairs in sequence-only and structure-grounded forms. For molecule--protein interactions, we collect sequence-level pairs from BindingDB~\cite{bindingdb_2024}, STITCH~\cite{stitch}, and the jglaser binding affinity dataset~\cite{jglaser}, and structure-grounded complexes from CrossDocked2020~\cite{crossdocked2020}. For protein--protein interactions, we use AlphaSeq~\cite{alphaseq} as the sequence-level source and PPIRef~\cite{ppiref} as the structure-grounded source. 

\subsection{Training Dynamics}
\label{sec:cpt_dynamics}
\begin{figure}[t]
    \centering
    \includegraphics[width=0.85\linewidth]{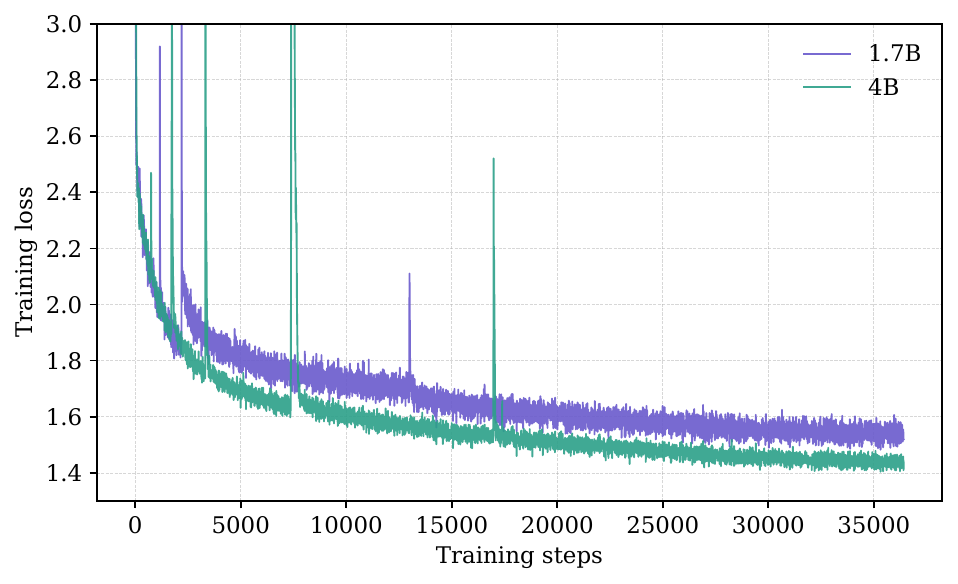}
    \caption{Training loss curves of \method-1.7B and \method-4B during multimodal continual pretraining. Both models converge stably despite occasional transient loss spikes, which leave downstream performance essentially unaffected.}
    \label{fig:cpt_loss}
\end{figure}

We conduct continual pretraining on top of the LLaMA-Factory~\cite{llamafactory} framework.
We train both \method-1.7B and \method-4B on 64 NVIDIA H100 GPUs with a global batch size of \num{1024} and a maximum sequence length of \num{8192} tokens, using the AdamW optimizer with a peak learning rate of \num{2.0e-4} under a cosine schedule with \num{2000} warmup steps. Training runs for approximately 36.4K steps (1 epoch), consuming the full 304.4B-token corpus described in Section~\ref{sec:cpt_data}.
The resulting training loss curves are shown in Figure~\ref{fig:cpt_loss}.

\paragraph{On loss spikes.} A closer inspection of the curves reveals several transient loss spikes for both model sizes, concentrated primarily in the early-to-middle phase of training. In all cases, the loss recovers to its pre-spike trajectory within a small number of steps without any manual intervention (e.g., learning-rate rewinding or optimizer-state rollback). To assess whether these spikes cause any latent degradation that is not visible in the training loss itself, we evaluated checkpoints before and after each prominent spike on a held-out validation set. Across all evaluated spikes, we observe no meaningful degradation: pre- and post-spike checkpoints achieve essentially comparable validation loss, and subsequent checkpoints continue to improve smoothly as training proceeds.

Given that the diversity of modalities inherently increases the risk of optimization instability, we interpret the observed spikes as localized optimization perturbations (most likely triggered by occasional high-gradient batches arising from the highly heterogeneous mixture of text, sequence tokens, and structure tokens) rather than genuine deterioration of the learned representations.

\section{Instruction Tuning across Biological Tasks}
\label{sec:sft}

After multimodal continual pretraining, we perform supervised instruction tuning to align \method with a broad spectrum of downstream biological tasks and to elicit its ability to follow natural-language instructions over heterogeneous biomolecular inputs and outputs. In this section, we first describe the task taxonomy and data sources of the instruction-tuning corpus (Section~\ref{sec:sft_data}), then detail how heterogeneous datasets are unified into a consistent instruction format with diversified prompts (Section~\ref{sec:sft_format}), and finally report the training configuration (Section~\ref{sec:sft_train}).

\begin{table}[t]
\centering
\caption{Overview of downstream tasks used for instruction tuning. The table defines 6 task categories spanning three domains (molecule, protein, interaction) across 1D and 3D modalities. The final entry (MegaScience, Text domain) is included only during instruction tuning to preserve general scientific reasoning capability, and is not part of the evaluation suite.} 
\label{tab:sft_tasks}
\resizebox{\textwidth}{!}{
\begin{tabular}{lc cl}
\toprule
\textbf{Domain} & \textbf{Modality} & \textbf{Task} & \textbf{Datasets} \\
\midrule
\multirow{11}{*}{Molecule}
 & \multirow{10}{*}{1D}
 & Unconditional Sequence Generation & MOSES~\cite{moses}, GuacaMol~\cite{guacamol} \\
 & & Name Conversion             & SMolInstruct~\cite{llasmol} \\
 & & Property Prediction         & SMolInstruct~\cite{llasmol} \\
 & & Molecule Captioning        & SMolInstruct~\cite{llasmol}, CheBI-20~\cite{molt5}, KnowMol-100K~\cite{knowmol} \\
 & & Text-Based Molecule Generation   & SMolInstruct~\cite{llasmol}, KnowMol-100K~\cite{knowmol}  \\
 & & Forward Synthesis        & SMolInstruct~\cite{llasmol} \\
 & & Retrosynthesis        & SMolInstruct~\cite{llasmol}, USPTO-50K~\cite{uspto_50k} \\
 & & Molecule Editing   & OpenMolIns/S\textsuperscript{2}-Bench~\cite{s2bench} \\
 & & Molecule Optimization &  OpenMolIns/S\textsuperscript{2}-Bench~\cite{s2bench}, MolOpt-Instructions~\cite{drugassist} \\ 
 & & Customized Molecule Generation & OpenMolIns/S\textsuperscript{2}-Bench~\cite{s2bench} \\
 & & Molecule Question Answering  & MoleculeQA~\cite{moleculeqa} \\
\cmidrule(lr){2-4}
 & \multirow{2}{*}{3D}
 & Unconditional Structure Generation & QM9-2014~\cite{qm9_2014} \\
 & & Conditional Structure Generation  & QM9-2014~\cite{qm9_2014} \\
\midrule
\multirow{10}{*}{Protein}
 & \multirow{5}{*}{1D}
 & Sequence Understanding  & Open Protein Instructions~\cite{opi} \\
 & & Annotation Prediction  & Open Protein Instructions~\cite{opi}, Mol-Instructions~\cite{mol_instructions} \\
 & & Knowledge Mining      & Open Protein Instructions~\cite{opi} \\
 & & Text-Based Protein Design         & CAMEO~\cite{cameo}, Molinst-SwissProtCLAP~\cite{mol_instructions} \\
  & & Unconditional Sequence Generation & DPLM-2~\cite{dplm2} \\
\cmidrule(lr){2-4}
 & \multirow{5}{*}{3D}
 & Structure Understanding        & PFUD~\cite{prottex} \\
 & & Structure Prediction (Folding)   & DPLM-2~\cite{dplm2}/PDB Date~\cite{dplm2} \\
 & & Inverse Folding                  & DPLM-2~\cite{dplm2}/PDB Date~\cite{dplm2} \\
 & & Sequence--Structure Co-Generation & DPLM-2~\cite{dplm2}  \\
 & & Unconditional Backbone Generation & DPLM-2~\cite{dplm2} \\
\midrule
\multirow{3}{*}{Interaction}
 & \multirow{2}{*}{1D}
 & Molecule--Protein Interaction    &  BindingDB~\cite{bindingdb}, PDBBindv2019~\cite{pdbbind_2019}, CASF-2016~\cite{casf_2016} \\
 & & Protein--Protein Interaction     &  Yeast~\cite{yeast_ppi}, Human~\cite{human_ppi}, PPI Affinity~\cite{ppi_affinity} \\
 \cmidrule(lr){2-4}
 & 3D & Molecule--Protein Interaction    &  PDBBindv2020~\cite{pdbbind} \\
\midrule
\bound{Text} & \bound{--} & \bound{--} & \bound{MegaScience~\cite{megascience}} \\
\bottomrule
\end{tabular}
}
\end{table}

\subsection{Task Taxonomy and Data Sources}
\label{sec:sft_data}

We curate a comprehensive instruction-tuning corpus that spans three biological entity scopes---molecule, protein, and interaction---and, within each scope, covers both sequence-based and structure-based tasks.
Table~\ref{tab:sft_tasks} summarizes the full task taxonomy, comprising 80 tasks drawn from a diverse collection of public benchmarks; this taxonomy is used consistently throughout the paper, with the paragraphs below giving a scope-level overview and Section~\ref{sec:eval_mol},~\ref{sec:eval_prot}, and~\ref{sec:eval_inter} reporting evaluation results organized along the same sub-task structure.
Table~\ref{tab:sft_data_stat_overall} reports the overall training data statistics grouped by domain and modality, and a detailed per-sub-task breakdown is further provided in Table~\ref{tab:sft_data_stat_detail}.

\paragraph{Molecular tasks.} 
For molecules, we cover a broad range of sequence-level tasks, including unconditional generation, name conversion, property prediction, captioning, text-based generation, forward and retrosynthesis, editing, customized generation, optimization, and question answering. 
For molecular structures, we include unconditional and property-conditioned conformer generation tasks.

\paragraph{Protein tasks.} 
For proteins, we cover sequence understanding, annotation prediction, knowledge mining, text-based protein design, and unconditional sequence generation. 
For protein structures, we include structure understanding, folding and inverse folding, sequence--structure co-generation, and unconditional structure generation.

\paragraph{Interaction tasks.} Beyond single-entity tasks, we additionally include molecule--protein and protein--protein interaction tasks in both sequence-only and structure-grounded forms, covering binding affinity prediction and interaction modeling at both levels.

\paragraph{General scientific reasoning.} Beyond the biology-specific tasks enumerated in Table~\ref{tab:sft_tasks}, we additionally incorporate MegaScience~\cite{megascience} into the instruction-tuning mixture as a source of general scientific reasoning supervision. Including this corpus helps preserve \method's broad scientific reasoning and instruction-following capabilities inherited from the base Qwen3 model, and prevents them from being eroded by the heavy concentration of domain-specific biomolecular tasks.

\subsection{Data Format Unification and Prompt Diversification}
\label{sec:sft_format}

The collected datasets exhibit substantial heterogeneity in their original formats, and only a fraction is natively distributed in instruction-style format. To bring them into a uniform interface compatible with \method, we apply three transformations as follows.

\paragraph{Modality serialization with special tokens.} Whenever an input or output field contains a biomolecular entity, we serialize it using the unified tokenization scheme and wrap it with the corresponding modality-specific start/end tokens, following the same protocol as in continual pretraining (Table~\ref{tab:tokenize_example}). 

\paragraph{Conversion to instruction format.} For datasets that are not natively in instruction form, we manually design conversion pipelines that map each raw example into a \emph{(instruction, input, response)} triplet. Depending on the task type, classification labels are verbalized into natural-language category names, regression targets are formatted as numerical strings with task-appropriate units and precision.

\paragraph{Prompt template diversification.} Using a single fixed prompt per task tends to produce models that overfit to the surface form of the instruction and generalize poorly to paraphrased or unseen phrasings at inference time. To mitigate this, for each sub-task in Table~\ref{tab:sft_tasks} we manually curate a pool of semantically equivalent prompt templates that vary in wording, sentence structure, and level of formality, while preserving the underlying task intent. Each training instance then samples one template uniformly at random from its task-specific pool to fill the instruction field.

\begin{table}[t]
\centering
\caption{Training data statistics of instruction tuning. 
}
\label{tab:sft_data_stat_overall}
\begin{tabular}{lcc}
\toprule
Domain & Modality & \#Train \\
\midrule
\multirow{2}{*}{Molecule}
 & 1D (SMILES)  & \num{9407182} \\
 & 1D (SELFIES) & \num{9425897} \\
 & 3D           & \num{390928} \\
\midrule
\multirow{2}{*}{Protein}
 & 1D & \num{3009742} \\
 & 3D & \num{1272911} \\
\midrule
\multirow{2}{*}{Interaction}
 & 1D & \num{67369} \\
 & 3D & \num{18013} \\
\midrule
Text & -- & \num{1253230} \\
\midrule
Total & -- & \num{24845248} \\
\bottomrule
\end{tabular}
\end{table}

\subsection{Training Configuration}
\label{sec:sft_train}
Instruction tuning is also performed using the LLaMA-Factory framework~\cite{llamafactory}.
We instruction-tune both \method-1.7B and \method-4B using the AdamW optimizer. The peak learning rate is set to \num{5.0e-5} under a cosine schedule with a linear warmup over the first 10\% of total steps. 
We use sample-level attention masking during training.
The maximum sequence length is \num{2048} tokens.

\section{Evaluation Setup}
\label{sec:eval_setup}

Our evaluation is designed to answer the two empirical questions:

\textbf{Q1: Feasibility of unified tokenization.} Can biomolecular structures be discretized into tokens that share a vocabulary with sequences and natural language, and decoded under a single next-token objective, while still supporting high-quality structural generation, prediction, and understanding across both molecules and proteins?

\textbf{Q2: Competitiveness against task-specialized models.} Does a single unified backbone---covering molecules, proteins, and their interactions across sequence, structure, and text within one vocabulary---match or surpass models that are purpose-built or purpose-trained for individual tasks?

To answer these questions, we organize the evaluation along the same taxonomy used for instruction tuning (Table~\ref{tab:sft_tasks}): single-entity molecular tasks across sequence and structure settings (Section~\ref{sec:eval_mol}), single-entity protein tasks organized analogously (Section~\ref{sec:eval_prot}), and molecule--protein as well as protein--protein interaction tasks (Section~\ref{sec:eval_inter}).

We emphasize that \method shares a single tokenization scheme, a single continually-pretrained backbone, and a single next-token training objective across every result reported in this section. 
The instruction-tuning stage, in contrast, is organized into several task-group-specific variants whose grouping is determined by data composition and per-task evaluation considerations. We make this division fully explicit in Table~\ref{tab:eval_variants}.

\paragraph{Evaluated variants.} All variants of \method evaluated in this section are instruction-tuned from the same continually-pretrained backbones and use the same unified tokenization scheme. The instruction-tuning stage, however, is organized into task-group-specific variants summarized in Table~\ref{tab:eval_variants}, and the numbers reported correspond to these task-group variants. 
For practical convenience, we release a single \emph{all-tasks merged} SFT model trained on the union of all sub-task corpora with mild oversampling of small-data sub-tasks (see Section~\ref{sec:discussion:why-grouped-sft}).

\paragraph{Baselines.} For each task, we compare \method against representative baselines drawn from the prior literature. The selected baselines span two complementary types of comparison: (a) \emph{generalist biological foundation models} that, like \method, attempt to cover multiple task categories under a shared framework (e.g., BioT5+~\cite{biot5+}, NatureLM~\cite{naturelm}, SciReasoner~\cite{scireasoner}, 3D-MoLM~\cite{3dmolm}, ESM3~\cite{esm3}, ProtTeX~\cite{prottex}), against which we test whether \method advances the multi-task frontier within the unified-tokenization paradigm; and (b) \emph{task-specialized state-of-the-art models} that represent the strongest published numbers on individual benchmarks (e.g., NExT-Mol~\cite{nextmol}, MolGPT~\cite{molgpt}, ESMFold~\cite{esm2_esmfold}, ProteinMPNN~\cite{proteinmpnn}, Uni-Mol~\cite{unimol}), against which we test whether \method's task-group variants, derived from the unified backbone, can match dedicated specialists. The specific baselines considered for each task are listed alongside the corresponding results in Sections~\ref{sec:eval_mol}--\ref{sec:eval_inter}. We additionally report both \method-1.7B and \method-4B throughout to provide a within-family scale comparison under an otherwise identical recipe.

\paragraph{Evaluation protocol.} For every sub-task in Table~\ref{tab:sft_tasks}, we follow the official data split and evaluation metric provided by its source benchmark. 
All inference is performed with vLLM~\cite{vllm}.
We adopt two decoding configurations depending on the nature of the task. For tasks whose evaluation does not involve output diversity---such as property prediction, name conversion, annotation prediction, question answering---we use greedy decoding with temperature $0$ to obtain a deterministic prediction. 
For tasks whose evaluation explicitly involves sampling diverse outputs---such as unconditional molecule generation or text-based protein design---we instead use nucleus sampling with $\text{top-}p = 1.0$, and adaptively tune the sampling temperature on a per-task basis to balance fidelity and diversity according to the conventions of each benchmark.

\section{Molecule Tasks}
\label{sec:eval_mol}

\begin{figure}[t]
    \centering
    \includegraphics[width=\linewidth]{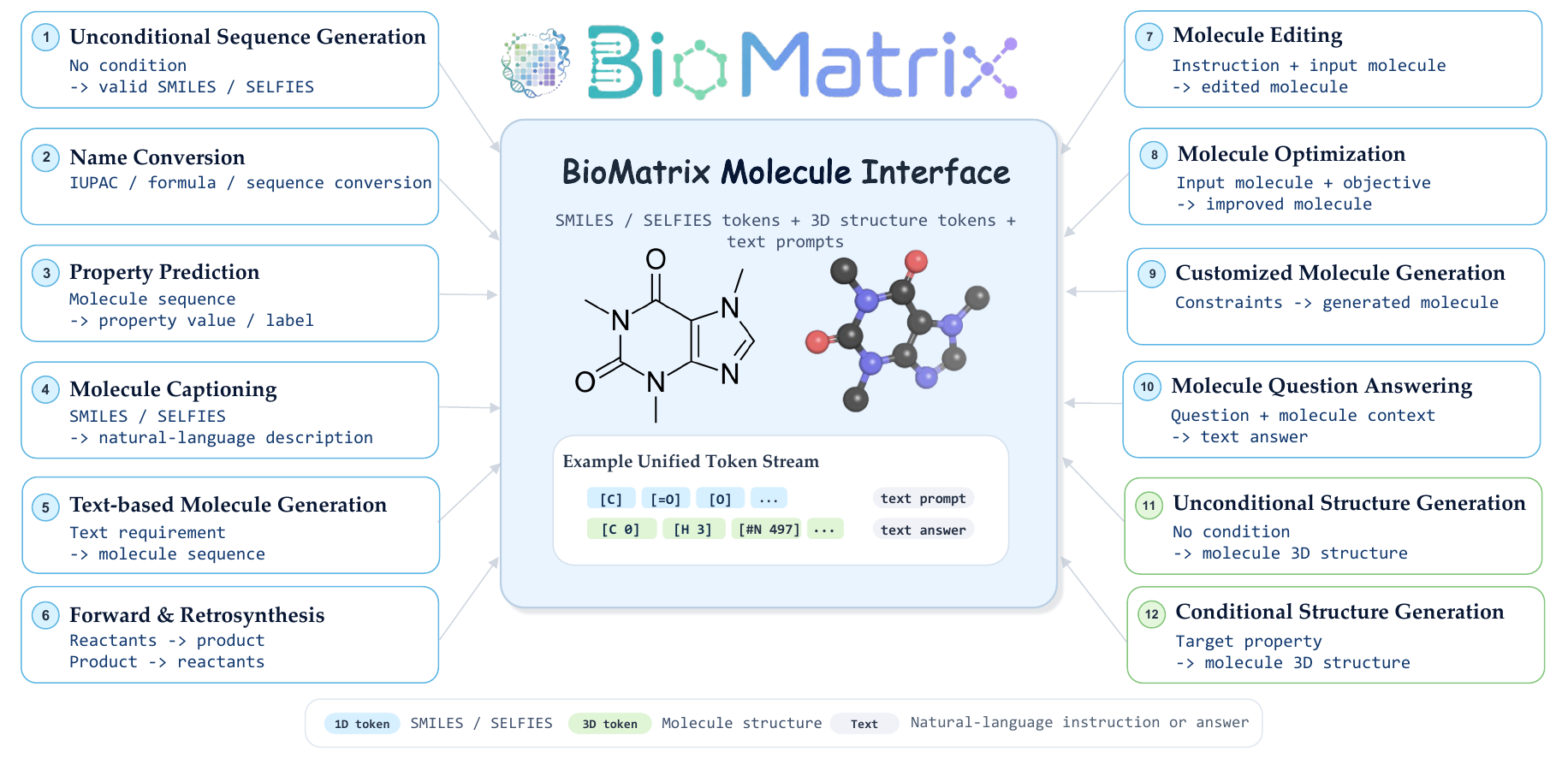}
    \caption{Overview of the molecular task suite evaluated in this section.}
    \label{fig:molecule_tasks}
\end{figure}

We evaluate \method on molecular tasks spanning both sequence-based and structure-based settings, as summarized in Figure~\ref{fig:molecule_tasks}. Section~\ref{sec:eval_mol_1d} covers tasks that take or produce molecular line notations together with natural language, while Section~\ref{sec:eval_mol_3d} covers tasks that involve molecular geometry. 

\begin{CorollaryBox}
\textbf{Key takeaways on molecule tasks.}
\begin{itemize}
    \item \textbf{Strong parameter efficiency against larger LLM baselines.} Across sequence-level molecular tasks—spanning unconditional generation, name conversion, property prediction, captioning, text-based generation, forward and retrosynthesis, editing, customized generation, optimization, and question answering—\method achieves the best or second-best result on the vast majority of metrics among LLM-based methods, often surpassing baselines that are several times larger in parameter count.
    \item \textbf{Native multimodality unlocks gains on cross-modal tasks.} The improvements over prior specialists are most pronounced on tasks that explicitly bridge modalities or conditioning signals: text-based molecule generation, molecule captioning, and especially property-conditioned conformer generation, where \method reduces MAE on electronic-structure targets by roughly $3$--$4\times$ over the strongest prior method. This pattern directly validates the value of placing sequence, structure, and text in a single discrete token space.
    \item \textbf{SMILES and SELFIES are complementary, not interchangeable.} By natively supporting both notations under an otherwise identical recipe, \method reveals a clean task--representation match: SELFIES dominates tasks rewarding distribution-level chemical validity (unconditional generation, property optimization), while SMILES dominates tasks rewarding surface-level structural anchoring (customized generation, synthesis prediction). Maintaining both notations is therefore a feature, not a redundancy.
    \item \textbf{Tokenization, not modeling capacity, bounds fine-grained geometric fidelity.} On unconditional conformer generation, \method matches diffusion-based baselines on distributional and dihedral metrics but exhibits a residual gap on bond-length precision. A single lightweight MMFF refinement closes most of this gap, indicating that the limitation lies in the irreducible quantization error of finite structural codebooks rather than in next-token modeling itself.
\end{itemize}
\end{CorollaryBox}

\subsection{1D Molecular Tasks}
\label{sec:eval_mol_1d}

\subsubsection{Unconditional Generation}
\label{sec:mol_1d_uncond}

This task evaluates the model's ability to learn the underlying distribution of drug-like molecules purely from data, without any conditioning signal. The model is asked to generate \num{10000} molecules from scratch as line-notation strings. We evaluate on two widely used benchmarks, MOSES~\cite{moses} and GuacaMol~\cite{guacamol}, following their official protocols. Both benchmarks assess the generated set along three standard axes: validity, uniqueness, and novelty. MOSES additionally reports internal diversity (IntDiv1/IntDiv2), which measures the average pairwise dissimilarity among generated molecules via Tanimoto similarity over molecular fingerprints, with higher values indicating that the generated set spans a broader range of distinct structures rather than collapsing onto similar ones.

\begin{table}[htbp]
\centering
\caption{Performance comparison of unconditional 1D molecule generation task across 10,000 samples on the MOSES~\cite{moses} dataset.
The baseline results are derived from~\cite{omggpt}.
}
\label{tab:moses_comparison}
\begin{tabular}{lccccc}
\toprule
Models & Validity & Unique & Novelty & \textit{IntDiv}$_1$ & \textit{IntDiv}$_2$ \\
\midrule
\multicolumn{6}{l}{\textit{Graph-Based}} \\
\midrule
JT-VAE~\cite{jt_vae}       & \textbf{1.0}    & \underline{0.999}           & 0.914           & 0.855           & 0.849 \\
GraphINVENT~\cite{graphinvent}  & 0.964           & 0.998           & --               & 0.857 & 0.851 \\
Digress~\cite{digress}      & 0.857           & 1.0             & 0.936           & 0.855           & 0.849 \\
\midrule
\multicolumn{6}{l}{\textit{Sequence-Based}} \\
\midrule
CharRNN~\cite{charrnn}      & 0.975           & \underline{0.999} & 0.842         & 0.856           & 0.850 \\
AAE~\cite{aae}          & 0.936           & \textbf{1.0}    & 0.793           & 0.855           & 0.850 \\
VAE~\cite{vae}          & 0.977           & 0.998           & 0.700           & 0.856           & 0.850 \\
LatentGAN~\cite{latentgan}    & 0.897           & 0.997           & 0.949 & 0.857 & 0.850 \\
MolGPT~\cite{molgpt}       & 0.988           & \textbf{1.0}    & 0.821           & 0.855           & 0.849 \\
Sc2Mol~\cite{sc2mol}       & 0.631           & 0.990           & \textbf{0.986}  & 0.866  & \textbf{0.872} \\
OMG-GPT~\cite{omggpt}     & 0.990 & \textbf{1.0}  & 0.903           & 0.855           & 0.849 \\
\midrule
\method-1.7B (SMILES) & 0.951 & \textbf{1.0} & 0.941 & 0.873 & \underline{0.866} \\
\method-1.7B (SELFIES) & 0.995 & \textbf{1.0} & 0.891 & 0.865 & 0.858 \\
\method-4B (SMILES) & 0.983 & \textbf{1.0} & \underline{0.954} & \underline{0.867} & 0.860 \\
\method-4B (SELFIES) & \underline{0.998} & \textbf{1.0} & 0.925 & \textbf{0.868} & 0.861 \\
\bottomrule
\end{tabular}
\end{table}
\begin{table}[htbp]
\centering
\caption{Performance comparison of unconditional 1D molecule generation task across 10,000 samples on the GuacaMol~\cite{guacamol} dataset.
The baseline results are derived from~\cite{omggpt}.
}
\label{tab:guacamol_comparison}
\begin{tabular}{lccc}
\toprule
Models & Validity & Unique & Novelty \\
\midrule
\multicolumn{4}{l}{\textit{Graph-Based}} \\
\midrule
Digress~\cite{digress}      & 0.852           & \textbf{1.0}    & 0.996           \\
NAGVAE~\cite{nagvae}       & 0.929           & 0.955           & \textbf{1.0}    \\
\midrule
\multicolumn{4}{l}{\textit{Sequence-Based}} \\
\midrule
SMILES LSTM~\cite{smiles_lstm}  & 0.959           & \textbf{1.0}    & 0.912           \\
VAE~\cite{vae}          & 0.870           & \underline{0.999} & 0.974         \\
AAE~\cite{aae}          & 0.822           & \textbf{1.0}    & \underline{0.998} \\
ORGAN~\cite{organ}        & 0.379           & 0.841           & 0.687           \\
MolGPT~\cite{molgpt}       & 0.979 & \underline{0.999} & \textbf{1.0} \\
OMG-GPT~\cite{omggpt}     & 0.984  & \textbf{1.0}    & 0.975           \\
\midrule
\method-1.7B (SMILES) & 0.994 & \underline{0.999} & 0.957\\
\method-1.7B (SELFIES) & \textbf{0.999} & \textbf{1.0} & 0.977\\
\method-4B (SMILES) & 0.991 & \textbf{1.0} & 0.978 \\
\method-4B (SELFIES) & \underline{0.998} & \textbf{1.0} & 0.986 \\
\bottomrule
\end{tabular}
\end{table}

As shown in Tables~\ref{tab:moses_comparison} and~\ref{tab:guacamol_comparison}, both \method-1.7B and \method-4B produce essentially valid and fully unique sets on both benchmarks, with validity reaching $0.951$--$0.999$ and uniqueness at or near $1.0$. On GuacaMol, \method-1.7B (SELFIES) attains the highest validity ($0.999$) among all sequence- and graph-based methods, while \method-4B (SELFIES) reaches $0.998$ validity together with $1.000$ uniqueness and $0.986$ novelty, jointly outperforming the strongest dedicated molecular generator OMG-GPT~\cite{omggpt} ($0.984\,/\,1.000\,/\,0.975$) across all three axes and remaining competitive with MolGPT~\cite{molgpt} ($0.979\,/\,0.999\,/\,1.000$). On MOSES, \method-4B (SELFIES) likewise delivers the best overall balance, with validity of $0.998$, uniqueness of $1.000$, novelty of $0.925$, and competitive internal diversity (IntDiv1\,=\,$0.868$, IntDiv2\,=\,$0.861$), improving over OMG-GPT ($0.990\,/\,1.000\,/\,0.903\,/\,0.855\,/\,0.849$) on every axis except validity, where the two are essentially tied.

\begin{DefinitionBox}
\textbf{\method occupies a more balanced validity--novelty--diversity operating point than prior generators.} Existing baselines tend to trade one axis for another: Sc2Mol attains the highest novelty (0.986) and internal diversity (IntDiv2 = 0.872) on MOSES but at the cost of validity collapsing to 0.631, meaning most of its ``novel'' outputs are not chemically valid. \method instead keeps validity ($\geq$0.951), uniqueness ($\approx$1.0), novelty ($\geq$0.891), and internal diversity (IntDiv1/2 above the $\sim$0.855 baseline cluster) jointly high. The SELFIES variants additionally guarantee syntactic validity by construction (0.995--0.999), suggesting that the unified next-token objective combined with SELFIES tokenization produces a generated set that is simultaneously valid, distinct, and broad in chemical-space coverage---without the need for hand-tuned trade-offs.
\end{DefinitionBox}

\subsubsection{Name Conversion}
\label{sec:mol_1d_name}
This task evaluates the model's ability to translate between different molecular naming conventions, testing whether the model has internalized the bidirectional mapping between structured line notations (SMILES / SELFIES) and human-readable identifiers (IUPAC names, molecular formulas). Following the protocol of SMolInstruct~\cite{llasmol}, we evaluate on five sub-tasks that collectively probe both directions of this mapping: IUPAC-to-formula (I2F), IUPAC-to-sequence (I2S), sequence-to-formula (S2F), and sequence-to-IUPAC (S2I), where the sequence refers to SMILES for the SMILES variant of \method and SELFIES for the SELFIES variant. For each sub-task we report exact match accuracy (EM) as the primary metric, and additionally report chemical validity for I2S, which produces line-notation strings that must parse into legal molecules.

\begin{table}[htbp]
\centering
\caption{Performance comparison of molecular name conversion task on the SMolInstruct~\cite{llasmol}. The baseline results are derived from~\cite{llasmol,scireasoner,biomedgpt_mol}.
}
\label{tab:mol_1d_name_conversion}
\begin{tabular}{lccccc}
\toprule
\multirow{2}{*}{Model} & I2F & \multicolumn{2}{c}{I2S} & S2F & S2I \\
\cmidrule(lr){2-2} \cmidrule(lr){3-4} \cmidrule(lr){5-5} \cmidrule(lr){6-6}
 & EM $\uparrow$ & EM $\uparrow$ & Validity $\uparrow$ & EM $\uparrow$ & EM $\uparrow$ \\
\midrule
\multicolumn{6}{l}{\textit{Task-specific, non-LLM based models}} \\
\midrule
Literature SOTA & \textbf{97.90} & 73.50 & 99.40 & \textbf{100.00} & \underline{56.50} \\
\midrule
\multicolumn{6}{l}{\textit{General / molecular LLM baselines}} \\
\midrule
GPT-4~\cite{gpt4} & 8.70 & 3.30 & 84.20 & 4.80 & 0.00 \\
GPT-oss~\cite{gpt_oss} & 33.24 & 7.58 & -- & 19.88 & 0.17 \\ 
Claude Opus 3~\cite{claude3} & 34.60 & 17.70 & 90.20 & 9.20 & 0.00 \\
LlaSMol\_Mistral~\cite{llasmol} & 87.90 & 70.10 & 99.60 & 93.20 & 29.00 \\
SciReasoner-1.7B~\cite{scireasoner} & 91.81 & 78.85 & --	 & 95.92 & 49.52 \\
SciReasoner-8B~\cite{scireasoner} & 92.65 & 84.40 & -- & 96.39 & \textbf{56.63} \\
BioMedGPT-Mol-8B~\cite{biomedgpt_mol} & 91.40 & 76.50 & -- & 95.70 & 44.90 \\
\midrule
\method-1.7B (SMILES) & 79.45 & \underline{87.22} & 99.12 & 99.39 & 44.75 \\
\method-1.7B (SELFIES) & 79.42 & 78.39 & \underline{99.78} & \underline{99.46} & 42.58 \\
\method-4B (SMILES) & \underline{94.09} & \textbf{92.83} & 99.75 & 99.28 & 55.21 \\
\method-4B (SELFIES) & \underline{94.12} & 83.10 & \textbf{99.82} & 99.31 & 53.40 \\
\bottomrule
\end{tabular}
\end{table}

As shown in Table~\ref{tab:mol_1d_name_conversion}, \method-4B establishes state-of-the-art performance across the board among LLM-based methods, achieving the best result on four of five metrics and remaining competitive on S2I EM with the strongest generalist baseline SciReasoner-8B while using roughly half the parameters. The improvement over SciReasoner-8B is most pronounced on I2S EM, rising from $84.40\%$ to $92.83\%$. At the $1.7$B scale, \method surpasses SciReasoner-1.7B on all EM metrics, indicating that the advantage is not simply a function of model size.

\subsubsection{Property Prediction}

This task evaluates the model's ability to predict physicochemical and biological properties of molecules directly from their line-notation representations, testing whether the model has internalized the mapping from molecular structure to functional behavior. Following the protocol of SMolInstruct~\cite{llasmol}, we evaluate on six widely used MoleculeNet~\cite{moleculenet} benchmarks spanning two regression tasks---ESOL and Lipo, measured by root-mean-square error (RMSE)---and four classification tasks---BBBP, ClinTox, HIV, and SIDER, measured by accuracy.

\begin{table}[htbp]
\centering
\caption{Performance comparison of property prediction task on the SMolInstruct~\cite{llasmol}. The baseline results are derived from~\cite{llasmol,scireasoner,biomedgpt_mol}
}
\label{tab:mol_1d_prop_pred}
\begin{tabular}{lcccccc}
\toprule
\multirow{2}{*}{Model} & ESOL & Lipo & BBBP & ClinTox & HIV & SIDER \\
\cmidrule(lr){2-2} \cmidrule(lr){3-3} \cmidrule(lr){4-4} \cmidrule(lr){5-5} \cmidrule(lr){6-6} \cmidrule(lr){7-7}
 & RMSE $\downarrow$ & RMSE $\downarrow$ & ACC $\uparrow$ & ACC $\uparrow$ & ACC $\uparrow$ & ACC $\uparrow$ \\
\midrule
\multicolumn{7}{l}{\textit{Task-specific, non-LLM based models}} \\
\midrule
Literature SOTA & \underline{0.819} & \textbf{0.612} & 85.30 & 92.40 & 97.00 & 70.00 \\
\midrule
\multicolumn{7}{l}{\textit{General / molecular LLM baselines}} \\
\midrule
GPT-4~\cite{gpt4} & 2.570 & 1.545 & 62.90 & 50.00 & 59.60 & 57.60 \\
GPT-oss~\cite{gpt_oss}  &  3.45  &   1.65  &   46.70 &  52.78 & 69.86 & 38.25 \\
Claude Opus 3~\cite{claude3} & 1.036 & 1.194 & 75.10 & 41.70 & 76.40 & 67.00 \\
LlaSMol\_Mistral~\cite{llasmol} & 1.150 & 1.010 & 74.60 & 93.10 & 96.70 & 70.70 \\
SciReasoner-1.7B~\cite{scireasoner} & 1.21 & 0.94 & 79.70 & 92.36 & 96.79 & 70.00 \\
SciReasoner-8B~\cite{scireasoner} & 1.080 & 0.820 & 82.74 & 91.67 & 96.81 & 68.53 \\
BioMedGPT-Mol-8B~\cite{biomedgpt_mol} & 0.916 & 0.973 & \textbf{87.60} & 93.10 & -- & -- \\
\midrule
\method-1.7B (SMILES) & \textbf{0.810} & 0.860 & 81.62 & 90.91 & 97.43 & \underline{72.39} \\
\method-1.7B (SELFIES) & 0.860 & 0.850 & 82.70 & 92.31 & 97.32 & 72.14 \\
\method-4B (SMILES) & 1.240 & \underline{0.780} & 84.86 & \underline{93.71} & \textbf{97.56} & 72.35 \\
\method-4B (SELFIES) & 1.740 & 0.840 & \underline{85.95} & \textbf{94.41} & \underline{97.48} & \textbf{72.48} \\
\bottomrule
\end{tabular}
\end{table}

As shown in Table~\ref{tab:mol_1d_prop_pred}, \method achieves the best or second-best result among LLM-based methods on five of six benchmarks, and surpasses the task-specific literature SOTA on three of them (HIV, SIDER, and ESOL for the 1.7B variant). \method-1.7B already outperforms the much larger SciReasoner-8B and BioMedGPT-Mol-8B on four of six benchmarks, again indicating that the advantage is not primarily driven by model scale.
The four classification benchmarks (BBBP, ClinTox, HIV, SIDER) yield tightly clustered results across all four \method variants, indicating stable performance on these tasks. The two regression benchmarks show larger fluctuations—ESOL RMSE in particular ranges from 0.810 to 1.740 across variants without a clean pattern—which we attribute to the higher sensitivity of RMSE as a continuous, unbounded metric compared to classification accuracy, rather than to systematic differences between variants.

\subsubsection{Molecule Captioning and Text-based Molecule Generation}
\label{sec:mol_1d_mc_mg}
These two tasks jointly evaluate the model's bidirectional alignment between molecular structures and natural-language descriptions: captioning asks the model to produce a textual description given a molecule, while text-based generation asks it to produce a molecule given a textual description. 
Following the protocol of SMolInstruct~\cite{llasmol}, we evaluate captioning with METEOR against reference descriptions, and text-based generation with exact match accuracy (EM), fingerprint-based Tanimoto similarity (FTS) to the reference molecule, and chemical validity of the generated line notations.

\begin{table}[htbp]
    \centering
    \caption{Performance comparison of molecule captioning (METEOR) and text-based molecule generation (EM, FTS, Valid) tasks on the SMolInstruct~\cite{llasmol} dataset. The baseline results are derived from~\cite{llasmol,scireasoner,biomedgpt_mol}.
         }
    \label{tab:mol_1d_mc_mg}
    \begin{tabular}{lcccc}
    \toprule
    Model & METEOR $\uparrow$ & EM $\uparrow$ & FTS $\uparrow$ & Validity $\uparrow$ \\
    \midrule
    \multicolumn{5}{l}{\textit{Task-specific, non-LLM based models}} \\
    \midrule
    Literature SOTA & 51.5 & 31.70 & 73.20 & 95.30 \\
    \midrule
    \multicolumn{5}{l}{\textit{General / molecular LLM baselines}} \\
    \midrule
    GPT-4~\cite{gpt4} & 18.8 & 6.40 & 42.60 & 81.40 \\
    GPT oss~\cite{gpt_oss} & 15.00 & 3.00 &--&--\\
    Claude Opus 3~\cite{claude3} & 21.9 & 12.30 & 57.60 & 92.60 \\
    LlaSMol\_Mistral~\cite{llasmol} & 45.2 & 19.20 & 61.70 & 99.70 \\
    SciReasoner-1.7B~\cite{scireasoner} & 59.0 & 39.00 & -- & --\\
    SciReasoner-8B~\cite{scireasoner} & 60.0 & 48.00 & -- & -- \\
    BioMedGPT-Mol-8B~\cite{biomedgpt_mol} & 51.5 & 29.60 & 77.50 & 98.90 \\
    \midrule
    \method-1.7B (SMILES) & 60.2 & 56.35 & 80.76 & \underline{99.72} \\
    \method-1.7B (SELFIES) & 59.7 & 53.94 & 79.81 & \textbf{99.82} \\
    \method-4B (SMILES) & \underline{61.4} & \textbf{65.07} & \textbf{85.47} & 99.52 \\
    \method-4B (SELFIES) & \textbf{61.5} & \underline{63.98} & \underline{84.75} & 99.47 \\
    \bottomrule
    \end{tabular}%
    \end{table}

As shown in Table~\ref{tab:mol_1d_mc_mg}, \method achieves the best result among LLM-based methods on all four metrics and additionally surpasses the task-specific literature SOTA on three of them (METEOR, EM, and FTS). The gains are most pronounced on text-based generation, where \method-4B (SMILES) improves EM from the prior LLM best of $48.00\%$ (SciReasoner-8B) to $65.07\%$, and FTS from $77.50\%$ (BioMedGPT-Mol-8B) to $85.47\%$. \method-1.7B already outperforms every LLM baseline on all four metrics, including the much larger SciReasoner-8B and BioMedGPT-Mol-8B.

\subsubsection{Forward Synthesis and Retrosynthesis}
\label{sec:mol_1d_fs_rs}
These two tasks jointly evaluate the model's understanding of chemical reactions in both directions: forward synthesis asks the model to predict the product given a set of reactants, while retrosynthesis asks it to predict plausible reactants given a target product. Following the protocol of SMolInstruct~\cite{llasmol}, we report exact match accuracy (EM), fingerprint-based Tanimoto similarity (FTS) to the reference molecule, and chemical validity of the generated line notations for both directions.

\begin{table}[htbp]
\centering
\caption{Performance comparison of forward synthesis and retrosynthesis tasks on the SMolInstruct~\cite{llasmol} dataset. The baseline results are derived from~\cite{llasmol,scireasoner,biomedgpt_mol}.}
\label{tab:mol_1d_forward_retro}
\begin{tabular}{lcccccc}
\toprule
\multirow{2}{*}{Model} & \multicolumn{3}{c}{Forward Synthesis} & \multicolumn{3}{c}{Retrosynthesis} \\
\cmidrule(lr){2-4} \cmidrule(lr){5-7}
 & EM $\uparrow$ & FTS $\uparrow$ & Validity $\uparrow$ & EM $\uparrow$ & FTS $\uparrow$ & Validity $\uparrow$ \\
\midrule
\multicolumn{7}{l}{\textit{Task-specific, non-LLM based models}} \\
\midrule
Literature SOTA & \textbf{78.70} & \underline{92.20} & \textbf{100.00} & \textbf{47.00} & \textbf{77.50} & 99.70 \\
\midrule
\multicolumn{7}{l}{\textit{General / molecular LLM baselines}} \\
\midrule
GPT-4~\cite{gpt4} & 1.60 & 40.50 & 87.00 & 0.00 & 33.40 & 42.60 \\
GPT-oss~\cite{gpt_oss} & 33.00 &--&-- & 8.00 &--&-- \\
Claude Opus 3~\cite{claude3} & 3.70 & 45.70 & 97.00 & 1.10 & 46.20 & 94.80 \\
LlaSMol\_Mistral~\cite{llasmol} & 63.30 & 84.90 & 99.80 & 32.90 & 70.40 & \textbf{100.00} \\
SciReasoner-1.7B~\cite{scireasoner} &69.00  & -- & -- &41.00 & -- & -- \\
SciReasoner-8B~\cite{scireasoner} & 75.00 & -- & -- & \underline{46.00} & -- & -- \\
BioMedGPT-Mol-8B~\cite{biomedgpt_mol} & 67.20 & 88.50 & 99.80 & 32.40 & \underline{75.20} & 99.90 \\
\midrule
\method-1.7B (SMILES) & 72.75 & \textbf{92.22} & 99.74 & 42.69 & 73.81 & 99.77 \\
\method-1.7B (SELFIES) & 62.78 & 84.24 & \underline{99.93} & 39.32 & 71.97 & \underline{99.93} \\
\method-4B (SMILES) & \underline{77.94} & 89.88 & \underline{99.93} & 45.16 & 74.88 & \underline{99.93} \\
\method-4B (SELFIES) & 69.08 & 87.01 & 99.85 & 42.90 & 73.63 & \underline{99.93} \\
\bottomrule
\end{tabular}
\end{table}

As shown in Table~\ref{tab:mol_1d_forward_retro}, \method achieves the best or second-best result among LLM-based methods on all six metrics. The SMILES variants consistently outperform the SELFIES variants on EM, with \method-4B (SMILES) reaching $77.94\%$ on forward synthesis EM and $45.16\%$ on retrosynthesis EM---substantially ahead of the prior LLM best (SciReasoner-8B at $75.00\%$ / $46.00\%$) on forward synthesis while remaining competitive on RS.

\subsubsection{Molecule Editing}
\label{sec:mol_1d_mol_edit}
This task evaluates the model's ability to perform localized structural modifications to a given molecule under a natural-language instruction—adding, deleting, or substituting a chemical component while leaving the rest of the molecule intact. It thereby probes a finer-grained capability than text-based generation: the model must locate the target substructure, apply the requested edit, and emit a chemically valid output that differs from the input only in the intended way. 
Following the protocol of S\textsuperscript{2}-Bench~\cite{s2bench}, we evaluate on three sub-tasks—component addition (AddComp), component deletion (DelComp), and component substitution (SubComp)—and we report success rate (SR) and weighted success rate (WSR), where the latter additionally weights each successful edit by its structural similarity to the input molecule, so that high WSR requires the edit to be both successful and localized.

\begin{table}[htbp]
    \centering
    \caption{Performance comparison of molecule editing task on the S\textsuperscript{2}-Bench~\cite{s2bench}. 
    The baseline results are derived from~\cite{s2bench}.
    }
    \label{tab:mol_1d_moledit}
    \begin{tabular}{lcccccc}
        \toprule 
        \multirow{2}{*}{Model}                                   & \multicolumn{2}{c}{AddComp} & \multicolumn{2}{c}{DelComp} & \multicolumn{2}{c}{SubComp} \\
        \cmidrule(lr){2-3} \cmidrule(lr){4-5} \cmidrule(lr){6-7} & SR $\uparrow$               & WSR $\uparrow$              & SR $\uparrow$              & WSR $\uparrow$     & SR $\uparrow$      & WSR $\uparrow$     \\
        \midrule GPT-4o~\cite{gpt_4o}                                          & 0.6188                      & 0.4197                      & 0.7012                     & 0.4234             & \underline{0.7992} & \underline{0.5774} \\
        GPT-4-turbo~\cite{gpt4_turbo}                                              & 0.6990                      & 0.4848                      & 0.7244                     & 0.4154             & 0.7778             & 0.5696             \\
        Claude-3.5~\cite{claude35_sonnet}                                               & 0.6832                      & 0.4794                      & 0.5414                     & 0.3615             & \textbf{0.8104}    & \textbf{0.5924}    \\
        Gemini-1.5-pro~\cite{gemini15}                                           & 0.7058                      & 0.4794                      & 0.7590                     & 0.4515             & 0.7148             & 0.5103             \\
        Llama3.1-8B (OpenMolIns-xlarge)~\cite{s2bench}                          & 0.7790                      & 0.5273                      & 0.8640                     & 0.5327             & 0.6100             & 0.4535             \\
        \midrule \method-1.7B (SMILES)                         & 0.8816                      & 0.6100                      & 0.9524                     & \underline{0.5760} & 0.6034             & 0.4556             \\
        \method-1.7B (SELFIES)                                 & 0.8522                      & 0.5853                      & \underline{0.9626}         & 0.5693             & 0.5858             & 0.4369             \\
        \method-4B (SMILES)                                    & \textbf{0.9294}             & \textbf{0.6447}             & 0.9568                     & \textbf{0.5774}    & 0.6082             & 0.4600             \\
        \method-4B (SELFIES)                                   & \underline{0.9124}          & \underline{0.6272}          & \textbf{0.9666}            & 0.5724             & 0.5990             & 0.4501             \\
        \bottomrule
    \end{tabular}
\end{table}

As shown in Table~\ref{tab:mol_1d_moledit}, \method establishes new state-of-the-art results on AddComp and DelComp across both metrics. On AddComp, \method-4B (SMILES) reaches an SR of $0.9294$ and a WSR of $0.6447$, improving over the strongest baseline Llama3.1-8B (OpenMolIns-xlarge)~\cite{s2bench} by a substantial margin on both axes despite using fewer than half the parameters. 
On DelComp, \method-4B (SELFIES) attains the highest SR of $0.9666$, with all four \method variants exceeding the prior best across both SR and WSR. \method-1.7B already outperforms every baseline on these two sub-tasks.
On SubComp, however, \method trails the strongest proprietary baseline Claude-3.5~\cite{claude35_sonnet}, whose SR of $0.8104$ remains well above the best \method result. 
We attribute this asymmetry to the intrinsic difficulty of substitution as an editing operation: unlike addition or deletion, which involve a single localized modification, substitution requires the model to jointly perform a deletion and an insertion at the same site while preserving the surrounding chemical context, leaving substantially less room for error. The fact that even the much larger Llama3.1-8B baseline drops to an SR of $0.6100$ on this sub-task (well below its own AddComp and DelComp scores) suggests that SubComp is broadly challenging for open-source instruction-tuned models.

\subsubsection{Customized Molecule Generation}
\label{sec:mol_1d_custom_gen}
This task evaluates the model's ability to generate molecules that satisfy explicit structural constraints—a specified atom count, bond count, or set of functional groups—rather than reproducing a particular molecule described in free text. 
Compared to text-based molecule generation (Section~\ref{sec:mol_1d_mc_mg}), where each prompt describes a single target molecule, customized generation specifies only a structural recipe and admits any chemically valid molecule satisfying it. This shifts the task from one of imitation to one of constrained enumeration: the model must internalize what each constraint means at the structural level (e.g., counting atoms of a given element, recognizing functional-group substructures) and then produce a molecule that simultaneously satisfies the constraint and remains chemically valid. We follow the protocol of S\textsuperscript{2}-Bench~\cite{s2bench} and evaluate on three sub-tasks—atom-number control (AtomNum), bond-number control (BondNum), and functional-group inclusion (FuncGroup)—reporting success rate (SR) and weighted success rate (WSR), where the latter additionally weights each successful generation by its novelty against a large external reference database (Zinc-250K~\cite{zinc15}), so that high WSR requires the model to satisfy the constraints with structurally novel molecules.

\begin{table}[htbp]
    \centering
    \caption{Performance comparison of customized molecule generation task on the S\textsuperscript{2}-Bench~\cite{s2bench}. The baseline results are derived from~\cite{s2bench}.}
    \label{tab:mol_1d_custom}
    \begin{tabular}{lcccccc}
    \toprule
    \multirow{2}{*}{MolCustom} & \multicolumn{2}{c}{AtomNum} & \multicolumn{2}{c}{BondNum} & \multicolumn{2}{c}{FuncGroup} \\
    \cmidrule(lr){2-3} \cmidrule(lr){4-5} \cmidrule(lr){6-7}
    & SR $\uparrow$ & WSR $\uparrow$ & SR $\uparrow$ & WSR $\uparrow$ & SR $\uparrow$ & WSR $\uparrow$ \\
    \midrule
    GPT-4o~\cite{gpt_4o} & 0.1998 & 0.1339 & 0.0650 & 0.0412 & 0.2330 & 0.1518 \\
    GPT-4-turbo~\cite{gpt4_turbo} & 0.1702 & 0.1190 & 0.0774 & 0.0488 & 0.2180 & 0.1440 \\
    Claude-3.5~\cite{claude35_sonnet} & 0.1928 & 0.1335 & 0.1058 & 0.0697 & 0.2364 & 0.1556 \\
    Gemini-1.5-pro~\cite{gemini15} & 0.1742 & 0.1202 & 0.0708 & 0.0462 & 0.2486 & 0.1659 \\
    Llama3.1-8B (OpenMolIns-xlarge)~\cite{s2bench} & 0.1186 & 0.0811 & \textbf{0.1196} & \textbf{0.0807} & \textbf{0.3548} & \textbf{0.2268} \\
    \midrule
    \method-1.7B (SMILES) & \underline{0.2176} & \underline{0.1471} & 0.0996 & 0.0702 & 0.2408 & 0.1497 \\
    \method-1.7B (SELFIES) & 0.0680 & 0.0492 & 0.0274 & 0.0178 & 0.0098 & 0.0062 \\
    \method-4B (SMILES) & \textbf{0.2774} & \textbf{0.1899} & \underline{0.1104} & \underline{0.0795} & \underline{0.2714} & \underline{0.1703} \\
    \method-4B (SELFIES) & 0.0262 & 0.0184 & 0.0056 & 0.0041 & 0.0048 & 0.0032 \\
    \bottomrule
\end{tabular}
\end{table}

As shown in Table~\ref{tab:mol_1d_custom}, \method-4B (SMILES) achieves the best result on AtomNum across both metrics, lifting AtomNum SR to $0.2774$ and clearly surpassing every proprietary and open-source baseline. On BondNum, \method-4B (SMILES) reaches an SR of $0.1104$, essentially matching the strongest baseline Llama3.1-8B (OpenMolIns-xlarge)~\cite{s2bench} ($0.1196$) despite using roughly half the parameters. A similar pattern holds on FuncGroup, where Llama3.1-8B retains the lead at SR $0.3548$ and \method-4B (SMILES) reaches $0.2714$, remaining competitive with the proprietary frontier models.

The most striking observation is the large gap between the SMILES and SELFIES variants across all three sub-tasks: on FuncGroup, for example, \method-4B (SMILES) reaches an SR of $0.2714$ whereas \method-4B (SELFIES) collapses to $0.0048$. This pattern, which is largely absent from the imitation-style tasks evaluated earlier, points to a structural mismatch between SELFIES and the kind of constraints this task imposes, and the mismatch manifests differently across the three sub-tasks.

For \emph{AtomNum}, the difficulty is that the number of atom tokens in a SELFIES string does not in general equal the number of heavy atoms in the resulting molecule. SELFIES uses control tokens such as \texttt{[Branch1]} and \texttt{[Ring1]} that are each followed by one token whose role is not to contribute an atom, but to act as an index into the SELFIES alphabet that encodes the branch length or ring size. Concretely, the two SELFIES strings \texttt{[C][C][Branch1][O][C][C][C]} and \texttt{[C][C][C][C][C][C][Ring1][N]} each contain six \texttt{[C]} tokens, yet decode to pentane (\texttt{CCCCC}, five carbons) and cyclohexane (\texttt{C1CCCCC1}, six carbons) respectively: in the first string, the \texttt{[C]} token immediately following \texttt{[Branch1]} is consumed as a branch-length index and contributes no atom, while in the second string the \texttt{[N]} token following \texttt{[Ring1]} encodes the ring size and likewise contributes no atom. A model cannot therefore satisfy a constraint such as ``contains six carbons'' by simply emitting six \texttt{[C]} tokens, whereas in SMILES every heavy-atom character is a heavy atom.

For \emph{BondNum}, the difficulty is that the markers needed to count bonds in SELFIES are not standalone tokens the model can enumerate. In SMILES, non-single bonds appear as dedicated characters—\texttt{=} for double and \texttt{\#} for triple—and ring-closing bonds as matching pairs of digits, so checking a constraint like ``contains two double bonds and one triple bond'' reduces to counting \texttt{=} and \texttt{\#} characters. In SELFIES, by contrast, the same information is encoded in two ways that both obscure it. First, double and triple bonds are fused into the following atom token rather than existing as standalone tokens: acetylene is written \texttt{C\#C} in SMILES (one \texttt{\#}) but \texttt{[C][\#C]} in SELFIES, with the triple-bond marker now baked into the second atom token. Second, ring-closing bonds are encoded through \texttt{[Ring1]} together with an index token specifying the ring size, and the bond type of the ring closure is itself carried as a prefix on that same index token (e.g., the benzene fragment in benzoic acid ends with \texttt{[Ring1][=Branch1]}, where \texttt{[=Branch1]} serves simultaneously as a ring-size index and as an indicator that the closing bond is a double bond). Counting bonds from a SELFIES string therefore requires parsing the internal structure of several distinct token classes rather than aggregating over one, whereas in SMILES every non-default bond has a dedicated, countable surface form.

For \emph{FuncGroup}, the difficulty lies in surface-form instability. Common functional groups in the benchmark pool (benzene rings, carboxyl groups, amides, and so on) correspond in SMILES to short, recurring substrings: every benzene ring is written as \texttt{c1ccccc1} (or its kekulized variants), every carboxyl group as \texttt{C(=O)O}, every primary amide as \texttt{C(=O)N}, regardless of where they appear in the molecule. In SELFIES, by contrast, the same functional group can be split across discontiguous token positions depending on its surrounding context. In benzoic acid (\texttt{OC(=O)c1ccccc1}), the benzene ring encodes into a contiguous block \texttt{[C][=C][C][=C][C][=C][Ring1][=Branch1]}. In biphenyl (\texttt{c1ccc(cc1)-c1ccccc1}), however, the encoder chooses to realize the first benzene ring as a ring whose tail is wrapped inside a branch: the first four ring atoms appear as \texttt{[C][=C][C][=C]}, followed by a \texttt{[Branch1][Branch1]} control pair that opens a branch containing the remaining two ring atoms and the ring-closing marker (\texttt{[C][=C][Ring1][=Branch1]}), after which the second benzene ring follows as its own contiguous block. The same chemical substructure (a benzene ring) therefore has no single canonical token signature in SELFIES, and a model cannot anchor a constraint such as ``contains a benzene ring'' to one recognizable token pattern.

\begin{DefinitionBox}
\textbf{SELFIES is great for open-ended generation, but this same feature makes it unsuitable for tasks with strict structural constraints.} Across all three customized-generation sub-tasks, the SELFIES variants of \method collapse to near-zero success rates while the SMILES variants achieve state-of-the-art or competitive results. The underlying failure modes---atom tokens reinterpreted as branch/ring indices, bond types folded into neighboring atom tokens, and substructures split across non-contiguous positions---all stem from SELFIES's context-dependent token semantics. This same property guarantees the syntactic validity that benefits unconditional generation (Section~\ref{sec:mol_1d_uncond}), but it prevents the model from anchoring structural constraints to recognizable token patterns.
\end{DefinitionBox}

\subsubsection{Molecule Optimization}
\label{sec:mol_1d_opt}

This task evaluates the model's ability to modify a given molecule under a natural-language instruction so as to improve one or more target properties while keeping the modified molecule structurally close to the original.
Compared to molecule editing (Section~\ref{sec:mol_1d_mol_edit}), which specifies a structural operation (add / delete / substitute a component), optimization specifies a property-level objective and leaves the model free to choose which substructures to modify in order to reach it. 
We evaluate \method on two complementary benchmarks: S2-Bench~\cite{s2bench}, which focuses on single-property optimization, and MolOpt~\cite{drugassist}, which additionally covers multi-property joint optimization.

\paragraph{Single-property optimization.} 
We follow the protocol of S2-Bench~\cite{s2bench} and evaluate on three single-property optimization sub-tasks corresponding to three widely used molecular properties: octanol-water partition coefficient (logP), molar refractivity (MR), and quantitative estimate of drug-likeness (QED). For each sub-task, the model is given an input molecule together with an instruction specifying a target property and the desired direction of change, and is asked to return a modified molecule. 
We report success rate (SR) and weighted success rate (WSR), where the latter additionally weights each successful optimization by its structural similarity to the input molecule.

\begin{table}[htbp]
    \centering
    \caption{Performance comparison of single-property molecule optimization task on the S\textsuperscript{2}-Bench~\cite{s2bench}. The baseline results are derived from~\cite{s2bench}.}
    \label{tab:mol_1d_opt_s2bench}
    \begin{tabular}{lcccccc}
    \toprule
    \multirow{2}{*}{MolOpt} & \multicolumn{2}{c}{logP} & \multicolumn{2}{c}{MR} & \multicolumn{2}{c}{QED} \\
    \cmidrule(lr){2-3} \cmidrule(lr){4-5} \cmidrule(lr){6-7}
    & SR $\uparrow$ & WSR $\uparrow$ & SR $\uparrow$ & WSR $\uparrow$ & SR $\uparrow$ & WSR $\uparrow$ \\
    \midrule
    GPT-4o~\cite{gpt_4o} & 0.7190 & 0.4735 & 0.6864 & 0.4407 & 0.3952 & 0.2442 \\
    GPT-4-turbo~\cite{gpt4_turbo} & 0.7662 & 0.5351 & 0.7388 & 0.5039 & 0.3946 & 0.2599 \\
    Claude-3.5~\cite{claude35_sonnet} & 0.7970 & 0.5678 & 0.6962 & 0.4951 & 0.5361 & 0.3775 \\
    Gemini-1.5-pro~\cite{gemini15} & 0.7712 & 0.5415 & 0.7876 & 0.5312 & 0.4704 & 0.2859 \\
    Llama3.1-8B (OpenMolIns-xlarge)~\cite{s2bench} & 0.8822 & 0.5877 & 0.6982 & 0.4673 & \textbf{0.8648} & \textbf{0.5825} \\
    \midrule
    \method-1.7B (SMILES) & \underline{0.9780} & \textbf{0.6635} & 0.9590 & \underline{0.6436} & 0.8210 & 0.5569 \\
    \method-1.7B (SELFIES) & 0.9660 & 0.6483 & \underline{0.9702} & \underline{0.6527} & 0.7980 & 0.5401 \\
    \method-4B (SMILES) & \textbf{0.9838} & \underline{0.6625} & \underline{0.9708} & 0.6502 & \underline{0.8368} & \underline{0.5672} \\
    \method-4B (SELFIES) & 0.9734 & 0.6495 & \textbf{0.9776} & \textbf{0.6546} & 0.8170 & 0.5504 \\
    \bottomrule
\end{tabular}
\end{table}

As shown in Table~\ref{tab:mol_1d_opt_s2bench}, \method establishes new state-of-the-art performance on logP and MR across both metrics, with all four \method variants surpassing every proprietary and open-source baseline by a clear margin. On QED, \method remains competitive with the strongest baseline Llama3.1-8B (OpenMolIns-xlarge)~\cite{s2bench}, with \method-4B (SMILES) trailing it slightly on both SR and WSR. 

\paragraph{Single- and multi-property optimization.}
We complement the S2-Bench evaluation with the MolOpt-Instructions benchmark~\cite{drugassist}, which extends molecule optimization in two directions: it enlarges the set of single-property objectives to cover drug-relevant attributes beyond logP/MR/QED---namely QED, hydrogen bond acceptor count (acceptor+), hydrogen bond donor count (donor+), aqueous solubility (solubility+), blood--brain barrier permeability (bbbp+), and hERG inhibition (herg-, where the goal is to decrease the value). It additionally introduces multi-property objectives (sol+\&acc+, qed+\&bbbp+) that require the model to optimize two attributes jointly. 
Following the original protocol of~\cite{drugassist}, each task is evaluated under two success criteria. Under the \emph{loose} criterion (L), an optimization is counted as successful whenever the optimized property is shifted in the requested direction relative to the input. Under the \emph{strict} criterion (S), the property must additionally clear a per-property threshold (e.g., $\Delta\text{QED}\geq 0.1$, $\Delta\text{acceptor}\geq 1$); for solubility specifically, the strict criterion further demands that the optimized value land within a prescribed range around the input value rather than merely move in the right direction, making it the most demanding sub-task in the benchmark. We report the fraction of generations satisfying the optimization objective (Correct) together with the fraction that are chemically valid (Validity), under both criteria.

\begin{table}[htbp]
    \centering
        \caption{Performance comparison of molecule optimization task on the MolOpt-Instructions~\cite{drugassist} dataset. The baseline results are derived from~\cite{drugassist}.}
        \label{tab:mol_1d_opt_molopt}
        \resizebox{0.84\textwidth}{!}{
        \begin{tabular}{llcccc}
        \toprule
        Task & Model & Correct (L) $\uparrow$ & Validity (L) $\uparrow$ & Correct (S) $\uparrow$ & Validity (S) $\uparrow$ \\
        \midrule
        \multirow{7}{*}{qed+}
    & GPT-3.5-turbo~\cite{chatgpt35} & 0.15 & 0.97 & 0.15 & 0.96 \\
    & BioMedGPT-LM~\cite{biomedgpt} & 0.15 & 0.34 & 0.09 & 0.32 \\
    & DrugAssist~\cite{drugassist} & 0.76 & \underline{0.99} & 0.63 & 0.97 \\
    \cmidrule(lr){2-6}
    & BioMatrix-1.7B (SMILES) & 0.73 & 0.84 & 0.51 & 0.76 \\
    & BioMatrix-1.7B (SELFIES) & \underline{0.87} & 0.95& \underline{0.688} & 0.95\\
    & BioMatrix-4B (SMILES) & 0.81 & 0.87 & 0.59 & 0.77 \\
    & BioMatrix-4B (SELFIES) & \textbf{0.93} & \textbf{0.99} & \textbf{0.74} & \textbf{0.98} \\
    \midrule
    \multirow{7}{*}{acceptor+}
    & GPT-3.5-turbo~\cite{chatgpt35} & 0.04 & \underline{0.98} & 0.06 & \underline{0.96} \\
    & BioMedGPT-LM~\cite{biomedgpt} & 0.18 & 0.45 & 0.13 & 0.39 \\
    & DrugAssist~\cite{drugassist} & 0.71 & 0.97 & 0.67 & 0.96 \\
    \cmidrule(lr){2-6}
    & BioMatrix-1.7B (SMILES) & 0.82 & 0.86 & 0.72 & 0.84 \\
    & BioMatrix-1.7B (SELFIES) & 0.83& 0.91 & 0.75 & 0.93\\
    & BioMatrix-4B (SMILES) & \underline{0.89} & 0.91 & \underline{0.84} & 0.88 \\
    & BioMatrix-4B (SELFIES) & \textbf{0.96} & \textbf{0.99} & \textbf{0.93} & \textbf{0.97} \\
    \midrule
    \multirow{7}{*}{donor+}
    & GPT-3.5-turbo~\cite{chatgpt35} & 0.10 & \textbf{0.98} & 0.04 & \underline{0.95} \\
    & BioMedGPT-LM~\cite{biomedgpt} & 0.17 & 0.46 & 0.09 & 0.46 \\
    & DrugAssist~\cite{drugassist} & 0.72 & \textbf{0.98} & 0.76 & \underline{0.95} \\
    \cmidrule(lr){2-6}
    & BioMatrix-1.7B (SMILES) & 0.91 & 0.94 & 0.85 & 0.89 \\
    & BioMatrix-1.7B (SELFIES) & 0.80& 0.91& 0.74& 0.93\\
    & BioMatrix-4B (SMILES) & \underline{0.94} & 0.95 & \underline{0.92} & 0.93 \\
    & BioMatrix-4B (SELFIES) & \textbf{0.95} & \textbf{0.98} & \textbf{0.93} & \textbf{0.98} \\
    \midrule
    \multirow{7}{*}{solubility+}
    & GPT-3.5-turbo~\cite{chatgpt35} & 0.16 & 0.94 & 0.05 & \underline{0.95} \\
    & BioMedGPT-LM~\cite{biomedgpt} & 0.18 & 0.27 & 0.09 & 0.35 \\
    & DrugAssist~\cite{drugassist} & 0.80 & \underline{0.98} & 0.41 & \textbf{0.98} \\
    \cmidrule(lr){2-6}
    & BioMatrix-1.7B (SMILES) & \underline{0.96} & 0.97 & \underline{0.63} & 0.80 \\
    & BioMatrix-1.7B (SELFIES) & 0.90& 0.98& 0.51 & 0.87\\
    & BioMatrix-4B (SMILES) & 0.94 & 0.94 & 0.58 & 0.71 \\
    & BioMatrix-4B (SELFIES) & \textbf{0.97} & \textbf{1.00} & \textbf{0.64} & 0.90 \\
    \midrule
    \multirow{7}{*}{bbbp+}
    & GPT-3.5-turbo~\cite{chatgpt35} & 0.10 & 0.97 & 0.10 & 0.95 \\
    & BioMedGPT-LM~\cite{biomedgpt} & 0.16 & 0.26 & 0.07 & 0.22 \\
    & DrugAssist~\cite{drugassist} & 0.82 & \underline{0.99} & 0.61 & \underline{0.98} \\
    \cmidrule(lr){2-6}
    & BioMatrix-1.7B (SMILES) & \textbf{0.95} & 0.97 & \underline{0.79} & 0.89 \\
    & BioMatrix-1.7B (SELFIES) & 0.90 & 0.98 & 0.62& 0.94\\
    & BioMatrix-4B (SMILES) & 0.93 & 0.94 & \textbf{0.81} & 0.89 \\
    & BioMatrix-4B (SELFIES) & \underline{0.94} & \textbf{1.00} & 0.67 & \textbf{0.99} \\
    \midrule
    \multirow{7}{*}{herg-}
    & GPT-3.5-turbo~\cite{chatgpt35} & 0.13 & 0.98 & 0.15 & 0.97 \\
    & BioMedGPT-LM~\cite{biomedgpt} & 0.13 & 0.20 & 0.12 & 0.18 \\
    & DrugAssist~\cite{drugassist} & 0.71 & \underline{0.99} & 0.67 & \underline{0.98} \\
    \cmidrule(lr){2-6}
    & BioMatrix-1.7B (SMILES) & 0.69 & 0.83 & 0.54 & 0.77 \\
    & BioMatrix-1.7B (SELFIES) & \underline{0.84} & 0.96 & \underline{0.72} & 0.93 \\
    & BioMatrix-4B (SMILES) & 0.71 & 0.85 & 0.64 & 0.82 \\
    & BioMatrix-4B (SELFIES) & \textbf{0.94} & \textbf{0.99} & \textbf{0.85} & \textbf{0.99} \\
    \midrule
    \multirow{7}{*}{sol+\&acc+}
    & GPT-3.5-turbo~\cite{chatgpt35} & 0.09 & 0.92 & 0.02 & \underline{0.91} \\
    & BioMedGPT-LM~\cite{biomedgpt} & 0.10 & 0.29 & 0.07 & 0.32 \\
    & DrugAssist~\cite{drugassist} & 0.50 & \underline{0.95} & \underline{0.27} & \textbf{0.95} \\
    \cmidrule(lr){2-6}
    & BioMatrix-1.7B (SMILES) & 0.83 & 0.90 & 0.21 & 0.59 \\
    & BioMatrix-1.7B (SELFIES) & 0.74& 0.93& 0.23& 0.77\\
    & BioMatrix-4B (SMILES) & \underline{0.89} & 0.92 & 0.24 & 0.50 \\
    & BioMatrix-4B (SELFIES) & \textbf{0.91} & \textbf{0.99} & \textbf{0.34} & 0.78 \\
    \midrule
    \multirow{7}{*}{qed+\&bbbp+}
    & GPT-3.5-turbo~\cite{chatgpt35} & 0.09 & 0.96 & 0.06 & 0.95 \\
    & BioMedGPT-LM~\cite{biomedgpt} & 0.16 & 0.35 & 0.11 & 0.36 \\
    & DrugAssist~\cite{drugassist} & 0.65 & \underline{0.99} & 0.41 & \textbf{0.98} \\
    \cmidrule(lr){2-6}
    & BioMatrix-1.7B (SMILES) & 0.70 & 0.86 & 0.45 & 0.73 \\
    & BioMatrix-1.7B (SELFIES) & \underline{0.84} & 0.96& 0.43& 0.91 \\
    & BioMatrix-4B (SMILES) & 0.73 & 0.85 & \underline{0.46} & 0.72 \\
    & BioMatrix-4B (SELFIES) & \textbf{0.88} & \textbf{1.00} & \textbf{0.46} & \underline{0.96} \\
    \bottomrule
    \end{tabular}
    }
\end{table}

As shown in Table~\ref{tab:mol_1d_opt_molopt}, \method-4B (SELFIES) achieves the best Correct and Validity scores on the majority of metric--task combinations under both criteria, with the remaining top positions filled by other \method variants. The improvement over the prior best baseline DrugAssist~\cite{drugassist} is consistent across both single- and multi-property tasks, and is particularly visible on hERG inhibition (widely regarded as one of the hardest properties to optimize due to its complex and data-scarce structure--activity relationships) and on the strict solubility setting, the most demanding sub-task in the benchmark, where the strict criterion requires landing inside a prescribed value range rather than merely shifting in the right direction. The advantage of \method on this benchmark also holds at the smaller scale: \method-1.7B (SELFIES) already surpasses every baseline on Correct for most tasks.

\begin{DefinitionBox}
\textbf{The same SELFIES property that hurts customized generation helps property optimization.} Across all eight MolOpt-Instructions sub-tasks and under both loose and strict success criteria, the SELFIES variants of \method dominate the Correct and Validity columns, with \method-4B (SELFIES) leading on the majority of metric--task combinations. This stands in direct contrast to the customized-generation results in Section~\ref{sec:mol_1d_custom_gen}, where the SELFIES variants collapsed to near-zero. The asymmetry reflects a clean task--representation match: optimization requires producing valid molecules with shifted properties, which SELFIES guarantees by construction and supports by abstracting away surface-form bookkeeping, whereas customized generation requires anchoring constraints to surface tokens, which SELFIES's context-dependent semantics actively obscure.
\end{DefinitionBox}

\subsubsection{Molecule Question Answering}
This task evaluates the model's factual accuracy in molecular knowledge, testing whether the model can correctly answer fine-grained questions about specific molecules rather than producing fluent but factually incorrect descriptions. We evaluate on MoleculeQA~\cite{moleculeqa}, the largest molecule-domain QA benchmark to date, which is designed to expose factual errors in molecular descriptions that traditional lexical-similarity metrics (e.g., BLEU, ROUGE) fail to detect. Following an MMLU~\cite{mmlu}-style multiple-choice format, MoleculeQA covers four sub-categories of molecular knowledge: \textit{Structure}, \textit{Source}, \textit{Property}, and \textit{Application}, with overall accuracy (Total) reported as the primary metric.

\begin{table}[htbp]
\centering
\caption{Performance comparison of molecule question answering task on the MoleculeQA~\cite{moleculeqa}. The baseline results are derived from~\cite{moleculeqa}.}
\label{tab:mol_1d_qa}
\resizebox{0.9\textwidth}{!}{
\begin{tabular}{lccccc}
\toprule
Model & Structure & Source & Property & Application & Total \\
\midrule
GPT-3.5-1106-turbo~\cite{chatgpt35} & 25.60 & 37.60 & 28.04 & 32.22 & 29.29 \\
Galactica-6.7B~\cite{galactica} & 32.35 & 41.92 & 31.05 & 28.21 & 33.96 \\
Mol-Instructions-7B~\cite{mol_instructions} & 37.46 & 47.36 & 32.69 & 29.88 & 38.37 \\
BioMedGPT-LM-7B~\cite{biomedgpt} & 54.19 & 60.01 & 38.85 & 40.9 & 52.23 \\
GPT-4-1106-preview~\cite{gpt4} & 60.94 & 50.19 & 35.57 & 43.91 & 53.47 \\
MolT5-base~\cite{molt5} & 58.01 & 65.85 & 45.14 & 42.24 & 55.39 \\
MoMu-base~\cite{momu} & 61.58 & 65.3 & 43.78 & 43.07 & 57.43 \\
BioT5-base~\cite{biot5} & 65.98 & 69.24 & \textbf{49.11} & 40.73 & 62.03 \\
MolCA-1.3B~\cite{molca} & 71.12 & 70.98 & 47.81 & 43.17 & 64.79 \\
\midrule
\method-1.7B (SMILES) & 79.28 & 74.76 & 43.23 & 44.41 & 70.07 \\
\method-1.7B (SELFIES) & 79.05 & 75.06 & 43.91 & 44.91 & 70.15 \\
\method-4B (SMILES) & \textbf{83.36} & \textbf{77.96} & \underline{47.88} & \textbf{46.24} & \textbf{73.78} \\
\method-4B (SELFIES) & \underline{83.26} & \underline{76.55} & 47.20 & \underline{45.58} & \underline{73.24} \\
\bottomrule
\end{tabular}}
\end{table}

As shown in Table~\ref{tab:mol_1d_qa}, \method-4B establishes new state-of-the-art performance on the Total accuracy ($73.78\%$) as well as on the Structure, Source, and Application sub-categories, substantially surpassing the prior best MolCA-1.3B ($64.79\%$). The improvement is most pronounced on the Structure sub-category, where \method-4B reaches $83.36\%$ versus the prior best of $71.12\%$, indicating that \method captures structural knowledge of molecules with notably higher fidelity. \method-1.7B already outperforms every baseline on Total accuracy and on the Structure, Source, and Application sub-categories, including specialist baselines such as BioT5-base and MolCA-1.3B that are explicitly designed for molecule--text understanding.

One observation worth noting is that the Property sub-category remains the hardest for all models, with \method-4B reaching $47.88\%$---competitive with but not exceeding the prior best BioT5-base ($49.11\%$). We attribute this to the fact that property-related questions often require predicting over physicochemical attributes (e.g., precise solubility values or activity ranges), which is a fundamentally different skill from the factual recall that dominates the other three sub-categories.

\subsection{3D Molecular Tasks}
\label{sec:eval_mol_3d}

\subsubsection{Unconditional 3D Structure Generation}
\label{sec:mol_3d_uncond}
This task evaluates the model's ability to jointly generate a chemically valid molecular graph and a physically plausible conformer, without any conditioning signal. 
We evaluate on QM9-2014~\cite{qm9_2014} following the protocol of~\cite{nextmol}, reporting two sets of metrics that target the 2D graph and 3D geometry of the generated molecules, respectively. The 2D-level metrics comprise atom stability, molecule stability, and validity \& completeness (V\&C) for chemical validity; validity \& uniqueness (V\&U) and validity \& uniqueness \& novelty (V\&U\&N) for diversity; and similarity to nearest neighbor (SNN), fragment similarity (Frag), scaffold similarity (Scaf), and Fr\'echet ChemNet Distance (FCD) for distributional similarity to the test set. The 3D-level metrics comprise atom stability and FCD computed on the 2D graphs reconstructed from the predicted coordinates, together with the maximum mean discrepancy (MMD) of bond lengths, bond angles, and dihedral angles against the reference distribution. 
Training-set statistics are also included as a reference upper bound.

\begin{table}[htbp]
    \centering
    \caption{Performances comparison of unconditional 3D molecule generation
    across \num{10000} samples on the QM9-2014~\cite{qm9_2014} dataset. The
    baseline results are derived from~\cite{nextmol}. $\dag$ denotes post-hoc
    refinement via MMFF force-field optimization. }
    \label{tab:mol_3d_uncond_gen} \resizebox{\textwidth}{!}{%
    \begin{tabular}{lccccccccc}
        \toprule 2D-Metric             & FCD $\downarrow$    & AtomStable $\uparrow$                     & MolStable $\uparrow$                         & V\&C $\uparrow$                             & V\&U $\uparrow$                                & V\&U\&N $\uparrow$  & SNN $\uparrow$      & Frag $\uparrow$     & Scaf $\uparrow$     \\
        \midrule {\color{gray}Train}   & {\color{gray}0.063} & {\color{gray}0.999}                       & {\color{gray}0.988}                          & {\color{gray}0.989}                         & {\color{gray}0.989}                            & {\color{gray}0.000} & {\color{gray}0.490} & {\color{gray}0.992} & {\color{gray}0.946} \\
        MolGPT~\cite{molgpt}           & 0.461               & 0.982                                     & 0.976                                        & 0.977                                       & 0.937                                          & 0.763               & \underline{0.523}   & 0.958               & 0.923               \\
        MolGen~\cite{molgen}           & 0.085               & \textbf{1.000}                            & \underline{0.988}                            & \textbf{1.000}                              & 0.955                                          & 0.479               & 0.500               & 0.988               & 0.934               \\
        CDGS~\cite{cdgs}               & 0.798               & 0.997                                     & 0.951                                        & 0.951                                       & 0.936                                          & \textbf{0.860}      & 0.493               & 0.973               & 0.784               \\
        JODO~\cite{jodo}               & 0.138               & \underline{0.999}                         & \underline{0.988}                            & 0.990                                       & 0.960                                          & 0.780               & 0.522               & 0.986               & 0.934               \\
        MiDi~\cite{midi}               & 0.187               & 0.998                                     & 0.976                                        & 0.980                                       & 0.954                                          & 0.769               & 0.501               & 0.979               & 0.882               \\
        EQGAT-diff~\cite{eqgat_diff}   & 2.157               & \textbf{1.000}                            & 0.972                                        & \textbf{1.000}                              & \textbf{0.996}                                 & 0.695               & 0.479               & 0.949               & 0.707               \\
        NExT-Mol~\cite{nextmol}        & 0.070               & \textbf{1.000}                            & \textbf{0.989}                               & \textbf{1.000}                              & \underline{0.967}                              & \underline{0.802}   & \textbf{0.530}      & \textbf{0.992}      & \textbf{0.945}      \\
        \midrule \method-1.7B          & \textbf{0.064}      & \textbf{1.000}                            & \underline{0.988}                            & \textbf{1.000}                              & 0.948                                          & 0.749               & 0.495               & \underline{0.991}   & \underline{0.944}   \\
        \method-4B                     & \underline{0.066}   & \textbf{1.000}                            & \underline{0.988}                            & \textbf{1.000}                              & 0.949                                          & 0.742               & 0.494               & \underline{0.991}   & 0.942               \\
        \midrule 3D-Metric             & FCD $\downarrow$    & \multicolumn{2}{c}{AtomStable $\uparrow$} & \multicolumn{2}{c}{Bond length $\downarrow$} & \multicolumn{2}{c}{Bond angle $\downarrow$} & \multicolumn{2}{c}{Dihedral angle $\downarrow$} \\
        \midrule {\color{gray}Train}   & {\color{gray}0.877} & \multicolumn{2}{c}{{\color{gray}0.994}}   & \multicolumn{2}{c}{{\color{gray}5.44E-04}}   & \multicolumn{2}{c}{{\color{gray}4.65E-04}}  & \multicolumn{2}{c}{{\color{gray}1.78E-04}}      \\
        G-SchNet~\cite{g_schnet}       & 2.386               & \multicolumn{2}{c}{0.957}                 & \multicolumn{2}{c}{3.62E-01}                 & \multicolumn{2}{c}{7.27E-02}                & \multicolumn{2}{c}{4.20E-03}                    \\
        G-SphereNet~\cite{g_spherenet} & 6.659               & \multicolumn{2}{c}{0.672}                 & \multicolumn{2}{c}{1.51E-01}                 & \multicolumn{2}{c}{3.54E-01}                & \multicolumn{2}{c}{1.29E-02}                    \\
        EDM~\cite{edm}                 & 1.285               & \multicolumn{2}{c}{0.986}                 & \multicolumn{2}{c}{\underline{1.30E-01}}     & \multicolumn{2}{c}{1.82E-02}                & \multicolumn{2}{c}{6.64E-04}                    \\
        MDM~\cite{mdm}                 & 4.861               & \multicolumn{2}{c}{\underline{0.992}}     & \multicolumn{2}{c}{2.74E-01}                 & \multicolumn{2}{c}{6.60E-02}                & \multicolumn{2}{c}{2.39E-02}                    \\
        JODO~\cite{jodo}               & 0.885               & \multicolumn{2}{c}{\underline{0.992}}     & \multicolumn{2}{c}{1.48E-01}                 & \multicolumn{2}{c}{\underline{1.21E-02}}    & \multicolumn{2}{c}{6.29E-04}                    \\
        MiDi~\cite{midi}               & 1.100               & \multicolumn{2}{c}{0.983}                 & \multicolumn{2}{c}{8.96E-01}                 & \multicolumn{2}{c}{2.08E-02}                & \multicolumn{2}{c}{8.14E-04}                    \\
        EQGAT-diff~\cite{eqgat_diff}   & 1.519               & \multicolumn{2}{c}{0.988}                 & \multicolumn{2}{c}{4.09E-01}                 & \multicolumn{2}{c}{1.91E-02}                & \multicolumn{2}{c}{1.14E-03}                    \\
        NExT-Mol~\cite{nextmol}        & 0.879               & \multicolumn{2}{c}{\textbf{0.993}}        & \multicolumn{2}{c}{\textbf{1.15E-01}}        & \multicolumn{2}{c}{\textbf{7.32E-03}}       & \multicolumn{2}{c}{\textbf{1.95E-04}}           \\
        \midrule \method-1.7B          & 1.040               & \multicolumn{2}{c}{0.897}                 & \multicolumn{2}{c}{1.05E+00}                 & \multicolumn{2}{c}{3.16E-02}                & \multicolumn{2}{c}{4.76E-04}                    \\
        \method-4B                     & 1.012               & \multicolumn{2}{c}{0.897}                 & \multicolumn{2}{c}{1.05E+00}                 & \multicolumn{2}{c}{3.09E-02}                & \multicolumn{2}{c}{6.09E-04}                    \\
        \method-1.7B$\dag$             & \underline{0.226}   & \multicolumn{2}{c}{0.985}                 & \multicolumn{2}{c}{1.15E+00}                 & \multicolumn{2}{c}{4.38E-02}                & \multicolumn{2}{c}{\underline{2.19E-04}}        \\
        \method-4B$\dag$               & \textbf{0.225}      & \multicolumn{2}{c}{0.985}                 & \multicolumn{2}{c}{1.16E+00}                 & \multicolumn{2}{c}{4.18E-02}                & \multicolumn{2}{c}{2.41E-04}                    \\
        \bottomrule
    \end{tabular}
    }
\end{table}

Results are shown in Table~\ref{tab:mol_3d_uncond_gen}. On the 2D side, \method delivers strong and well-rounded performance: \method-1.7B attains FCD\,=\,$0.064$, essentially matching the training reference ($0.063$) and the strongest diffusion-based baseline NExT-Mol ($0.070$), and is within $0.001$--$0.002$ of the training distribution on Frag ($0.991$ vs.\ $0.992$) and Scaf ($0.944$ vs.\ $0.946$). Atom stability, molecule stability, and V\&C all reach $1.000\,/\,0.988\,/\,1.000$, on par with or above every baseline. The two axes where \method trails NExT-Mol are V\&U\&N ($0.749$ vs.\ $0.802$) and SNN ($0.495$ vs.\ $0.530$), both reflecting a generated distribution that sits slightly closer to the training set than NExT-Mol's.

On the geometric side, \method shows a more nuanced picture. In the default decoding setting, the raw conformers exhibit competitive dihedral-angle MMD but comparatively loose bond-length fidelity (MMD\,=\,$1.05$) and lower atom stability ($0.897$). The bond-length MMD is highly sensitive to small shifts in the distributional mean, and we therefore interpret this number as indicating a modest systematic offset in predicted bond lengths rather than a catastrophic failure of geometry modeling.
We attribute this primarily to two structural limitations of the MolStrucTok-based reconstruction pipeline. First, its $512$-entry codebook must quantize a four-dimensional continuous descriptor space, imposing an intrinsic quantization error on every per-atom geometric code that no amount of upstream modeling can recover. Second, coordinates are reconstructed autoregressively by sequentially traversing the molecule and rebuilding each local spherical frame atom-by-atom, which causes per-atom decoding errors to accumulate along the traversal and occasionally produces geometrically implausible configurations; this stands in contrast to diffusion-based approaches such as NExT-Mol, which denoise all atomic coordinates jointly and therefore enjoy a global geometric consistency that autoregressive reconstruction does not afford.
\begin{DefinitionBox}
\textbf{Autoregressive structure reconstruction trades global geometric consistency for unification, but the gap is largely closable post-hoc.} \method's raw conformers match diffusion-based baselines on dihedral-angle MMD but exhibit looser bond-length fidelity (MMD = $1.05$) and lower atom stability ($0.897$) than NExT-Mol, reflecting irreducible quantization error from a finite codebook compounded by per-atom error accumulation along the autoregressive traversal. A single lightweight MMFF refinement step closes most of this gap (FCD: $1.04 \rightarrow 0.23$, atom stability: $0.897 \rightarrow 0.985$), indicating that the limitation is concentrated in fine-grained geometric consistency rather than chemical plausibility, and that combining a unified token-level generator with cheap geometric post-processing is a practical path to high-fidelity structural outputs.
\end{DefinitionBox}

\subsubsection{Property-conditioned 3D Structure Generation}
\label{sec:mol_3d_cond}
This task evaluates the model's ability to generate molecular conformers whose quantum-chemical properties match a prescribed target value, testing whether a model has internalized the mapping from molecular structure to physical observables rather than merely reproducing the training distribution. Following the protocol of~\cite{nextmol}, we condition generation on each of six quantum properties from QM9-2014~\cite{qm9_2014}: dipole moment $\mu$, polarizability $\alpha$, heat capacity $C_v$, HOMO energy $\varepsilon_{\text{HOMO}}$, LUMO energy $\varepsilon_{\text{LUMO}}$, and HOMO--LUMO gap $\Delta\varepsilon$. 
To quantify controllability, we follow prior work~\cite{nextmol} and partition the QM9-2014 training set into two disjoint halves of 50K molecules each, $D_a$ and $D_b$: an auxiliary property predictor $\phi_c$~\cite{edm} is trained on $D_a$ and used to estimate the property values of generated molecules, while \method is trained on $D_b$. We then report the mean absolute error (MAE) between the specified target value and the prediction of $\phi_c$ on each generated sample. Following~\cite{nextmol}, we additionally include $\phi_c$'s own MAE on $D_b$ as a lower bound (L-Bound) that reflects the predictor's intrinsic error and is unreachable by any generator.

\begin{table}[htbp]
\centering
\caption{Performance (MAE) comparison of conditional 3D molecule generation across \num{10000} samples for each property on the QM9-2014~\cite{qm9_2014} dataset. The baseline results are derived from \cite{nextmol}.}
\label{tab:mol_3d_cond_gen}
\vspace{-0.2cm}

\begin{tabular}{lccccccc}
  \toprule
  Method & \multicolumn{1}{c}{$\mu\ (\textnormal{D})$} & \multicolumn{1}{c}{$\alpha\ (\textnormal{Bohr}^3)$} & \multicolumn{1}{c}{$C_{v}\ \left(\frac{\textnormal{cal}}{\textnormal{mol}}\textnormal{K}\right)$} & \multicolumn{1}{c}{$\varepsilon_{\textnormal{HOMO}}\ (\textnormal{meV})$} & \multicolumn{1}{c}{$\varepsilon_{\textnormal{LUMO}}\ (\textnormal{meV})$} & \multicolumn{1}{c}{$\Delta\varepsilon\ (\textnormal{meV})$} \\ \midrule
  \bound{L-Bound}         &\bound{0.043}&\bound{0.09}&\bound{0.040}&\bound{\phantom{0}39}&\bound{\phantom{00}36}&\bound{\phantom{00}65}\\ 
  EDM~\cite{edm}                     &       1.123 &       2.78 &       1.065 &                 371 &       \phantom{0}601 &       \phantom{0}671 \\
  EEGSDE~\cite{eegsde}                  &       0.777 &       2.50 &       0.941 &                 302 &       \phantom{0}447 &       \phantom{0}487 \\
  GeoLDM~\cite{geoldm}                  &       1.108 &       2.37 &       1.025 &                 340 &       \phantom{0}522 &       \phantom{0}587 \\
  JODO~\cite{jodo}                    &       0.628 &       1.42 &       0.581 &                 226 &       \phantom{0}256 &       \phantom{0}335 \\ 
  NExT-Mol~\cite{nextmol}         & \underline{0.507} & 1.16& 0.512&           205& \phantom{0}235& \phantom{0}297\\
 \midrule
\method-1.7B            & \textbf{0.211}& \underline{0.53}         & \underline{0.474}         & \textbf{\phantom{0}53}& \phantom{0}\underline{59}           & \phantom{0}\underline{84} \\
\method-4B              & \textbf{0.211}& \textbf{0.47}& \textbf{0.288}& \phantom{0}\underline{57}           & \textbf{\phantom{0}54}  & \textbf{\phantom{0}81} \\
  \bottomrule
\end{tabular}
\vspace{-0.2cm}
\end{table}

As shown in Table~\ref{tab:mol_3d_cond_gen}, \method substantially advances the state of the art across all six properties. \method-4B attains the best MAE on five of six targets and the second-best on the remaining one, with \method-1.7B occupying the complementary positions and itself already surpassing every baseline on every property. On the electronic-structure targets in particular, the improvement over the strongest prior method NExT-Mol~\cite{nextmol} is especially pronounced: MAE drops from $205$ to $53$\,meV ($\varepsilon_{\text{HOMO}}$), $235$ to $54$\,meV ($\varepsilon_{\text{LUMO}}$), and 297 to 81\,meV ($\Delta\varepsilon$), representing roughly a $3$--$4\times$ error reduction. On the remaining three properties, \method-4B likewise more than halves NExT-Mol's error (e.g., $\mu$: $0.507\,\to\,0.211$; $\alpha$: $1.16\,\to\,0.470$). 
\begin{DefinitionBox}
\textbf{Joint tokenization of properties and geometry yields outsized gains on property-conditioned conformer generation.} On the three electronic-structure targets of QM9-2014, \method-4B reduces MAE over the strongest prior method NExT-Mol by roughly $3$--$4\times$ ($\varepsilon_{\text{HOMO}}$: $205 \rightarrow 53$~meV; $\varepsilon_{\text{LUMO}}$: $235 \rightarrow 54$~meV; $\Delta\varepsilon$: $297 \rightarrow 81$~meV), and more than halves the error on the remaining three. The relative improvement here is substantially larger than what we observe on unconditional conformer generation against the same baseline, suggesting that exposing target property values in the same discrete token space as the structure under a single next-token objective is a particularly effective form of conditioning.
\end{DefinitionBox}

\section{Protein Tasks}
\label{sec:eval_prot}

\begin{figure}[h]
    \centering
    \includegraphics[width=\linewidth]{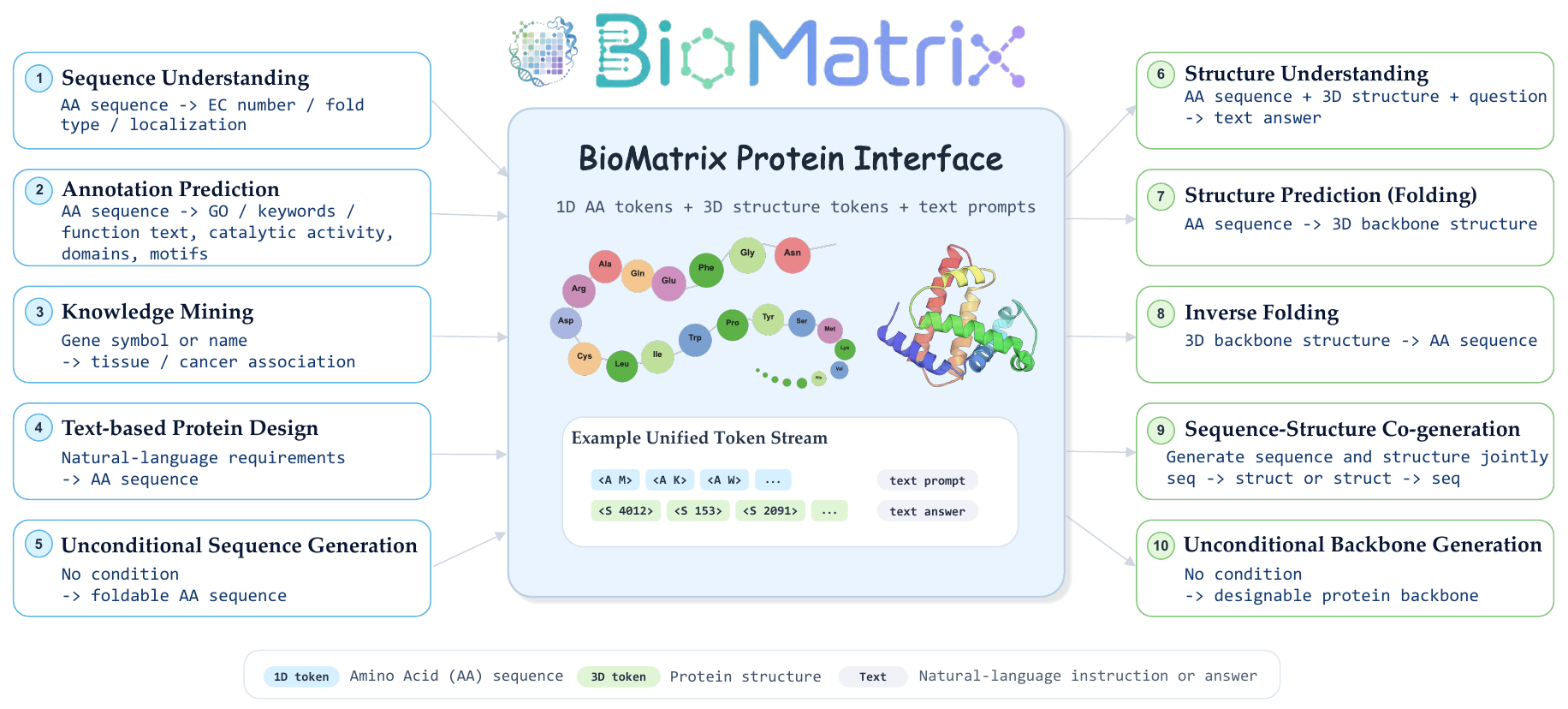}
    \caption{Overview of the protein task suite evaluated in this section.}
    \label{fig:protein_tasks}
\end{figure}

We next evaluate \method on protein tasks spanning both sequence-based and structure-based settings, as summarized in Figure~\ref{fig:protein_tasks}. Section~\ref{sec:eval_prot_1d} covers tasks that take or produce amino acid sequences together with natural language, while Section~\ref{sec:eval_prot_3d} covers tasks involving backbone geometry.

\begin{CorollaryBox}
\textbf{Key takeaways on protein tasks.}
\begin{itemize}
    \item \textbf{Parameter-efficient state-of-the-art across protein understanding.} Across sequence understanding, annotation prediction, and knowledge mining, \method consistently matches or exceeds the performance of substantially larger baselines (including the $8$B-scale SciReasoner and OPI-Llama variants) at both the $1.7$B and $4$B scales, while frontier general-purpose LLMs such as GPT-O3 and DeepSeek-R1 fail to generalize to these domains. 
    \item \textbf{Residue-aligned tokenization makes folding and inverse folding strict symmetric duals.} By preserving a strict one-to-one correspondence between each AA token and its per-residue structure token, folding and inverse folding reduce to clean token-to-token alignment problems on the same vocabulary. \method achieves competitive folding accuracy and state-of-the-art amino acid recovery on inverse folding without any specific architecture or design.
    \item \textbf{Designable backbones and foldable sequences emerge as the default mode of the next-token prior.} On sequence--structure co-generation, unconditional backbone generation, and unconditional sequence generation, \method matches or surpasses dedicated specialists (RFDiffusion, FrameDiff, FoldFlow) and the strongest multimodal baseline DPLM-2 on every standard self-consistency metric, in some cases approaching the native-PDB reference. The consistency of this advantage across decoding orders and conditioning settings suggests that high-quality joint generation is a natural property of the unified token space.
    \item \textbf{Text-grounded protein design without specialist architecture.} On text-based protein design, \method generates sequences that are simultaneously diverse, low-redundancy, and structurally plausible across both CAMEO and MolInst-SwissProtCLAP, remaining competitive with task-specialized baselines under a single unified backbone.
\end{itemize}
\end{CorollaryBox}

\subsection{1D Protein Tasks}
\label{sec:eval_prot_1d}

\subsubsection{Sequence Understanding}

This suite of tasks evaluates whether a model can infer high-level biological properties directly from protein primary sequences, including enzyme function annotation, structural family recognition, and cellular destination prediction. Concretely, we consider four sequence understanding subsets from the OPI benchmark~\cite{opi}: EC Number Prediction, in which the model predicts Enzyme Commission (EC) annotations from sequence alone; Fold Type Prediction, which evaluates hierarchical structural classification at the family, superfamily, and fold levels; and subcellular localization prediction, which measures whether the model can identify the cellular compartment in which a protein is expected to reside. Together, these tasks probe complementary aspects of protein sequence understanding, including sequence-to-function mapping, sequence-to-structure abstraction, and sequence-to-localization reasoning. Following the evaluation protocol established in~\cite{opi}, we employ specific metrics tailored to the nature of each task: Precision, Recall, and the F1-score are utilized for EC number prediction to account for the multi-level classification structure of the task, whereas Accuracy is reported for fold type identification and subcellular localization tasks. These metrics collectively measure the proficiency of the model in recognizing conserved sequence motifs and capturing the global structural and functional implications encoded within the primary sequence.

\begin{table}[htbp]
\centering
\caption{Performance comparison of protein sequence understanding tasks across EC number prediction benchmark. The baseline results are derived from~\cite{opi,scireasoner}}
\label{tab:pro_1d_opi_su_ec}
\begin{tabular}{lcccccc}
\toprule
\multirow{2}{*}{Model} & \multicolumn{3}{c}{CLEAN EC Number Price} & \multicolumn{3}{c}{CLEAN EC Number New} \\ 
\cmidrule(lr){2-4} \cmidrule(lr){5-7} 
 & Precision & Recall & F1 & Precision & Recall & F1 \\ 
\midrule
OPI-Galactica-6.7B~\cite{opi} & 2.68 & 2.68 & 2.68 & 27.00 & 26.63 & 25.96 \\
OPI-Llama-3.1-8B-Instruct~\cite{opi} & 7.38 & 7.38 & 7.38 & 37.24 & 33.74 & 34.68 \\
SciReasoner-1.7B~\cite{scireasoner} & 10.00 & 10.00 & 10.00 & 40.00 & 39.00 & 39.00 \\
SciReasoner-8B~\cite{scireasoner} & 21.00 & 23.00 & 22.00 & \textbf{55.00} & \underline{44.00} & \textbf{54.00} \\
\midrule
\method-1.7B & \textbf{33.89} & \textbf{35.57} & \textbf{34.34} & 42.86 & 43.37 & 43.03 \\
\method-4B & \underline{27.74} & \underline{28.86} & \underline{28.08} & \underline{43.24} & \textbf{44.13} & \underline{43.54}  \\ 
\bottomrule
\end{tabular}
\end{table}

Results for EC number prediction are summarized in Table~\ref{tab:pro_1d_opi_su_ec}. The proposed models demonstrate a substantial improvement in performance compared to existing LLM baselines. Notably, \method-1.7B achieves an F1-score of $34.34$\% on the CLEAN EC Number Price dataset, which significantly outperforms SciReasoner-1.7B ($10.00$\%) and even surpasses the SciReasoner-8B ($22.00$\%) by a large margin. 
On the CLEAN EC Number New split, \method-4B remains competitive with SciReasoner-8B (F1: 43.54\% vs. 54.00\%) despite using roughly half the parameters, with comparable Recall (44.13\% vs. 44.00\%) but lower Precision. These results suggest that \method is particularly effective on the more challenging Price split, while remaining competitive on the New split at substantially smaller scale.

\begin{table}[htbp]
\centering
\caption{Performance comparison of fold type and subcellular localization tasks on the OPI benchmark. The baseline results are derived from~\cite{opi,scireasoner}.}
\label{tab:pro_1d_opi_su_fold}
\resizebox{\textwidth}{!}{
\begin{tabular}{lcccc}
\toprule
\multirow{2}{*}{Model} & \multicolumn{3}{c}{Fold type prediction} & Subcellular localization \\
\cmidrule(lr){2-4} \cmidrule(lr){5-5}
 & Family level & Superfamily level & Fold level & Hold-out \\
\midrule
Gemini 2.5-pro~\cite{gemini25} & -- & -- & -- & 2.00 \\
GPT-O3~\cite{o3_o4mini} & -- & -- & -- & 8.00 \\
DeepSeek-R1~\cite{deepseek_r1} & -- & -- & -- & 5.00 \\
OPI-Galactica-6.7B~\cite{opi} & 49.00 & 13.00 & 8.00 & 78.00 \\
OPI-Llama-3.1-8B-Instruct~\cite{opi} & 61.00 & 15.00 & 10.00 & 42.00 \\
SciReasoner-1.7B~\cite{scireasoner} & -- & -- & -- & 84.00 \\
SciReasoner-8B~\cite{scireasoner} & 83.40 & 28.63 & 15.04 & \textbf{91.00}  \\
\midrule
\method-1.7B  & \underline{85.84} & \underline{29.51} & \textbf{16.71} & 89.90 \\
\method-4B & \textbf{87.25} & \textbf{30.22} & \underline{15.60} & \underline{90.69} \\
\bottomrule
\end{tabular}
}
\end{table}

As shown in Table~\ref{tab:pro_1d_opi_su_fold}, \method also demonstrates strong performance on structural and localization tasks. In fold type prediction, \method-4B achieves state-of-the-art Accuracy at both the Family level with $87.25\%$ and the Superfamily level with $30.22\%$, consistently outperforming all baselines. Even at the more challenging Fold level, which requires remote homology detection from distant evolutionary sequences, \method-1.7B attains an Accuracy of $16.71$\%, the highest among all evaluated models. For subcellular localization, while general-purpose models such as GPT-O3 and DeepSeek-R1 fail to generalize to this domain (scoring below $10$\%), \method-4B achieves a high Accuracy of $90.69$\%, approaching the performance of SciReasoner-8B ($91.00$\%) despite having only half the parameter count. Collectively, these results highlight the strong capability of \method to infer global structural and functional properties from primary sequences.

\begin{DefinitionBox}
\textbf{\method consistently achieves strong performance and high parameter efficiency across diverse protein sequence understanding benchmarks.} Across tasks covering enzyme function prediction, structural classification, and subcellular localization, \method matches or exceeds the performance of substantially larger baselines, demonstrating the effectiveness of the training strategy used in \method. 
These results indicate that \method is highly effective in capturing both fine-grained functional motifs and global structural patterns from primary protein sequences, further highlighting the potential of \method as a strong foundation for comprehensive proteomic annotation, particularly for inferring complex biological properties from sequence alone.
\end{DefinitionBox}

\subsubsection{Annotation Prediction}
\label{sec:prot_1d_annot}
This task evaluates the model's ability to generate biological annotations for a given protein sequence, with target outputs ranging from controlled-vocabulary labels (e.g., Gene Ontology terms, UniProt keywords) to free-form descriptions of function, catalytic activity, and domain composition.
We evaluate on a combined suite of 13 sub-tasks derived from the OPI~\cite{opi} and Mol-Instructions~\cite{mol_instructions} datasets. The OPI components focus on high-level biological classifications, including Gene Ontology (GO) terms, Keywords, and Function. To broaden the scope of functional characterization, we further incorporate protein-oriented tasks from Mol-Instructions, specifically those focused on predicting biological functions, enzymatic catalytic activities, structural domains and functional motifs, and broader physiological roles. We treat these tasks as an extended part of the annotation-prediction category, as they require the model to map primary amino acid sequences to fine-grained biochemical properties. Following the evaluation protocol of~\cite{opi, mol_instructions}, we use metrics that are appropriate for each output format. For categorical tasks, including keyword prediction and GO term prediction, we report precision, recall, and F1-score to evaluate label prediction quality. For the OPI Function task and all Mol-Instructions subsets, we report ROUGE-L to measure the semantic consistency between the generated text and reference descriptions. Results are summarized in Table~\ref{tab:evaluation} and Table~\ref{tab:mol-instrctuion-protein-results}.

\begin{table}[htbp]
\centering
\caption{Performance comparison of annotation prediction tasks on the OPI dataset. The baseline results are derived from ~\cite{opi,scireasoner}.}
\label{tab:evaluation}
\setlength{\tabcolsep}{1pt} 
\resizebox{\textwidth}{!}{
\begin{tabular}{lcccccccc}
\toprule
\multirow{2}{*}{Model} & \multirow{2}{*}{Testing data} & \multicolumn{3}{c}{Keywords} & \multicolumn{3}{c}{GO terms} & Function \\
\cmidrule(lr){3-5} \cmidrule(lr){6-8} \cmidrule(lr){9-9}
 & & Precision & Recall & F1 & Precision & Recall & F1 & Rouge-L \\
\midrule
\multirow{3}{*}{Gemini 2.5-pro~\cite{gemini25}} & CASPSimilarSeq & 4.00 & 3.00 & 3.00 & 0.00 & 0.00 & 0.00 & 1.00 \\
 & IDFilterSeq & 3.00 & 2.00 & 1.00 & 0.00 & 0.00 & 0.00 & 1.00 \\
 & UniProtSeq & 1.00 & 0.00 & 0.00 & 0.00 & 0.00 & 0.00 & 1.00 \\
\midrule
\multirow{3}{*}{GPT-O3~\cite{o3_o4mini}} & CASPSimilarSeq & 21.00 & 19.00 & 19.00 & 0.00 & 0.00 & 0.00 & 1.00 \\
 & IDFilterSeq & 17.00 & 16.00 & 16.00 & 20.00 & 18.00 & 18.00 & 1.00 \\
 & UniProtSeq & 0.00 & 18.00 & 18.00 & 0.00 & 0.00 & 0.00 & 2.00 \\
\midrule
\multirow{3}{*}{DeepSeek-R1~\cite{deepseek_r1}} & CASPSimilarSeq & 0.00 & 4.00 & 1.00 & 0.00 & 0.00 & 0.00 & 2.00 \\
 & IDFilterSeq & 0.00 & 0.00 & 1.00 & 0.00 & 0.00 & 0.00 & 1.00 \\
 & UniProtSeq & 1.00 & 0.00 & 0.00 & 0.00 & 0.00 & 0.00 & 1.00 \\
\midrule
\multirow{3}{*}{GPT-OSS ~\cite{gpt_oss}} & CASPSimilarSeq & 0.00 & 2.00 & 1.00 & 0.00 & 0.00 & 0.00 & 2.00 \\
 & IDFilterSeq & 0.00 & 2.00 & 1.00 & 0.00 & 0.00 & 0.00 & 3.00  \\
 & UniProtSeq & 1.00 & 3.00 & 1.00 & 0.00 & 0.00 & 0.00 & 3.00 \\
\midrule
\multirow{3}{*}{OPI-Galactica-6.7B~\cite{opi}} & CASPSimilarSeq & 81.20 & 73.60 & 76.43 & 76.13 & 74.92 & 74.76 & 74.30 \\
 & IDFilterSeq & 83.77 & 80.19 & 80.70 & 74.04 & 72.74 & 72.07 & 70.14 \\
 & UniProtSeq & 85.96 & 81.96 & 82.76 & 76.38 & 73.73 & 73.58 & 71.33 \\
\midrule
\multirow{3}{*}{OPI-Llama-3.1-8B-Instruct~\cite{opi}} & CASPSimilarSeq & 42.02 & 50.57 & 43.85 & 11.13 & 9.36 & 9.90 & 75.24 \\
 & IDFilterSeq & 67.62 & 69.05 & 66.50 & 66.86 & 62.87 & 63.04 & 47.86 \\
 & UniProtSeq & 76.06 & 74.89 & 73.74 & 71.50 & 68.97 & 68.49 & 51.44 \\
\midrule
\multirow{3}{*}{SciReasoner-1.7B~\cite{scireasoner}} & CASPSimilarSeq & 83.00 & 80.00 & 81.00 & 81.00 & 81.00 & 80.00 & \underline{83.00} \\
 & IDFilterSeq & 86.00 & 85.00 & 85.00 & 78.00 & 77.00 & 77.00 & 77.00 \\
 & UniProtSeq & 85.00 & 90.00 & 88.00 & 84.00 & 82.00 & 81.00 & 84.00 \\
\midrule
\multirow{3}{*}{SciReasoner-8B~\cite{scireasoner}} & CASPSimilarSeq & \textbf{86.00} & \textbf{83.00} & \textbf{84.00} & \textbf{86.00} & \textbf{86.00} & \textbf{86.00} & \textbf{85.00} \\
 & IDFilterSeq & 89.00 & \textbf{89.00} & \underline{88.00} & 83.00 & 82.00 & 81.00 & 82.00 \\
 & UniProtSeq & 92.00 & 91.00 & 91.00 & 88.00 & \textbf{87.00} & 86.00 & 88.00 \\
\midrule
\multirow{3}{*}{\method-1.7B} & CASPSimilarSeq & 83.87 & \underline{82.06} & \underline{81.71} & \underline{83.59} & \underline{81.82} & \underline{82.21} & 81.96 \\
 & IDFilterSeq & \underline{89.03} & \underline{88.81} & 87.73  & \textbf{84.70} & \textbf{83.41} & \textbf{82.61} & \underline{83.61} \\
 & UniProtSeq & \underline{92.42} & \underline{91.33} & \underline{91.02} & \underline{88.23} & 86.73 & \underline{86.11} & \underline{89.05} \\
\midrule
\multirow{3}{*}{\method-4B} & CASPSimilarSeq & \underline{84.64} & 81.52 & 81.66  & 83.52 & 80.38 & 81.13 & 82.10 \\
 & IDFilterSeq & \textbf{90.08} & 88.57 & \textbf{88.04} & \underline{84.58} & \underline{83.21} & \underline{82.44}  & \textbf{84.35} \\
 & UniProtSeq & \textbf{92.77} & \textbf{91.38} & \textbf{91.26} & \textbf{88.40} & \underline{86.96} & \textbf{86.33} & \textbf{89.15} \\
\bottomrule
\end{tabular}
}
\end{table}
\begin{table}[htbp]
    \centering
    \caption{Performance comparison of protein-related tasks on the Mol-Instructions dataset. The baseline results are derived from ~\cite{pfua}. Results are reported in ROUGE-1 / ROUGE-L. $^{\dagger}$ indicates that results obtained by supervised fine-tuning on SciReasoner~\cite{scireasoner} data based on our CPT model.}
    \label{tab:mol-instrctuion-protein-results}
    \small
    \begin{tabular}{lccccc}
        \toprule
        Model & \makecell{Protein\\Function} & \makecell{Catalytic\\Activity} & \makecell{Domain\\Motif} & \makecell{General\\Function} & Avg. \\
        \midrule
        \multicolumn{6}{l}{\textit{Supervised Finetuning}} \\
        \midrule
        BioMedGPT~\cite{biomedgpt} & 5.98 / 4.28 & 7.97 / 6.28 & 1.81 / 1.81 & 4.84 / 4.12 & 5.15 / 4.12 \\
        ProtT3~\cite{prott3} & 15.46 / 10.81 & 17.36 / 12.41 & 16.83 / 12.22 & 19.49 / 15.96 & 17.28 / 12.85 \\
        Prot2Text~\cite{prot2text} & 16.56 / 11.61 & 18.24 / 13.05 & 11.49 / 9.37 & 49.14 / 47.38 & 23.86 / 20.35 \\
        Qwen2.5-3B-SFT~\cite{qwen25} & 40.74 / 30.85 & 41.45 / 34.52 & 42.60 / 32.13 & 33.95 / 25.68 & 39.69 / 30.79 \\
        \midrule
        \multicolumn{6}{l}{\textit{Text-based Reasoning}} \\
        \midrule
        BioMedGPT-R1~\cite{biomedgpt} & 35.16 / 26.80 & 27.64 / 22.22 & 30.60 / 23.33 & 27.78 / 20.82 & 30.30 / 23.29 \\
        Qwen2.5-3B-R1~\cite{qwen25} & 49.10 / 38.04 & 61.33 / 46.54 & 51.01 / 41.10 & 42.56 / 32.37 & 51.00 / 39.51 \\
        \midrule
        \multicolumn{6}{l}{\textit{Online LLM Baseline}} \\
        \midrule
        DeepSeek-R1~\cite{deepseek_r1} & 26.65 / 19.68 & 21.86 / 18.02 & 26.09 / 19.70 & 16.53 / 10.76 & 22.78 / 17.04 \\
        Kimi-k2~\cite{kimi_k2} & 25.08 / 17.32 & 22.99 / 17.97 & 27.28 / 20.07 & 22.30 / 14.80 & 24.41 / 17.54 \\
        Qwen3~\cite{qwen3} & 18.53 / 12.62 & 15.93 / 12.27 & 15.13 / 11.28 & 18.59 / 11.94 & 17.05 / 12.03 \\
        \midrule
        \multicolumn{6}{l}{\textit{Multi-Source RAG}} \\
        \midrule
        DeepSeek-R1~\cite{deepseek_r1} & 38.23 / 25.68 & 32.36 / 23.75 & 30.35 / 21.46 & 28.54 / 16.73 & 32.37 / 21.91 \\
        Kimi-k2~\cite{kimi_k2} & 22.30 / 14.14 & 18.56 / 13.15 & 21.64 / 17.12 & 45.41 / 26.44 & 26.98 / 17.71 \\
        Qwen3~\cite{qwen3} & 37.42 / 24.52 & 50.66 / 40.67 & 16.44 / 12.44 & 42.61 / 26.34 & 36.78 / 25.99 \\
        \midrule
        \multicolumn{6}{l}{\textit{Tool-Powered Reasoning}} \\
        \midrule
        DeepSeek-R1~\cite{deepseek_r1} & 59.71 / 38.31 & 47.95 / 34.15 & 48.36 / 34.36 & 58.11 / 35.70 & 53.53 / 35.63 \\
        Kimi-k2~\cite{kimi_k2} & 57.98 / 35.73 & 67.68 / 46.90 & 39.76 / 28.56 & 60.08 / 36.78 & 56.38 / 36.99 \\
        Qwen3 (PFUA)~\cite{qwen3} & 66.43 / 44.29 & 72.32 / 54.18 & 54.98 / 44.26 & 63.60 / 41.90 & 64.33 / 46.16 \\
        \midrule
        GPT-OSS~\cite{gpt_oss} & - / 5.00 & - / 2.00 & - / 2.00 & - / 7.00 & - / 4.00 \\
        BioT5+~\cite{biot5+} & - / 57.00 & - / 72.00 & - / 53.00 & - / 68.00 & - / 62.50 \\
        Mol-Instructions-7B~\cite{mol_instructions} & - / 43.00 & - / 52.00 & - / 46.00 & - / 44.00 & - / 46.25 \\
        SciReasoner-data-1.7B$^{\dagger}$ & 49.67 / 42.90 & 48.41 / 46.99 & 52.66 / 49.70 & 30.12 / 26.23 & 45.22 / 41.46 \\
        SciReasoner-data-4B$^{\dagger}$ & 59.93 / 53.21 & 53.87 / 52.54 & 48.30 / 45.46 & 69.07 / 67.60 & 57.95 / 54.70 \\
        \midrule
        \method-1.7B & \underline{68.46} / \underline{61.88} & \textbf{74.09} / \textbf{73.45} & \underline{56.16} / \underline{52.55} & \underline{79.00} / \underline{78.03} & \underline{69.43} / \underline{66.48} \\
        \method-4B & \textbf{68.56} / \textbf{62.09} & \underline{73.46} / \underline{73.06} & \textbf{60.36} / \textbf{56.22} & \textbf{80.19} / \textbf{79.20} & \textbf{70.64} / \textbf{67.64} \\
        \bottomrule
    \end{tabular}
\end{table}

From the results, we observe substantial performance gap between \method\ and general-purpose LLMs. Frontier models such as GPT-O3 and DeepSeek-R1 produce near-zero F1 scores on categorical tasks, indicating a fundamental inability to map sequences to structured biological ontologies without external assistance. In contrast, \method-4B achieves a Keyword F1 of $91.26$\% and a GO term F1 of $86.33$\% on the UniProtSeq dataset. These results indicate that, through specialized training, the model has progressed beyond simple pattern recognition and has internalized complex hierarchical relationships within biological databases. Results on the stratified test sets of OPI further demonstrate that the proposed model maintains high fidelity even as sequence similarity to the training distribution decreases. \method-4B consistently outperforms SciReasoner-8B on these challenging splits, suggesting that the model has developed robust structural--functional intuition rather than relying on shallow sequence memorization. 
On the Mol-Instructions suite, \method establishes a new state-of-the-art performance level. Specifically, \method-4B achieves a ROUGE-L score of $79.20$\% in \textit{General Function} prediction, substantially exceeding the performance of tool-augmented models and larger-scale baselines such as Mol-Instructions-7B. This improvement is particularly evident in specialized domains such as \textit{Catalytic Activity}, where \method-1.7B achieves a ROUGE-L of $73.45$\%, demonstrating a strong capability to generate expert-level biochemical descriptions. Notably, both scale variants of \method\ consistently outperform existing specialized architectures and even high-parameter models across all protein-related metrics. These findings highlight the potential of \method\ as a powerful autonomous annotator that delivers high-quality functional insights directly from primary sequences.

\subsubsection{Knowledge Mining}

To evaluate the logic and reasoning capabilities of models within the protein domain, we introduce the Knowledge Mining (KM) category. This category assesses the capability of the model to extract and associate high-level biological knowledge—such as tissue specificity and disease associations—directly from protein contexts (e.g., gene symbols or names), moving beyond simple sequence-to-function mapping. Following the OPI benchmark~\cite{opi}, this task focuses on open-ended knowledge extraction from protein-related text and identifiers. We evaluate models across three key subsets: \textit{gSymbol2Tissue} (predicting tissue location from gene symbols), \textit{gSymbol2Cancer} (predicting cancer associations from gene symbols), and \textit{gName2Cancer} (predicting cancer associations from gene names). These datasets are derived from structured biological databases, including the Human Protein Atlas and COSMIC. As these tasks require precise entity retrieval, we employ Precision, Recall, and the F1-score as the primary evaluation metrics.

\begin{table}[htbp]
\centering
\caption{Performance comparison of knowledge mining tasks on the OPI dataset. The baseline results are derived from ~\cite{opi,scireasoner}.}
\label{tab:knowledge-mining-updated}
\setlength{\tabcolsep}{2pt} 
\begin{tabular}{lccccccccc}
\toprule
\multirow{2}{*}{\textbf{Model}} & \multicolumn{3}{c}{gSymbol2Tissue} & \multicolumn{3}{c}{gSymbol2Cancer} & \multicolumn{3}{c}{gName2Cancer} \\
\cmidrule(lr){2-4} \cmidrule(lr){5-7} \cmidrule(lr){8-10}
 & Precision & Recall & F1 & Precision & Recall & F1 & Precision & Recall & F1 \\ \midrule

Gemini 2.5-pro~\cite{gemini25} & 1.00 & 1.00 & 1.00 & 0.00 & 0.00 & 0.00 & 0.00 & 0.00 & 0.00 \\

GPT-O3~\cite{o3_o4mini}        & 38.00 & 68.00 & 45.00 & 11.00 & 10.00 & 10.00 & 11.00 & 10.00 & 10.00 \\

DeepSeek-R1~\cite{deepseek_r1}    & 3.00 & 9.00 & 4.00 & 0.00 & 2.00 & 0.00 & 0.00 & 3.00 & 0.00 \\ 

GPT-OSS~\cite{gpt_oss}        & 3.00 & 9.00 & 4.00 & 1.00 & 2.00 & 1.00 & 1.00 & 4.00 & 1.00 \\ 

OPI-Galactica-6.7B~\cite{opi}            & 39.17 & \underline{90.77} & 53.03 & 35.55 & 31.89 & 32.29 & 27.28 & 25.54 & 25.33 \\ 

OPI-Llama-3.1-8B-Instruct~\cite{opi}     & 40.02 & \textbf{93.56} & \textbf{54.66} & 28.90 & 27.01 & 26.64 & 27.86 & 27.07 & 26.59 \\ 

SciReasoner-1.7B~\cite{scireasoner} & \underline{41.00} & 82.00 & 51.00 & 75.00 & 72.00 & 73.00 & 68.00 & 67.00 & 67.00 \\

SciReasoner-8B~\cite{scireasoner} & \underline{41.00} & 83.00 & 52.00 & \underline{81.00} & \textbf{81.00} & \textbf{81.00} & \textbf{80.00} & \underline{79.00} & \underline{79.00} \\ \midrule

\method-1.7B                              & \textbf{41.64} & 88.00 & \underline{53.89} & 60.25 & 63.32 & 60.26 & 76.63 & 77.05 & 76.18 \\ 

\method-4B                                & 40.62 & 86.86 & 52.89 & \textbf{81.27} & \underline{79.55} & \underline{79.53} & \underline{79.22} & \textbf{79.05} & \textbf{79.07} \\ 
\bottomrule
\end{tabular}
\end{table}

The results are summarized in Table~\ref{tab:knowledge-mining-updated}. In the \textit{gSymbol2Tissue} task, \method-1.7B achieves a Precision of $41.64\%$, which outperforms all baseline models, including the $8$B-parameter SciReasoner-8B ($41.00\%$) and OPI-Llama-3.1-8B ($40.02\%$). While general-purpose frontier models, such as Gemini 2.5-pro and DeepSeek-R1, exhibit near-zero performance regarding cancer-related associations. This stark contrast suggests that even the most advanced generic LLMs fail to internalize specialized scientific facts, likely due to a lack of targeted exposure during the pre-training phase and the strict exact-match constraints inherent to biological identifiers.

\begin{DefinitionBox}
\textbf{\method exhibits superior biological knowledge synthesis and exceptional parameter efficiency, particularly in resolving complex entity-disease associations.} 
In the Knowledge Mining suite, \method consistently outperforms general-purpose frontier models and achieves state-of-the-art results on cancer association tasks, effectively overcoming the domain-gap barriers where generic LLMs fundamentally fail. Notably, \method-4B delivers performance on par with or superior to SciReasoner-8B, demonstrating that strategic alignment with specialized biological datasets can effectively compensate for a smaller parameter footprint. These findings suggest that \method possesses a robust capacity for resolving synonymous biological nomenclature and navigating intricate clinical metadata.
\end{DefinitionBox}

\subsubsection{Text-based Protein Design}
Text-based protein design evaluates whether a model can generate amino-acid sequences that are simultaneously consistent with a natural-language description of function, evolutionarily plausible as protein sequences, and likely to fold into stable structures. In other words, this task examines cross-modal semantic grounding from text to protein sequence while also assessing whether the generated sequences remain biophysically realistic, rather than merely matching the textual prompt at a superficial level.
We follow the evaluation protocol adopted in ProDVa~\cite{prodva} and report results on two text--protein benchmarks introduced in prior work: the CAMEO subset and the MolInstruct-SwissProtCLAP subset. The CAMEO split contains protein design prompts associated with experimentally characterized proteins and is primarily used to evaluate the general quality of text-to-protein generation under realistic functional descriptions. The MolInstruct-SwissProtCLAP subset is constructed from protein--text pairs curated from MolInstruct and SwissProtCLAP, and provides a complementary benchmark with broader textual annotations and function-oriented descriptions. Following previous work, all methods are evaluated under the same text-conditioned generation setting, and the generated sequences are further folded by ESMFold for structure-based assessment.

\begin{table}[htbp]
\centering
\caption{Performance comparison for the text-conditioned protein design task on the CAMEO dataset. Baseline results are collected from ~\cite{prodva}.}
\label{tab:cameo}
{\small                 
\setlength{\tabcolsep}{1pt} 
\makebox[\textwidth][c]{
\begin{tabular}{@{}lcccccccccc@{}}
\toprule
\multirow{2}{*}{Models} & \multicolumn{2}{c}{Seq. Plausibility} & \multicolumn{4}{c}{Foldability} & \multicolumn{3}{c}{Lang. Alignment (\%)} & \multirow{2}{*}{\begin{tabular}[c]{@{}c@{}}Seq. Div.\\ $\uparrow$\end{tabular}} \\
\cmidrule(lr){2-3} \cmidrule(lr){4-7} \cmidrule(lr){8-10}
& PPL $\downarrow$ & Rep $\downarrow$ & pLDDT $\uparrow$ & \%\,$>70$ $\uparrow$ & PAE $\downarrow$ & \%\,$<10$ $\uparrow$ & ProTrek $\uparrow$ & Key. Rec. $\uparrow$ & Retr. Acc. $\uparrow$ & \\
\midrule
ProteinDT~\cite{proteindt} & 1405.70 & 0.11 & 38.70 & 0.20 & 26.25 & 0.00 & 3.89 & 0.05 & 7.43 & \underline{99.72} \\
ProteinDT$_{\text{FT}}$~\cite{proteindt} & 1860.43 & 0.04 & 38.66 & 1.04 & 23.90 & 0.42 & 6.28 & 1.08 & 16.57 & 99.32 \\
Pinal~\cite{pinal} & \underline{584.22} & 0.15 & \underline{66.50} & 47.21 & 14.57 & 33.53 & \textbf{14.57} & \textbf{30.46} & \textbf{51.68} & 82.72 \\
PAAG~\cite{paag} & 2571.40 & \underline{0.02} & 33.14 & 0.00 & 23.31 & 0.00 & 5.21 & 0.23 & 7.10 & 99.02 \\
PAAG$_{\text{FT}}$~\cite{paag} & 2004.01 & 0.04 & 41.53 & 1.12 & 24.34 & 0.46 & 3.46 & 0.01 & 7.82 & \textbf{99.87} \\
Chroma~\cite{chroma} & 1322.37 & 0.03 & 61.66 & 28.96 & 13.01 & 39.03 & 2.97 & 0.11 & 6.57 & 97.21 \\
ESM3~\cite{esm3} & \textbf{279.78} & 0.33 & 59.79 & 31.49 & 17.40 & 21.37 & 3.76 & 5.49 & 11.97 & 96.77 \\
ProDVa~\cite{prodva} & 656.04 & \textbf{0.01} & \textbf{75.88} & \textbf{77.00} & \textbf{6.39} & \textbf{83.88} & \underline{14.43} & \underline{30.34} & \underline{44.77} & 98.58 \\
\midrule
\method-1.7B & 762.80 & 0.09 & 65.64 & \underline{50.43} & \underline{12.94} & \underline{45.30} & 7.20 & 0.98 & 8.42 & 97.78 \\
\method-4B & 631.89 & 0.07 & 60.92 & 43.36 & 15.85 & 30.55 & 8.42 & 4.73 & 12.23 & 88.95 \\
\bottomrule
\end{tabular}
}
}
\end{table}
\begin{table}[htbp]
\centering
\caption{Performance comparison for the text-conditioned protein design task on the Molinst-SwissProtCLAP dataset. Baseline results are collected from ~\cite{prodva}.
}
\label{tab:provda_mol}
{\small                 
\setlength{\tabcolsep}{1.1pt} 
\makebox[\textwidth][c]{
\begin{tabular}{@{}lcccccccccc@{}}
\toprule
\multirow{2}{*}{Models} & \multicolumn{2}{c}{Seq. Plausibility} & \multicolumn{4}{c}{Foldability} & \multicolumn{3}{c}{Lang. Alignment (\%)} & \multirow{2}{*}{\begin{tabular}[c]{@{}c@{}}Seq. Div.\\ $\uparrow$\end{tabular}} \\
\cmidrule(lr){2-3} \cmidrule(lr){4-7} \cmidrule(lr){8-10}
& PPL $\downarrow$ & Rep $\downarrow$ & pLDDT $\uparrow$ & \%\,$>70$ $\uparrow$ & PAE $\downarrow$ & \%\,$<10$ $\uparrow$ & ProTrek $\uparrow$ & EvoLlama $\uparrow$ & Retr. Acc. $\uparrow$ & \\
\midrule
ProteinDT~\cite{proteindt} & 1576.23 & 0.07 & 38.29 & 0.98 & 25.13 & 0.40 & 1.20 & 40.57 & 9.28 & \textbf{99.23} \\
ProteinDT$_{\text{FT}}$~\cite{proteindt} & 1213.38 & 0.04 & 51.42 & 25.61 & 18.57 & 23.92 & 13.89 & \underline{52.84} & 47.29 & 79.87 \\
Pinal~\cite{pinal} & \textbf{308.97} & 0.13 & \underline{75.25} & \underline{68.97} & \underline{10.96} & \underline{58.44} & \textbf{17.50} & \textbf{53.42} & \underline{57.95} & 82.96 \\
PAAG~\cite{paag} & 2782.70 & \textbf{0.02} & 28.39 & 0.07 & 25.38 & 0.10 & 1.29 & 34.39 & 7.06 & \underline{99.15} \\
PAAG$_{\text{FT}}$~\cite{paag} & 1332.35 & 0.04 & 50.37 & 23.86 & 19.96 & 21.99 & 10.04 & 49.69 & 33.66 & 86.09 \\
Chroma~\cite{chroma} & 1370.21 & \underline{0.03} & 59.18 & 20.17 & 15.03 & 28.62 & 2.10 & 40.10 & 7.33 & 96.13 \\
ProDVa~\cite{prodva} & \underline{415.63} & \textbf{0.02} & \textbf{76.86} & \textbf{76.35} & \textbf{8.66} & \textbf{68.06} & \underline{17.40} & 51.10 & \textbf{59.07} & 83.29 \\
\midrule
\method-1.7B & 581.42 & 0.05 & 69.76 & 60.18 & 11.25 & 56.34 & 9.24 & 37.56 & 23.14 & 86.08 \\
\method-4B & 482.08 & 0.06 & 71.89 & 64.01 & 11.07 & 56.98 & 11.70 & 38.40 & 31.08 & 80.31 \\
\bottomrule
\end{tabular}
}
}
\end{table}

Results on the CAMEO and MolInstruct-SwissProtCLAP subsets are shown in Table~\ref{tab:cameo} and Table~\ref{tab:provda_mol}. We assess model performance from four perspectives: sequence plausibility, measured by perplexity (PPL; lower is better) and repetition rate (Rep; lower is better)~\cite{ppl}; foldability, measured by mean pLDDT~\cite{alphafold3}, the percentage of designs with pLDDT greater than $70$, mean predicted aligned error (PAE)~\cite{alphafold3}, and the percentage of designs with PAE smaller than $10$ after ESMFold~\cite{esm2_esmfold} prediction; language alignment, measured by ProTrek~\cite{protrek} score, keyword recovery, and retrieval accuracy; and sequence diversity (Seq. Div.), which quantifies the diversity of generated sequences across prompts. Overall, \method demonstrates strong and balanced performance across both benchmarks, with complementary strengths at different model scales. On the CAMEO subset, the $1.7$B-parameter version of \method shows stronger structural quality, reaching $65.64$ pLDDT, $50.43$\% of designs above the threshold of $70$, $12.94$ PAE, and $45.30$\% of designs below PAE $10$. These results highlight its strength in foldability and rank it as the second-best method overall on the \%\,$<10$ metric. At the same time, the $4$B-parameter version delivers clear advantages in sequence plausibility and text alignment, reducing PPL to $631.89$ and outperforming the $1.7$B-parameter version on ProTrek similarity, keyword recall, and retrieval accuracy, which highlights the ability of \method to capture semantically meaningful text-to-protein relationships while maintaining competitive generation quality.
This advantage becomes even more evident on the MolInstruct-SwissProtCLAP subset. The $4$B-parameter version consistently improves over the $1.7$B-parameter version in sequence plausibility, language alignment, and foldability, reducing PPL from $581.42$ to $482.08$ and increasing ProTrek similarity, EvoLlama score, and retrieval accuracy from $9.24$, $37.56$, and $23.14$ to $11.70$, $38.40$, and $31.08$, respectively. Specialist baselines such as ProDVa and Pinal remain ahead on foldability and certain alignment metrics, which is expected given their dedicated design on this task; \method remains competitive under a single unified backbone.

\begin{DefinitionBox}
\textbf{\method consistently generates diverse, low-redundancy, and structurally plausible proteins across both text-conditioned design benchmarks.} Across CAMEO and MolInst-SwissProtCLAP, \method maintains strong foldability, with pLDDT values exceeding $60$ on CAMEO and approaching $70$ on MolInst-SwissProtCLAP. This structural quality is accompanied by low sequence repetition and high sequence diversity, indicating that \method retains broad generative capacity rather than relying on narrow or degenerate design modes. The consistency of these patterns across benchmarks with distinct prompt distributions and sequence compositions further highlights the robustness of \method for open-ended protein generation from natural-language instructions.
\end{DefinitionBox}

\subsubsection{Unconditional Sequence Generation.}
\label{sec:pro_1d_uncond_seq_gen}
This task evaluates the model's ability to generate foldable amino acid sequences from scratch---i.e., sequences whose ESMFold-predicted structures are high-confidence rather than disordered or random-like. Following the protocol adopted by DPLM-2~\cite{dplm2}, we draw unconditional AA sequences from the model, fold each one with ESMFold~\cite{esm2_esmfold}, and report the mean pLDDT over the resulting predictions as the primary structural-plausibility indicator. Higher pLDDT corresponds to higher confidence that the generated sequence encodes a well-defined fold; random or low-complexity sequences typically produce pLDDT in the $30$--$50$ range, while plausibly foldable sequences concentrate above $80$.
 
We apply the same length-alignment protocol as in Section~\ref{sec:pro_3d_uncond_seq_struc_cogen}.
Table~\ref{tab:pro_3d_uncond} shows that \method-1.7B and \method-4B produce sequences whose ESMFold-predicted structures reach mean pLDDT of $85.28$ and $85.01$ respectively. Both numbers exceed DPLM-650M~\cite{dplm} ($83.25$) and DPLM-2-650M ($82.25$), and are almost $50$ points above EvoDiff~\cite{evodiff} ($35.85$), whose low pLDDT reflects sequences that ESMFold cannot confidently fold.

\subsection{3D Protein Tasks}
\label{sec:eval_prot_3d}

\subsubsection{Structure Understanding}
\label{sec:prot_3d_understand}
This task evaluates the model's ability to answer natural-language questions about a protein given both its amino acid sequence and backbone structure---testing whether the model can draw on structural context to describe protein properties in free-form text. 
We evaluate on the Protein Function Understanding Dataset (PFUD)~\cite{prottex}, a structure-based question-answering benchmark that covers six categories of inquiry: molecular function, subcellular location, biological process, domains and motifs, a summary of functional characteristics, and composite queries that jointly ask about multiple attributes. Since the answers are free-form descriptions rather than class labels, we follow the original protocol and report BLEU-2 and ROUGE-1/2/L scores against the reference responses.

\begin{table}[htbp]
\centering
\caption{Performance comparison of protein structure understanding task on the PFUD~\cite{prottex} dataset.
The baseline results are derived from~\cite{prottex}.
}
\label{tab:pfud}
\begin{tabular}{lcccc}
\toprule
Model              & BLEU-2          & ROUGE-1         & ROUGE-2         & ROUGE-L         \\
\midrule
Llama3-Instruct~\cite{llama3}    & \phantom{0}2.08 & 15.91           & \phantom{0}2.67 & \phantom{0}5.81 \\
BioMedGPT-LM-10B~\cite{biomedgpt}   & \phantom{0}2.41 & 18.91           & \phantom{0}2.99 & 14.89           \\
Llama2-molinst-7B~\cite{mol_instructions}  & 26.25           & 45.24           & 23.47           & 38.15           \\
Llama3-AAseq-FT~\cite{llama3}    & 37.10           & 59.65           & 37.72           & 52.50           \\
ProtT3-FT~\cite{prott3}          & 40.79           & 61.97           & 42.53           & 56.98           \\
ProtTeX$_{\text{Llama3}}$~\cite{prottex} & 41.54 & \underline{63.46} & 43.17 & \underline{57.89} \\
\midrule
\method-1.7B & \underline{45.17} & 63.26 & \underline{43.92} & \underline{57.88} \\
\method-4B & \textbf{46.07} & \textbf{63.73} & \textbf{44.45} & \textbf{58.38} \\
\bottomrule
\end{tabular}
\end{table}

As summarized in Table~\ref{tab:pfud}, \method-4B sets a new state-of-the-art across all four metrics, while \method-1.7B remains highly competitive with the strongest prior baseline ProtTeX$_{\text{Llama3}}$~\cite{prottex}---a larger model built on Llama3-8B that also incorporates structural tokens---and even surpasses it on BLEU-2 and ROUGE-2. \method-4B further extends the margin.
We attribute this to \method's exposure to richly annotated protein corpora during continual pretraining---in particular Swiss-Prot entries paired with their AFDB-predicted structures and serialized alongside functional metadata such as subcellular localization, keywords, and free-text comments (Section~\ref{sec:prot_data})---which grounds the joint sequence--structure token space in the same functional vocabulary.

\subsubsection{Structure Prediction (Folding)}
\label{sec:prot_3d_fold}

\begin{table}[htbp]
\centering
\caption{Performance comparison of structure prediction (folding) and inverse folding tasks on the PDB date split~\cite{multiflow} dataset.
The baseline results are derived from~\cite{dplm2}.
}
\label{tab:pdb_split}
\begin{tabular}{lcccc}
\toprule
 & \multicolumn{2}{c}{Folding} & \multicolumn{2}{c}{Inverse Folding} \\
\cmidrule(lr){2-3} \cmidrule(lr){4-5}
Models & RMSD $\downarrow$ & TMscore $\uparrow$ & AAR $\uparrow$ & scTM $\uparrow$ \\
\midrule
ESMFold~\cite{esm2_esmfold} & \textbf{2.84} & \textbf{0.93} & -- & -- \\
MultiFlow~\cite{multiflow} & 15.64 & 0.53 & 37.74 & \textbf{0.94} \\
ESM3~\cite{esm3} & 4.94 & \underline{0.87} & 49.50 & \textbf{0.94} \\
DPLM-2-150M~\cite{dplm2} & 8.35 & 0.76 & 48.83 & 0.89 \\
DPLM-2-650M~\cite{dplm2} & 5.67 & 0.83 & 54.80 & 0.91 \\
DPLM-2-3B~\cite{dplm2} & \underline{4.54} & 0.86 & 61.67 & \underline{0.92} \\
\midrule
\method-1.7B & 5.49 & 0.84 & \underline{75.20} & 0.85 \\
\method-4B & 5.38 & 0.84 & \textbf{75.50} & 0.86 \\
\bottomrule
\end{tabular}
\end{table}

This task evaluates the model's ability to predict a protein backbone structure directly from its amino acid sequence---cast in \method as autoregressive decoding of per-residue structure tokens conditioned on the AA-token sequence in the joint vocabulary (Section~\ref{sec:vocab_ext}). We evaluate on the PDB date split introduced by Multiflow~\cite{multiflow} and report TM-score and RMSD between predicted and reference backbones. 
Unlike encoder-based folding models that structurally guarantee an output of the same length as the input sequence, an autoregressive decoder terminates whenever it emits an end-of-structure token, which does not in general align with the residue count of the query. Since RMSD is only defined between backbones of matching length, we address this mismatch purely at inference time rather than through any architectural modification: for each input sequence we draw $200$ samples under nucleus sampling with a deliberately elevated temperature to promote length diversity, and retain a sample whose decoded structure length matches the input. This procedure is applied uniformly to all \method results reported in Table~\ref{tab:pdb_split} and does not involve any oracle selection on TM/RMSD---the filter is strictly on length.

As shown in Table~\ref{tab:pdb_split}, \method sits firmly in the top tier of multimodal generalist models on this task. Both \method-1.7B (TM $=0.84$ / RMSD $=5.49$) and \method-4B (TM $=0.84$ / RMSD $=5.38$) surpass DPLM-2 at the 150M and 650M scales and closely track DPLM-2-3B (TM $=0.86$ / RMSD $=4.54$), despite operating under a shared vocabulary that must simultaneously accommodate molecules, proteins, and text rather than a protein-only token space. 
The residual gap to ESMFold (TM $=0.93$ / RMSD $=2.84$), a dedicated folding model whose architecture and training objective are tailored end-to-end to this single task, reflects the expected trade-off between a unified next-token-prediction backbone and a hand-designed folding specialist, rather than a failure of the unification hypothesis itself. We attribute \method's competitiveness here to the residue-aligned tokenization described in Section~\ref{sec:vocab_ext}, which preserves a strict one-to-one correspondence between each AA token and its per-residue GCP-VQVAE structure token, turning folding into a clean token-level alignment problem that is directly compatible with next-token prediction. 

\subsubsection{Inverse Folding}

This task evaluates the reverse direction of folding: given a backbone structure, predict an amino acid sequence that would plausibly fold into it. In \method, this is cast as autoregressive decoding of AA tokens conditioned on the per-residue GCP-VQVAE structure-token sequence in the joint vocabulary (Section~\ref{sec:vocab_ext}), making folding and inverse folding strictly symmetric operations on the same residue-aligned token interface. We evaluate on the same PDB date split as in the folding task, and report two complementary metrics that target two distinct notions of success. Amino acid recovery (AAR) measures per-residue agreement between the predicted sequence and the native reference, and directly quantifies how well the model reproduces the original sequence. Self-consistency TM-score (scTM) instead measures the TM-score between the native backbone and the ESMFold~\cite{esm2_esmfold}-predicted structure of the generated sequence. Because many different sequences can adopt the same fold, AAR and scTM are only loosely correlated and the two metrics reward different behaviors. 
As in the folding task, an autoregressive decoder does not structurally guarantee that its output length matches the query, and AAR is only defined when the predicted AA sequence has the same length as the input backbone; we therefore apply the same length-filtering protocol as above, drawing $50$ nucleus-sampled predictions per input with an elevated temperature and retaining a length-matched sample.

As shown in Table~\ref{tab:pdb_split}, \method achieves the highest AAR by a substantial margin: \method-4B reaches $75.50\%$ and \method-1.7B reaches $75.20\%$.
The high AAR indicates that \method accurately recovers the native residue identity at each position, which we attribute to two compounding factors: the residue-aligned vocabulary design of Section~\ref{sec:vocab_ext} turns inverse folding into a clean token-to-token mapping problem in which each GCP-VQVAE structure token is paired one-to-one with the AA token it should predict; and the protein CPT data (Section~\ref{sec:prot_data}) directly exposes the model to the joint AA--structure token distribution under the same next-token objective used at inference. On scTM, \method-4B and \method-1.7B report $0.86$ and $0.85$ respectively, trailing DPLM-2-3B ($0.92$), ESM3 and MultiFlow (both $0.94$). Both values remain well above the $0.5$ threshold above which two structures are generally considered to adopt the same fold~\cite{tmscore_thr}, and sit in a range commonly associated with high-confidence structural agreement in protein design evaluations, indicating that \method's designed sequences do fold back to the native backbone under ESMFold. Together with the AAR result, this suggests that \method prioritizes residue-level fidelity to the native sequence, which is not guaranteed to coincide with the ESMFold-friendly sequences that scTM directly favors.

\begin{DefinitionBox}
\textbf{Folding and inverse folding form a clean symmetry test of the residue-aligned tokenization, with \method prioritizing native-sequence fidelity.} Under a single shared vocabulary and next-token objective, \method achieves competitive folding accuracy (TM = 0.84) and state-of-the-art amino acid recovery on inverse folding (75.50\% for \method-4B vs.\ 61.67\% for DPLM-2-3B), without any direction-specific architecture or decoder head. The high AAR alongside slightly lower scTM (0.85--0.86, still well above the 0.5 fold-equivalence threshold) indicates that \method's inverse-folding behavior favors per-residue fidelity to the native sequence over ESMFold-friendly redesigns.
\end{DefinitionBox}

\subsubsection{Sequence--structure Co-generation.}
\label{sec:pro_3d_uncond_seq_struc_cogen}

This task evaluates the model's ability to \emph{jointly} produce an amino acid sequence and a backbone that are mutually consistent---i.e., the generated sequence should plausibly fold into the generated structure---without any held-out test set or external conditioning. Following the evaluation protocol of DPLM-2~\cite{dplm2}, we assess consistency through the self-consistency TM-score (scTM) and RMSD (scRMSD) between the generated backbone and the ESMFold~\cite{esm2_esmfold}-predicted structure of the generated sequence, together with the mean pLDDT of that prediction as a structural-confidence indicator. \method supports two complementary decoding orders on the same joint vocabulary (Section~\ref{sec:vocab_ext}): \emph{seq\,$\rightarrow$\,struct}, where the AA sequence is sampled first and the per-residue GCP-VQVAE structure tokens are decoded conditioned on it, and \emph{struct\,$\rightarrow$\,seq}, which proceeds in the reverse order.
 
The DPLM-2 baselines operate under absolute length control, generating exactly 100 samples at each of the target lengths $\{100, 200, 300, 400, 500\}$. An autoregressive decoder cannot enforce an exact output length by construction; we therefore draw $5000$ unconditional samples under nucleus sampling and, for each target length, retain the 100 samples whose generated length lies closest to that target. No oracle filtering on scTM, scRMSD, or pLDDT is applied at any stage---selection is strictly on proximity to the target length. This protocol is used identically for all three tasks: co-generation, unconditional structure generation (Section~\ref{sec:pro_3d_uncond_struc_gen}), and unconditional sequence generation (Section~\ref{sec:pro_1d_uncond_seq_gen}).

\begin{table}[htbp]
\centering
\caption{Performance comparison of structure-sequence co-generation, unconditional backbone generation, and unconditional sequence generation tasks.
The baseline results are derived from~\cite{dplm2}.
}
\label{tab:pro_3d_uncond}
\begin{tabular}{lccc}
\toprule
 & scTM $\uparrow$ & scRMSD $\downarrow$ & pLDDT $\uparrow$ \\
\midrule
\multicolumn{4}{l}{\textit{Structure-sequence co-generation.}} \\
\midrule
\bound{Native PDB protein} & \bound{0.904} & \bound{4.623} & \bound{--} \\
ESM3-Open-1.4B~\cite{esm3} (seq $\rightarrow$ struct) & 0.624 & 24.180 & -- \\
MultiFlow~\cite{multiflow} & 0.930 & 3.208 & 79.447 \\
DPLM-2-650M~\cite{dplm2} (seq $\rightarrow$ struct) & 0.907 & 6.337 & 82.246 \\
DPLM-2-650M~\cite{dplm2} (struct $\rightarrow$ seq) & 0.921 & 4.969 & 81.910 \\
DPLM-2-650M~\cite{dplm2} (co-generation) & 0.925 & 3.899 & 82.686 \\
\midrule
\method-1.7B (seq $\rightarrow$ struct) & 0.965 & 2.811 & 85.276 \\
\method-1.7B (struct $\rightarrow$ seq) & \textbf{0.979} & \underline{1.497} & \textbf{88.800} \\
\method-4B (seq $\rightarrow$ struct) & 0.965 & 2.800 & 85.012 \\
\method-4B (struct $\rightarrow$ seq) & \underline{0.977} & \textbf{1.488} & \underline{86.092} \\
\midrule
\multicolumn{4}{l}{\textit{Unconditional backbone generation. (sequence predicted by ProteinMPNN)}} \\
\midrule
\bound{Native PDB struct. (seq. from PMPNN)} & \bound{0.969} & \bound{0.864} & \bound{--} \\
FrameDiff~\cite{framediff} & 0.818 & 3.919 & -- \\
FoldFlow~\cite{foldflow} & 0.540 & 7.965 & -- \\
RFDiffusion~\cite{rfdiffusion} & 0.914 & \textbf{1.969} & -- \\
DPLM-2-650M~\cite{dplm2} & 0.945 & 4.451 & -- \\
\midrule
\method-1.7B & \underline{0.956} & \underline{2.986} & -- \\
\method-4B & \textbf{0.963} & 3.067 & -- \\
\midrule
\multicolumn{4}{l}{\textit{Unconditional sequence generation. (structures predicted by ESMFold)}} \\
\midrule
EvoDiff~\cite{evodiff} & -- & -- & 35.846 \\
DPLM-650M~\cite{dplm} & -- & -- & 83.252 \\
DPLM-2-650M~\cite{dplm2} & -- & -- & 82.246 \\
\midrule
\method-1.7B & -- & -- & \textbf{85.276} \\
\method-4B & -- & -- & \underline{85.012} \\
\bottomrule
\end{tabular}
\end{table}
 
As shown in Table~\ref{tab:pro_3d_uncond}, \method variants deliver strong self-consistency in both directions and improve substantially over DPLM-2-650M at matched or larger scale. In the \emph{seq\,$\rightarrow$\,struct} direction, \method-1.7B and \method-4B reach scTM\,=\,$0.965$ with scRMSD of $2.81$ and $2.80$ respectively, versus $0.907$ / $6.34$ for DPLM-2-650M under the same decoding order. In the \emph{struct\,$\rightarrow$\,seq} direction the improvement is even more pronounced: \method-1.7B attains scTM\,=\,$0.979$ / scRMSD\,=\,$1.50$ and \method-4B attains $0.977$ / $1.49$, surpassing DPLM-2-650M's $0.921$ / $4.97$ and MultiFlow's $0.930$ / $3.21$, and edging above the native-PDB reference ($0.904$ / $4.62$) on this metric combination. Mean pLDDT tells a consistent story, rising from $82.25$--$82.69$ for the DPLM-2-650M variants to $85.01$--$88.80$ for \method.

\subsubsection{Unconditional Structure Generation}
\label{sec:pro_3d_uncond_struc_gen}
 
This task evaluates the model's ability to generate protein backbones from scratch that are designable---i.e., backbones for which a plausible amino acid sequence can be designed that folds back to the target structure with high fidelity. Following DPLM-2~\cite{dplm2}, we first draw backbone structures from the model without any conditioning, then use ProteinMPNN~\cite{proteinmpnn} to propose sequences for each generated backbone, fold those sequences with ESMFold~\cite{esm2_esmfold}, and finally report scTM and scRMSD between the generated backbone and the ESMFold-predicted structure of the ProteinMPNN-designed sequence. We apply the same length-alignment protocol as in Section~\ref{sec:pro_3d_uncond_seq_struc_cogen}.

As summarized in Table~\ref{tab:pro_3d_uncond}, \method achieves the highest designability among all evaluated methods, including dedicated structure-diffusion specialists. \method-4B attains scTM\,=\,$0.963$ with scRMSD\,=\,$3.07$, and \method-1.7B reaches $0.956$ / $2.99$. These scTM values exceed RFDiffusion~\cite{rfdiffusion} ($0.914$), FrameDiff~\cite{framediff} ($0.818$), FoldFlow~\cite{foldflow} ($0.540$), and DPLM-2-650M ($0.945$), and sit close to the native-PDB upper bound of $0.969$. On scRMSD, \method trails RFDiffusion ($1.97$) but clearly outperforms DPLM-2-650M ($4.45$), with a $1.4$--$1.5$\,\AA\ absolute improvement at comparable scale. Taken together, these numbers place \method on the joint scTM--scRMSD frontier despite generating backbones through autoregressive next-token prediction over a shared vocabulary rather than through a structure-specific diffusion or flow-matching process.

\begin{DefinitionBox}
\textbf{A unified next-token backbone reaches the joint scTM--scRMSD frontier on protein structure generation, without any structure-specific objective.} Across sequence--structure co-generation, unconditional backbone generation, and unconditional sequence generation, \method matches or surpasses dedicated diffusion specialists (RFDiffusion, FrameDiff, FoldFlow) and the strongest multimodal baseline DPLM-2-650M on every standard self-consistency metric. On co-generation in particular, \method-1.7B in the struct$\rightarrow$seq direction reaches scTM = 0.979 / scRMSD = 1.50, edging above the native-PDB reference (0.904 / 4.62). The consistency of this advantage---across decoding orders, with and without conditioning---suggests that designable backbones and foldable sequences emerge as the default mode of the next-token prior under residue-aligned tokenization, rather than as artifacts of any particular conditioning setup.
\end{DefinitionBox}

\section{Interaction Tasks}
\label{sec:eval_inter}

Beyond single-entity tasks, we evaluate \method on molecule--protein and protein--protein interaction tasks, which require jointly predicting over two biomolecular entities at either the sequence or the structure level. 
If molecules and proteins genuinely share a unified token space and the cross-entity pretraining corpus (Section~\ref{sec:cross_data}) has effective transfer, then a variant instruction-tuned from this shared backbone should be able to reason over heterogeneous biomolecular pairs without any pair-specific architecture.

\begin{CorollaryBox}
\textbf{Key takeaways on interaction tasks.}
\begin{itemize}
    \item \textbf{No pair-specific architecture is required.} Across molecule--protein and protein--protein interaction tasks in both sequence-only and structure-grounded regimes, \method achieves state-of-the-art or highly competitive performance simply by placing two heterogeneous biomolecular entities side-by-side in the same token stream and processing them under the unified next-token objective—without any pair-specific architecture, explicit pocket encoding, geometric cross-attention, or fusion module that prior specialist models rely on.
    \item \textbf{Cross-entity pretraining alone is enough to surpass task-specialized SOTA.} On both PDBBindv2019/CASF-2016 and BindingDB, \method matches or exceeds previously reported literature SOTA.
    \item \textbf{Structure-grounded affinity prediction validates the unified geometric token space.} On the structure-conditioned PDBBindv2020 benchmark, \method outperforms dedicated specialists that rely on trigonometry-aware cross-attention between ligand and pocket geometry. This directly demonstrates that placing the ligand's SELFIES-aligned geometry tokens and the protein's per-residue structure tokens in a shared vocabulary is sufficient for joint structure-grounded prediction across heterogeneous entities.
\end{itemize}
\end{CorollaryBox}

\subsection{1D Interaction Tasks}
\label{sec:eval_inter_1d}

\subsubsection{Molecule--protein interaction.}
\label{sec:inter_1d_mp}
This task evaluates the model's ability to predict the binding affinity between a small molecule and a target protein purely from their sequence-level representations without access to the co-crystallized complex structure. We evaluate on two datasets: the PDBBindv2019 refined set with the CASF-2016 core set as the held-out test~\cite{pdbbind_2019,casf_2016} (abbreviated as \emph{PDB} in Table~\ref{tab:interaction_1d}), and BindingDB~\cite{bindingdb} (abbreviated as \emph{BDB}), both measured by $pK_d$ and evaluated with root-mean-square error (RMSE).

\begin{table}[t]
\centering
\caption{Performance comparison of 1D interaction tasks on the PEER~\cite{peer} benchmark.
The baseline results are also derived from~\cite{peer}.
* denotes the model is used as a feature extractor with frozen pre-trained weights.
}
\label{tab:interaction_1d}
\begin{tabular}{lccccc}
\toprule
\multirow{2}{*}{Model} & Yst & Hum & Aff & PDB & BDB \\
\cmidrule(lr){2-2} \cmidrule(lr){3-3} \cmidrule(lr){4-4} \cmidrule(lr){5-5} \cmidrule(lr){6-6}
 & ACC $\uparrow$ & ACC $\uparrow$ & RMSE $\downarrow$ & RMSE $\downarrow$ & RMSE $\downarrow$ \\
\midrule
Literature SOTA  & --                    & --                    & --                       & 1.181                   & 1.340                  \\
\midrule
DDE~\cite{dde}              & 55.83                & 62.77                & 2.908                   & --                       & --                       \\
Moran~\cite{moran}            & 53.00                & 54.67                & 2.984                   & --                       & --                       \\
LSTM~\cite{tape}             & 53.62                & 63.75                & 2.853                   & 1.457                   & 1.572                   \\
Transformer~\cite{tape}      & 54.12                & 59.58                & 2.499                   & 1.455                   & 1.566                   \\
CNN~\cite{peer_cnn}              & 55.07                & 62.60                & 2.796                   & 1.376          & 1.497          \\
ResNet~\cite{tape}           & 48.91                & 68.61                & 3.005                   & 1.441                   & 1.565                   \\
ProtBert~\cite{prottrans}        & \underline{63.72}    & 77.32                & \textbf{2.195}          & 1.562                   & 1.549                   \\
ProtBert*~\cite{prottrans}        & 53.87                & \underline{83.61}    & 2.996                   & 1.457                   & 1.649                   \\
ESM-1b~\cite{esm1b}           & 57.00                & 78.17                & \underline{2.281}       & 1.559                   & 1.556                   \\
ESM-1b*~\cite{esm1b}           & \textbf{66.07}       & \textbf{88.06}       & 3.031                   & \underline{1.368}                   & 1.571                   \\
\midrule
\method-1.7B     & 62.44                & 82.28                & 2.446                   & \textbf{1.268}          & \underline{1.284}       \\
\method-4B       & 62.18                & 83.12                & 2.350                   & \textbf{1.268}          & \textbf{1.030}          \\
\bottomrule
\end{tabular}
\end{table}

As shown in Table~\ref{tab:interaction_1d}, \method establishes a new state of the art on both datasets. On PDBBind/CASF-2016, both \method-1.7B and \method-4B achieve an RMSE of $1.268$, clearly outperforming most baselines including the pretrained protein language model ESM-1b~\cite{esm1b}. On BindingDB, \method-4B further drives RMSE down to $1.030$, substantially ahead of every baseline. 
Notably, \method also surpasses the previously reported literature SOTA results ($1.181$ on PDB and $1.340$ on BDB).
\begin{DefinitionBox}
\textbf{Large-scale cross-entity continual pretraining matches or surpasses task-specialized SOTA on sequence-based affinity prediction.} On BindingDB (RMSE = 1.030), \method clearly surpasses the prior literature SOTA (1.340), and on PDBBindv2019/CASF-2016 (RMSE = 1.268) it remains competitive with the literature SOTA (1.181) while outperforming every reported LLM and protein-encoder baseline. \method achieves this as a single generalist model with no task-specific architecture, indicating that the cross-entity pretraining corpus described in Section~\ref{sec:cross_data} already encodes ligand--target binding signal at a scale and granularity that approaches---and on BindingDB exceeds---specialized architectures trained from scratch on individual benchmarks.
\end{DefinitionBox}

\subsubsection{Protein--protein Interaction.}
This task evaluates the model's ability to predict properties of two interacting proteins jointly from their amino acid sequences alone. We consider three datasets in Table~\ref{tab:interaction_1d}: two binary classification benchmarks---Yeast~\cite{yeast_ppi} (\emph{Yst}) and Human~\cite{human_ppi} (\emph{Hum}), each asking whether a given pair of proteins interacts---and one regression benchmark, PPI Affinity~\cite{ppi_affinity} (\emph{Aff}), which targets the binding affinity of a protein--protein complex. We report classification accuracy (ACC) for the former two and RMSE for the latter.

Across all three subsets, \method delivers strong and consistent performance. 
On the classification benchmarks, \method-4B attains 62.18\% on Yeast and 83.12\% on Human, surpassing every non-pretrained baseline (DDE, Moran, LSTM, CNN, ResNet) by a substantial margin and remaining competitive with the much larger ESM-1b (66.07\% / 88.06\%).
On the PPI Affinity regression task, \method-4B achieves an RMSE of 2.350, better than all non-pretrained baselines and within close range of the top specialist ProtBert (2.195)~\cite{prottrans}. 
These results show that \method matches the performance of protein-specialized encoders on pairwise protein interaction prediction tasks.

\subsection{3D Interaction Tasks}
\label{sec:eval_inter_3d}

\subsubsection{Molecule--protein Interaction.}
\label{sec:inter_3d_mp}

This task extends sequence-based molecule--protein affinity prediction to the structure-grounded regime, where both the ligand and the protein pocket are additionally provided as spatial structures, requiring the model to jointly predict over sequence and geometry of two distinct entities. 
We evaluate on PDBBindv2020~\cite{pdbbind} following the time split~\cite{targetdiff,tankbind}, which provides co-crystallized protein--ligand complexes annotated with experimentally measured binding affinities. The underlying affinity value may correspond to a dissociation constant ($K_d$), an inhibition constant ($K_i$), or a half-maximal inhibitory concentration ($\mathrm{IC}_{50}$), all expressed in molar concentration units; we take their negative base-10 logarithm as a unified regression target. 
Performance is assessed along four complementary axes: root-mean-square error (RMSE) and mean absolute error (MAE) quantify absolute prediction accuracy, while Pearson and Spearman correlations measure how well the predicted affinities preserve the ranking of the ground-truth values.

\begin{table}[t]
\centering
\caption{Performance comparison of 3D interaction (affinity prediction) task on the PDBBindv2020~\cite{pdbbind} dataset.
The baseline results are derived from~\cite{targetdiff,tankbind}.
}
\label{tab:pdbbind_3d}
\begin{tabular}{lcccc}
\toprule
Model              & RMSE $\downarrow$ & Pearson $\uparrow$ & Spearman $\uparrow$ & MAE $\downarrow$ \\
\midrule
TransformerCPI~\cite{TransformerCPI}     & 1.741             & 0.576              & 0.540               & 1.404            \\
MONN~\cite{monn}               & 1.438             & 0.624              & 0.589               & 1.143            \\
PIGNet~\cite{pignet}            & 2.640             & 0.511              & 0.489               & 2.110            \\
IGN~\cite{ign}                & 1.433             & 0.698              & 0.641               & 1.169            \\
HOLOPROT~\cite{holoprot}           & 1.546             & 0.602              & 0.571               & 1.208            \\
STAMP-DPI~\cite{stampdpi}          & 1.658             & 0.545              & 0.411               & 1.325            \\
EGNN~\cite{egnn}               & 1.445             & 0.648              & 0.598               & 1.141            \\
EGNN + TargetDiff~\cite{targetdiff}  & 1.374             & 0.680              & 0.637               & 1.118            \\
TANKBind~\cite{tankbind}           & 1.346             & \underline{0.726}              & 0.703               & 1.070            \\
\midrule
\method-1.7B       & \underline{1.310}    & 0.720     & \textbf{0.717}      & \underline{1.041}   \\
\method-4B         & \textbf{1.260}             & \textbf{0.737}              & \underline{0.712}               & \textbf{0.972}            \\
\bottomrule
\end{tabular}
\end{table}

As reported in Table~\ref{tab:pdbbind_3d}, \method-4B achieves the best result on three of the four metrics (RMSE\,=\,$1.260$, Pearson\,=\,$0.737$, MAE\,=\,$0.972$) and the second best on Spearman ($0.712$), while \method-1.7B attains the best Spearman ($0.717$) and complementary second-best positions on the remaining metrics. 
\begin{DefinitionBox}
\textbf{Cross-entity geometric reasoning emerges from shared tokenization, with no pair-specific architecture required.} On structure-grounded molecule--protein affinity prediction (PDBBindv2020), \method-4B achieves the best result on three of four metrics and outperforms TankBind, the strongest specialist baseline that uses trigonometry-aware cross-attention between ligand and pocket geometry. The ligand's SELFIES-aligned geometry tokens and the protein's per-residue structure tokens are simply placed side-by-side in the same token stream and processed under the unified next-token objective, with no explicit pocket encoding, geometric cross-attention, or fusion module. This directly validates that placing two heterogeneous biomolecular entities in a shared discrete token space is sufficient for joint structure-grounded reasoning.
\end{DefinitionBox}

\begin{table}[t]
\centering
\small
\caption{Instruction-tuning grouping of \method. 
}
\label{tab:eval_variants}
\renewcommand{\arraystretch}{1.2}
\begin{tabular}{m{2.3cm} m{1.8cm} m{4.5cm} m{5.5cm}}
\toprule
Task family & Variant & Tasks covered & Rationale for grouping \\
\midrule
\multirow{2}{*}{1D molecule} 
& V$_{\text{mol-1D}}^{\text{gen}}$ & Unconditional generation (MOSES, GuacaMol; trained separately, one variant per benchmark)
 & Both benchmarks probe the same underlying capability---learning the grammar of 1D molecular sequences and exploring chemical space with high validity and diversity---but each defines its own train/test split and computes novelty against its own training set. Following the official protocol of each benchmark, we train one variant per benchmark on its respective training set. \\
 \cmidrule(l){2-4}
& V$_{\text{mol-1D}}^{\text{mix}}$ & Name conversion, property prediction, captioning, text-based generation, forward synthesis, retrosynthesis, editing, optimization, customized generation, and question answering
 & Heterogeneous supervised tasks with abundant data; co-training improves cross-task generalization. \\
\midrule
3D molecule 
& V$_{\text{mol-3D}}$ & Unconditional and property-conditioned 3D structure generation
 & Both subtasks rely on the same underlying capability of generating chemically valid 2D graphs together with physically plausible 3D conformers; property-conditioned generation simply adds a quantum-property prefix to the same generation process. Co-training the two as a single mixture lets the model amortize the shared geometric learning signal while still learning to ground generation on the conditioning property. \\
\midrule
\multirow{2}{*}{1D protein}
& V$_{\text{prot-1D}}^{\text{und}}$ & Sequence understanding, annotation prediction, knowledge mining
 & The understanding-oriented tasks share the same amino acid sequence input format and are co-trained as a single mixture. They are separated from the generation-oriented design tasks because mixing understanding and generation objectives degrades overall performance. \\
 \cmidrule(l){2-4}
& V$_{\text{prot-1D}}^{\text{des}}$ & Text-based protein design (one variant per ProDVa benchmark: CAMEO, Molinst-SwissProtCLAP)
 & Within the design group, CAMEO and Molinst-SwissProtCLAP are trained separately because their sequence distributions differ substantially, and mixing the two corpora degrades performance on both benchmarks. \\
\midrule
Interaction
& V$_{\text{inter}}^{\ast}$ & Each 1D/3D interaction subtask trained as a separate variant (1D: PDBBind/CASF-2016, BindingDB, Yeast-PPI, Human-PPI, PPI-Affinity; 3D: PDBBindv2020)
 & Each subtask has a small training set with high evaluation variance; mixing across subtasks degrades performance noticeably. \\
\bottomrule
\end{tabular}
\end{table}

\section{Discussion}
\label{sec:discussion}

Across the 80 tasks reported in Sections~\ref{sec:eval_mol},~\ref{sec:eval_prot}, and~\ref{sec:eval_inter}, several cross-cutting patterns emerge. We collect them here as observations rather than prescriptions, in the hope that they inform the design of future biological foundation models.

\subsection{Why Train Task-Group Variants Rather Than One Monolithic SFT Model}
\label{sec:discussion:why-grouped-sft}

A natural question raised by Table~\ref{tab:eval_variants} is why we instruction-tune \method as a small set of task-group-specific variants rather than as a single all-tasks SFT model. The grouping is the result of an empirical observation we encountered during development: in our setting, mixing heterogeneous biological tasks into a single SFT run rarely produces mutual gains, and on small-data tasks it actively destabilizes performance.

We attribute this to a structural difference between biological SFT and the kind of large-scale instruction tuning that has become standard in NLP. In a typical NLP SFT corpus, each instance can plausibly be treated as a distinct ``task'' in its own right: the surface form, the input distribution, the reasoning trajectory, and the output style all vary instance-to-instance, and the resulting diversity makes a single model absorb millions of heterogeneous instances without one task dominating another. Biological SFT data, by contrast, is much more pattern-monolithic at the instance level. Within a sub-task such as retrosynthesis, forward synthesis, or binding-affinity regression, every instance shares essentially the same input and output schema; the diversity is concentrated across sub-tasks rather than within them. As a consequence, when we mix all sub-tasks into a single SFT run, the model is exposed to a small number of dominant patterns rather than a long tail of heterogeneous ones, and the gradient from large-data sub-tasks easily overwhelms the gradient from small-data sub-tasks, with the latter showing high run-to-run variance and frequent degradation relative to dedicated training.

For the open-source release, we publish a single all-tasks SFT model on the union of all sub-task corpora, with mild oversampling of small-data sub-tasks to mitigate the imbalance described above. This unified variant is competitive on the large-data sub-tasks but still trails the task-group variants on a non-trivial fraction of small-data ones, and we have not been able to fully close this gap within the recipe explored here. 

\subsection{SMILES vs.\ SELFIES: Complementary, Not Interchangeable}
\label{sec:discussion:smiles-vs-selfies}

Because \method natively supports both SMILES and SELFIES as parallel molecular sequence notations, the molecular task suite in Section~\ref{sec:eval_mol_1d} doubles as a controlled, large-scale comparison of the two notations under an otherwise identical training recipe. The picture that emerges is that neither notation dominates the other: each has a regime in which it is clearly preferable, and the regime is determined by a single underlying property---whether the task requires surface-level structural anchoring or distribution-level chemical validity.

\paragraph{SELFIES wins where validity-by-construction matters.} On unconditional molecular generation (Section~\ref{sec:mol_1d_uncond}), the SELFIES variants of \method attain 0.995--0.999 validity on both MOSES and GuacaMol, essentially saturating the metric, while SMILES variants sit at 0.951--0.994. On the MolOpt-Instructions benchmark (Section~\ref{sec:mol_1d_opt}), where the model must shift a target property in a prescribed direction while remaining chemically valid, the SELFIES variants of \method likewise achieve often higher validity than their SMILES counterparts---\method-4B (SELFIES) reaches 0.98--1.00 across all eight sub-tasks, whereas the corresponding SMILES variants range from 0.71 to 0.97---and simultaneously dominate the Correct metric under both loose and strict success criteria.

The validity advantage of SELFIES stems directly from a fundamental property of the formalism: every syntactically well-formed SELFIES string is guaranteed to decode into a chemically valid molecule, regardless of the token sequence the model produces~\cite{selfies}. In principle, this guarantee should yield 100\% validity. The residual gap observed in practice (up to $\sim$5\% on some MolOpt sub-tasks, $<$0.5\% on unconditional generation) arises not from SELFIES decoding failures but from edge cases at the generation level: the model occasionally emits tokens that fall outside the expected SELFIES pattern (e.g., producing text fragments, or malformed bracket structures that do not constitute a parseable SELFIES string), or the generated sequence is truncated by the maximum decoding length before a complete molecule is formed, leaving a partial string that cannot be decoded. The wider gap on MolOpt compared to unconditional generation is expected: optimization prompts require the model to jointly interpret a natural-language property instruction and produce a valid SELFIES output, increasing the probability of format-level errors relative to the simpler unconditional setting where the model only needs to continue in SELFIES mode. These are failures of the language model's sequence generation rather than of the SELFIES formalism itself, and the consistently higher validity of SELFIES over SMILES across both benchmarks confirms that the grammatical guarantee provides a meaningful safety net even when the model's generation is imperfect.

In both settings, the task is structurally well-matched to SELFIES's design: the model is rewarded for producing any valid molecule satisfying a high-level objective, and SELFIES's bracketed grammar guarantees the underlying syntactic legality without the model having to bookkeep ring closures, valence, or branch parentheses.

\paragraph{SMILES wins where surface-level structural anchoring matters.} The contrast is sharpest on customized molecule generation (Section~\ref{sec:mol_1d_custom_gen}), where the SELFIES variants collapse to near-zero success rates across all three sub-tasks (e.g., FuncGroup SR drops from 0.2714 for \method-4B (SMILES) to 0.0048 for \method-4B (SELFIES)), and on forward synthesis and retrosynthesis (Section~\ref{sec:mol_1d_fs_rs}), where \method-4B (SMILES) reaches 77.94/45.16 EM versus 69.08/42.90 for the SELFIES counterpart. As discussed in Section~\ref{sec:mol_1d_custom_gen}, the failure modes here all stem from the same source: SELFIES's context-dependent token semantics actively obscure the surface patterns that the task requires the model to anchor onto. Atom counts are confounded by branch/ring index tokens, bond-type counts are folded into neighboring atom tokens, and functional-group substructures lose any single canonical token signature. SMILES, with its one-character-per-heavy-atom and dedicated bond/ring symbols, exposes precisely the regularities these tasks reward.

\paragraph{Tasks where the choice is essentially neutral.} On property prediction, molecule captioning, text-based generation, molecule editing, single-property optimization on $S^2$-Bench, and MoleculeQA, the two notations track each other to within metric noise. These are tasks in which neither validity-by-construction nor surface-level pattern matching is the bottleneck: the model is either reading molecular sequence content as input (with SMILES and SELFIES carrying equivalent information modulo the tokenizer) or generating short outputs whose correctness is dominated by semantic content rather than by the grammar of the notation.

\subsection{Where Scaling from 1.7B to 4B Helps, and Where It Does Not}
\label{sec:discussion:scaling}

Our two within-family checkpoints (\method-1.7B and \method-4B), trained under an otherwise identical recipe, allow a clean per-task readout of how much each capability benefits from a $\sim$2.4$\times$ increase in parameter count. The results are markedly heterogeneous, and we find it useful to group sub-tasks into three regimes.

\begin{figure}[t]
    \centering
    \includegraphics[width=0.95\linewidth]{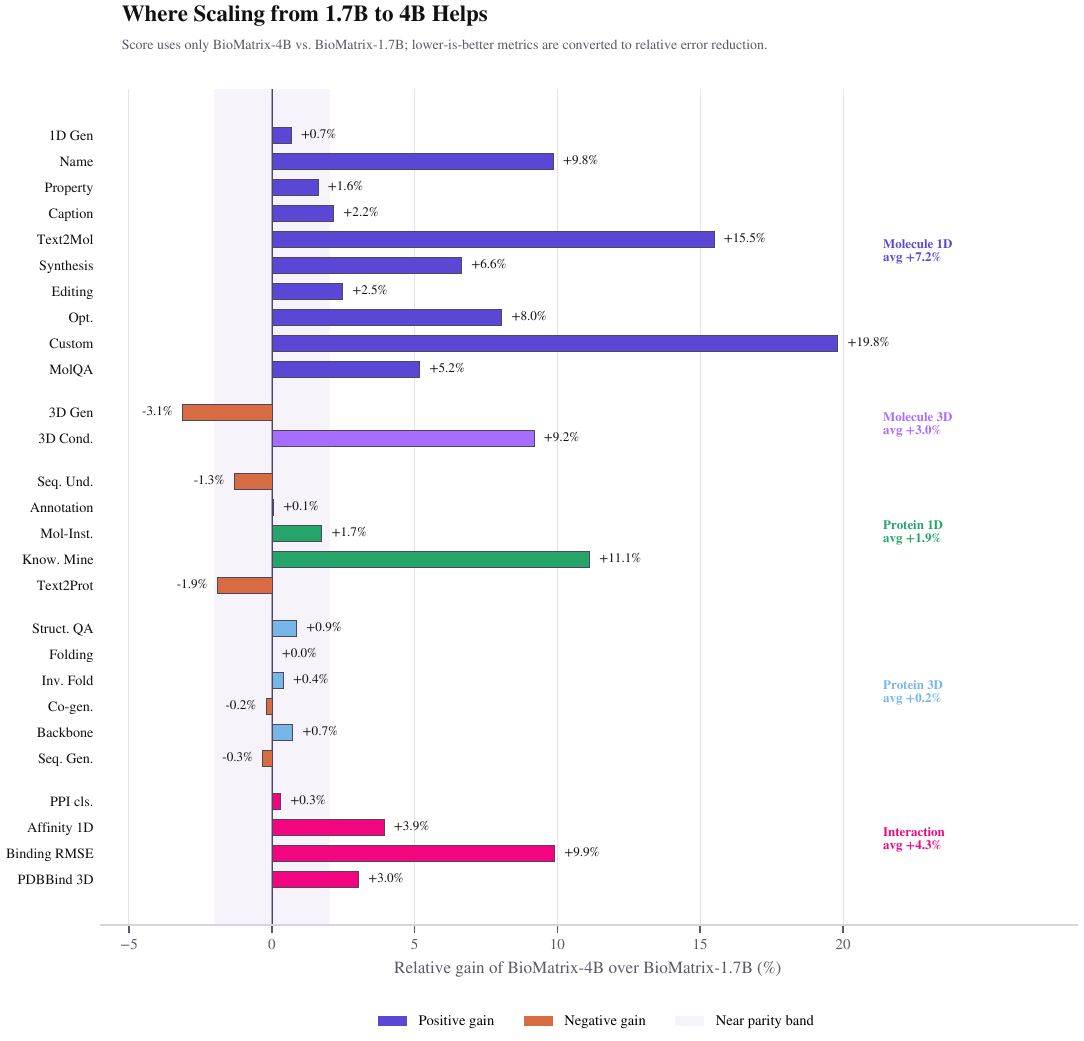}
    \caption{Relative gain from scaling \method from 1.7B to 4B. The score compares only the two \method scales: higher-is-better metrics use the relative score increase over \method-1.7B, while lower-is-better metrics use the corresponding relative error reduction.}
    \label{fig:scaling_gain}
\end{figure}

\paragraph{Strongly scaling tasks.} The largest and most consistent gains from 1.7B to 4B appear on tasks that combine non-trivial knowledge recall with structured generation. Molecular name conversion (Section~\ref{sec:mol_1d_name}) gains 14--15 absolute points on I2F EM (79.45 $\to$ 94.09 for SMILES) and 5--6 on I2S EM. Text-based molecule generation (Section~\ref{sec:mol_1d_mc_mg}) gains 9--10 absolute EM points (56.35 $\to$ 65.07 for SMILES), while customized generation shows a large relative gain despite remaining a difficult low-success regime. Forward synthesis EM gains $\sim$5 points (72.75 $\to$ 77.94), optimization improves across the aggregate success metrics, and MoleculeQA gains $\sim$3.5 absolute points on Total accuracy. On the protein side, the clearest relative gain is knowledge mining, where the 4B model improves substantially over the 1.7B variant. We interpret these as tasks where the bottleneck is the model's ability to compress a large, heterogeneous training distribution into reusable factual and structural knowledge---a regime in which additional parameters reliably help.

\paragraph{Weakly scaling or saturated tasks.} A second cluster of tasks shows little to no improvement from scaling, typically because the 1.7B variant has already saturated the relevant metric or because the task is dominated by benchmark noise. Unconditional molecular generation is essentially saturated at 1.7B (validity, uniqueness, and novelty all sit at or near their ceilings on both MOSES and GuacaMol). Property prediction and molecule captioning show only modest relative gains, and the aggregate annotation and structure-QA scores on the protein side are likewise close to parity between the two scales. Sequence understanding (EC number, fold type, subcellular localization) and inverse folding AAR are within 1 point between the two scales. These flat scaling trends should therefore not be read as poor task performance: in several cases \method is already at or near the best reported result, leaving little headroom for a larger checkpoint to show a measurable gain.

\paragraph{Anti-scaling and high-variance tasks.} A small number of tasks exhibit non-monotone behavior between the two scales. The most visible case is regression on small datasets: ESOL RMSE rises from 0.810 (\method-1.7B SMILES) to 1.240 (\method-4B SMILES), and a similar but smaller fluctuation appears on Lipo. On EC number prediction (Price split) and a few interaction sub-tasks, the 1.7B variant slightly outperforms the 4B variant. Furthermore, interaction tasks should not be read as a strong scaling regime despite occasional positive deltas in Figure~\ref{fig:scaling_gain}: they are mostly small-data, single-task SFT variants, where optimization noise and benchmark variance can be comparable to the apparent gain.

\subsection{Native Multimodality Pays Off Most on Cross-Modal and Cross-Entity Tasks}
\label{sec:discussion:native-multimodal}

A subtler but, we believe, important pattern across the evaluation is where the unified-tokenization design pays off most relative to specialized baselines. On tasks that operate within a single modality and a single entity type---unconditional molecule generation, sequence-only protein understanding, classification on MoleculeNet---\method matches but does not dramatically exceed dedicated specialists; the unification does not hurt, but it also does not unlock new headroom. On tasks that explicitly cross modalities or entities, by contrast, the gains over the strongest specialist baselines widen substantially: property-conditioned molecular conformer generation, text-based molecule generation, sequence-based molecule--protein affinity prediction, and structure-grounded molecule--protein affinity prediction.

Taken together, the data directly validate the effectiveness of this design: the value of placing sequences, structures, and natural language for both molecules and proteins in a single discrete token space is not primarily that it improves any individual single-modality task, but that it makes the cross-modal and cross-entity tasks tractable under one next-token objective, without bolt-on encoders, projectors, or fusion modules.

\subsection{Tokenization Is the Bottleneck on Fine-Grained 3D Geometry}
\label{sec:discussion:3d-bottleneck}

A final cross-cutting observation concerns the limits of the current tokenizers on geometry. On tasks evaluated at the level of distributions or backbone topology---unconditional molecular conformer generation on graph-aggregated metrics, unconditional protein backbone generation, sequence--structure co-generation, structure understanding---\method matches or surpasses dedicated diffusion specialists. On tasks evaluated at the level of fine-grained per-atom geometry, however, a residual gap remains: bond-length MMD on QM9-2014 is loose relative to NExT-Mol (Section~\ref{sec:mol_3d_uncond}), and folding RMSD trails ESMFold on the PDB date split (Section~\ref{sec:prot_3d_fold}). Both gaps are concentrated in the geometric reconstruction step rather than in upstream modeling: a single MMFF refinement closes most of the QM9 gap (FCD: 1.04 $\to$ 0.23, atom stability 0.897 $\to$ 0.985), suggesting that the limitation lies in the irreducible quantization error of finite codebooks combined with autoregressive coordinate reconstruction, rather than in the language model's ability to learn the geometric distribution.

Two implications follow. First, the unified-tokenization paradigm is not in tension with high-fidelity structural outputs---a cheap geometric post-processing step recovers most of the precision an end-to-end diffusion specialist provides, and pairing a unified token-level generator with such a refinement step is a practical design pattern. Second, advances in structural tokenization---larger or hierarchical codebooks, joint tokenization of side chains, or non-autoregressive decoding schemes that reconstruct coordinates globally rather than atom-by-atom---are likely to translate directly into improvements for any model built on top of them.

\section{Conclusion}
\label{section:conclusion}

We presented \method, a multimodal foundation model that natively integrates sequences, structures, and natural language for both molecules and proteins within a single decoder-only architecture. By consolidating all modalities into a shared discrete token space, \method handles heterogeneous biomolecular inputs and outputs uniformly under a single next-token prediction objective, without external encoders, projection adapters, or modality-specific output heads. Through continual pretraining on a large-scale multimodal corpus and instruction tuning across a comprehensive task suite, \method achieves state-of-the-art or competitive performance on the vast majority of evaluated benchmarks, demonstrating that a single generalist backbone can match or surpass specialized models across a broad spectrum of biological tasks. We hope \method serves as a useful step toward unified biological foundation models that bridge sequences, structures, and language across diverse biomolecular entities.

\section{Limitations}
\label{section:limitations}

Despite the breadth and consistency of the empirical results, \method has several limitations that we believe are important to acknowledge, and that point to concrete directions for future work.

\paragraph{Molecular and protein 3D structures are tokenized in disjoint spaces.} \method employs two structure tokenizers that operate independently of each other, with codebooks anchored in disjoint geometric reference frames. While this design works well within a single biomolecular entity, neither tokenizer is equipped to express the relative pose between a small molecule and a protein pocket. As a result, \method cannot natively represent or generate biomolecular \emph{complexes}, which limits its applicability to tasks such as structure-based docking and pocket-conditioned ligand design. A unified 3D tokenizer that operates on a shared geometric reference frame across entity types is the next step.

\paragraph{Potential overlap between continual pretraining and SFT/evaluation data.} We did not perform dedicated entity-level filtering between the continual pretraining corpus and the downstream SFT/evaluation corpora. Consequently, molecules, proteins, and their associated annotations appearing in SFT or evaluation tasks may have been implicitly observed during continual pretraining, either through related database entries, literature text, or cross-entity resources. This is a practical limitation of building broad biological foundation models from public molecular and protein corpora, where the same entities and metadata are repeatedly reused across databases and benchmarks. Our goal in this work is therefore not to claim a strictly contamination-free benchmark setting, but to build a strong and comprehensive base model that the community can further post-train on custom tasks with task-specific data curation.

\paragraph{General-purpose capabilities are partially eroded by domain specialization.} Although we incorporate FineWeb-Edu and MegaScience into the pretraining and instruction-tuning corpora to mitigate catastrophic forgetting, the heavy concentration of biomolecular tokens in the 304.4B-token corpus inevitably shifts the model away from the general language and reasoning capabilities of the underlying Qwen3-Base. We have not systematically benchmarked \method on general-purpose instruction-following or reasoning suites, and we expect a measurable gap relative to the original Qwen3 checkpoints on such tasks. 

\paragraph{Task-group SFT rather than a single unified SFT model.} As detailed in Section~\ref{sec:discussion:why-grouped-sft}, we train a small set of task-group-specific SFT variants rather than one monolithic instruction-tuned model, because in our setting the pattern-monolithic nature of biological SFT data causes large-data sub-tasks to overwhelm small-data ones when mixed naively. While we release a single all-tasks SFT model with mild oversampling for practical use, this unified variant still trails the task-group variants on a non-trivial fraction of small-data sub-tasks. Closing this gap (via more principled data balancing, curriculum design, or training dynamics tailored to highly imbalanced multi-task biological corpora) is an open problem that we have not fully solved within the recipe explored here.

\paragraph{Coverage of biomolecular entities and modalities.} \method covers small molecules and proteins, but does not currently model other biologically important entities such as nucleic acids (DNA, RNA), carbohydrates, or lipids. Extending the unified tokenization scheme to these additional entities and modalities is conceptually compatible with the framework presented here, and we view it as a natural direction for scaling the paradigm.

\bibliographystyle{plainnat}
\setcitestyle{numbers}
\bibliography{paper}

\beginappendix

\begin{landscape}
\begin{longtable}{lcllccc}
\caption{Detailed training and test data statistics for each sub-task. 
}
\label{tab:sft_data_stat_detail} \\
\toprule
Domain & Modality & Task & Name & \#Train & \#Test & Source \\
\midrule
\endfirsthead

\multicolumn{7}{c}{\tablename\ \thetable\ -- \textit{Continued from previous page}} \\
\toprule
\textbf{Domain} & \textbf{Modality} & \textbf{Task} & \textbf{Name} & \textbf{\#Train} & \textbf{\#Test} & \textbf{Source} \\
\midrule
\endhead

\midrule
\multicolumn{7}{r}{\textit{Continued on next page}} \\
\endfoot

\bottomrule
\endlastfoot

\multirow{40}{*}{Molecule} & \multirow{37}{*}{1D}
  & \multirow{2}{*}{Unconditional Sequence Generation}
   & MOSES & \num{1936962} & \num{10000} & OMG-GPT~\cite{omggpt} \\
 & & & GuacaMol & \num{1260508} & \num{10000} & OMG-GPT~\cite{omggpt} \\
\cmidrule(lr){3-7}
& & \multirow{4}{*}{Name Conversion}
   & IUPAC-to-formula (I2F) & 300000 & 2993 &  SMolInstruct~\cite{llasmol} \\
 & & & IUPAC-to-sequence (I2S) & 274053 & 2764 &  SMolInstruct~\cite{llasmol} \\
 & & & sequence-to-formula (S2F) & 274053 & 2764 &  SMolInstruct~\cite{llasmol} \\
 & & & sequence-to-IUPAC (S2I) & 274053 & 2764 &  SMolInstruct~\cite{llasmol} \\
 
\cmidrule(lr){3-7}
 & & \multirow{6}{*}{Property Prediction}
    & ESOL & 888 & 112 & SMolInstruct~\cite{llasmol} \\
 & & & Lipo & 3359 & 420 & SMolInstruct~\cite{llasmol} \\
 & & & BBBP & 1489 & 185 & SMolInstruct~\cite{llasmol} \\
 & & & ClinTox & 1131 & 143 & SMolInstruct~\cite{llasmol} \\
 & & & HIV & 30370 & 3813 & SMolInstruct~\cite{llasmol} \\
 & & & SIDER & 20240 & 2340 & SMolInstruct~\cite{llasmol} \\
 
\cmidrule(lr){3-7}
 & & \multirow{3}{*}{Molecule Captioning}
   & SMolInstruct   & 55611 & 2327 & SMolInstruct~\cite{llasmol} \\
 & & & CheBI-20 & 26407 & 3300 & MolT5~\cite{molt5} \\
 & & & KnowMol-100K & 399996 & -- & KnowMol-100K~\cite{knowmol} \\
 
\cmidrule(lr){3-7}
 & & \multirow{3}{*}{Text-Based Molecule Generation}
   & SMolInstruct & 55611 & 2282 & SMolInstruct~\cite{llasmol} \\
 & & & CheBI-20 & 26407 & 3300 & MolT5~\cite{molt5} \\
 & & & KnowMol-100K & 99999 & -- & KnowMol-100K~\cite{knowmol} \\

\cmidrule(lr){3-7}
 & & \multirow{3}{*}{Customized Molecule Generation}
     & AtomNum & 133334 & 5000 & S\textsuperscript{2}-Bench~\cite{s2bench} \\
 & & & BondNum & 133334 & 5000 & S\textsuperscript{2}-Bench~\cite{s2bench} \\
 & & & FuncGroup & 133334 & 5000 & S\textsuperscript{2}-Bench~\cite{s2bench} \\
 
\cmidrule(lr){3-7}
 & & Forward Synthesis & SMolInstruct & 971809 & 4062 & SMolInstruct~\cite{llasmol} \\
 
\cmidrule(lr){3-7}
 & & \multirow{2}{*}{Retrosynthesis}
   & SMolInstruct  & 941735 & 4156 & SMolInstruct~\cite{llasmol} \\
 & & & USPTO-50K & 40008 & 5007 & USPTO-50K~\cite{uspto_50k} \\
 
\cmidrule(lr){3-7}
 & & \multirow{3}{*}{Molecule Editing}
   & AddComp & 133333 & 5000 & S\textsuperscript{2}-Bench~\cite{s2bench} \\
 & & & DelComp & 133333 & 5000 & S\textsuperscript{2}-Bench~\cite{s2bench} \\
 & & & SubComp & 133333 & 5000 & S\textsuperscript{2}-Bench~\cite{s2bench}\\
 
\cmidrule(lr){3-7}
 & & \multirow{11}{*}{Molecule Optimization}
    & qed+ & 145991 &8122 & MolOpt-Instructions~\cite{drugassist} \\
& & & acceptor+ & 154492 &8583 & MolOpt-Instructions~\cite{drugassist} \\
& & & donor+ & 154492 &8583 & MolOpt-Instructions~\cite{drugassist} \\
& & & solubility+ & 154493 &8583 & MolOpt-Instructions~\cite{drugassist} \\
& & & bbbp+ & 154493 &8583 & MolOpt-Instructions~\cite{drugassist} \\
& & & herg- & 154493 &8582 & MolOpt-Instructions~\cite{drugassist} \\
& & & others & 263501 & -- & MolOpt-Instructions~\cite{drugassist} \\
& & & logP & 133333 &5000 & S\textsuperscript{2}-Bench~\cite{s2bench} \\
& & & MR & 133333 &5000 & S\textsuperscript{2}-Bench~\cite{s2bench} \\
& & & QED & 133333 & 5000 &S\textsuperscript{2}-Bench~\cite{s2bench} \\

\cmidrule(lr){3-7}
 & & \multirow{4}{*}{Molecule Question Answering}
  & Structure & 32176 &3113 & MoleculeQA~\cite{moleculeqa} \\
& & & Source & 11062 &1343 & MoleculeQA~\cite{moleculeqa} \\
& & & Property & 4838 &731 & MoleculeQA~\cite{moleculeqa} \\
& & & Usage & 1917 &599 & MoleculeQA~\cite{moleculeqa} \\
\cmidrule(lr){2-7}

 & \multirow{7}{*}{3D}
 & Unconditional Structure Generation & QM9-2014 & \num{97834} & \num{10000} & NExT-Mol~\cite{nextmol} \\
\cmidrule(lr){3-7}
 & & \multirow{6}{*}{Conditional Structure Generation}
   & QM9-2014-$\mu$ & \num{48849} & \num{10000} & NExT-Mol~\cite{nextmol} \\
 & & & QM9-2014-$\alpha$ & \num{48849} & \num{10000} & NExT-Mol~\cite{nextmol} \\
 & & & QM9-2014-$C_{v}$ & \num{48849} & \num{10000} & NExT-Mol~\cite{nextmol} \\
 & & & QM9-2014-$\varepsilon_{\textnormal{HOMO}}$ & \num{48849} & \num{10000} & NExT-Mol~\cite{nextmol} \\
 & & & QM9-2014-$\varepsilon_{\textnormal{LUMO}}$ & \num{48849} & \num{10000} & NExT-Mol~\cite{nextmol} \\
 & & & QM9-2014-$\Delta\varepsilon$ & \num{48849} & \num{10000} & NExT-Mol~\cite{nextmol} \\
\midrule

\multirow{30}{*}{Protein} & \multirow{25}{*}{1D}
& \multirow{6}{*}{Sequence Understanding}
   & CLEAN\_EC\_number\_new & \multirow{2}{*}{\num{227362}} & 392 & OPI~\cite{opi} \\
& & & CLEAN\_EC\_number\_price & & 149 & OPI~\cite{opi} \\
\cmidrule(lr){4-7}
& & & Subcellular\_localization & \num{11230} & 2772 & OPI~\cite{opi} \\
\cmidrule(lr){4-7}
& & & Fold\_holdout & \multirow{3}{*}{\num{12311}} & 718 & OPI~\cite{opi} \\
& & & Superfamily\_holdout &  & \num{1254} & OPI~\cite{opi} \\
& & & Family\_holdout &  & \num{1272} & OPI~\cite{opi} \\
 
\cmidrule(lr){3-7}
 & & \multirow{13}{*}{Annotation Prediction}
   & CASPSimilarSeq\_function & \multirow{3}{*}{\num{451618}} & 184 & OPI~\cite{opi} \\
   & & & IDFilterSeq\_function & & \num{1112} & OPI~\cite{opi} \\
   & & & UniProtSeq\_function & & \num{4562} & OPI~\cite{opi} \\
   \cmidrule(lr){4-7}
   & & & CASPSimilarSeq\_go & \multirow{3}{*}{\num{451618}} & 184 & OPI~\cite{opi} \\
   & & & IDFilterSeq\_go & & \num{1112} & OPI~\cite{opi} \\
   & & & UniProtSeq\_go & & \num{4562} & OPI~\cite{opi} \\
   \cmidrule(lr){4-7}
   & & & CASPSimilarSeq\_keywords & \multirow{3}{*}{\num{451618}} & 184 & OPI~\cite{opi} \\
   & & & IDFilterSeq\_keywords & & \num{1112} & OPI~\cite{opi} \\
   & & & UniProtSeq\_keywords & & \num{4562} & OPI~\cite{opi} \\
   \cmidrule(lr){4-7}
 & & & Catalytic Activity & \num{51573} & \num{1601} & Mol-Instructions~\cite{mol_instructions} \\
 & & & Domain Motif & \num{43700} & \num{5876} & Mol-Instructions~\cite{mol_instructions} \\
 & & & Protein Function & \num{110689} & \num{3494} & Mol-Instructions~\cite{mol_instructions} \\
 & & & General Function & \num{83939} & \num{2633} & Mol-Instructions~\cite{mol_instructions} \\
\cmidrule(lr){3-7}
 & & \multirow{3}{*}{Knowledge Mining}
   & Gene\_name\_to\_cancer & 590 & 148 & OPI~\cite{opi} \\
   & & & Gene\_symbol\_to\_cancer & 590 & 148 & OPI~\cite{opi} \\
   & & & Gene\_symbol\_to\_tissue & 8,723 & 2,181 & OPI~\cite{opi} \\
\cmidrule(lr){3-7}
 & & \multirow{2}{*}{Text-Based Protein Design}
   & CAMEO & 391,933 & 507 & ProDVa~\cite{prodva} \\
 & & & Molinst-SwissProtCLAP & 712,248 & 5,876 & ProDVa~\cite{prodva} \\
\cmidrule(lr){3-7}
 & & Unconditional Sequence Generation & PDB, AFDB & \num{220285} & \num{500} & DPLM-2~\cite{dplm2} \\
\cmidrule(lr){2-7}

 & \multirow{5}{*}{3D}
 & Structure Understanding          & PFUD & \num{391771} & \num{7702} & ProtTex~\cite{prottex} \\
\cmidrule(lr){3-7}
 & & Structure Prediction (Folding) & PDB, AFDB & \num{220285} & \num{447} & DPLM-2~\cite{dplm2} \\
\cmidrule(lr){3-7}
 & & Inverse Folding                & PDB, AFDB & \num{220285} & \num{447} & DPLM-2~\cite{dplm2} \\
\cmidrule(lr){3-7}
 & & Sequence--Structure Co-Generation & PDB, AFDB & \num{220285} & \num{500} & DPLM-2~\cite{dplm2} \\
\cmidrule(lr){3-7}
 & & Unconditional Structure Generation & PDB, AFDB & \num{220285} & \num{500} & DPLM-2~\cite{dplm2} \\
\midrule

\multirow{6}{*}{Interaction} & \multirow{5}{*}{1D}
 & \multirow{2}{*}{Molecule--Protein Interaction}
   & BindingDB & \num{7900} & \num{5230} & PEER~\cite{peer} \\
 & & & PDBBindv2019 & \num{16434} & \num{285} & PEER~\cite{peer} \\
\cmidrule(lr){3-7}
 & & \multirow{3}{*}{Protein--Protein Interaction}
   & Yeast & \num{4945} & \num{394} & PEER~\cite{peer} \\
 & & & Human & \num{35669} & \num{237} & PEER~\cite{peer} \\
 & & & PPI & \num{2421} & \num{326} & PEER~\cite{peer} \\
\cmidrule(lr){2-7}

 & 3D & Molecule--Protein Interaction & PDBBindv2020 & \num{18013} & \num{359} & PDBBindv2020~\cite{pdbbind} \\
\midrule

Text & -- & -- & MegaScience & \num{1253230} & -- & MegaScience~\cite{megascience} \\

\end{longtable}
\end{landscape}

\end{document}